\pdfoutput=1
\documentclass{article}
\usepackage{iclr2027_conference,times}
\usepackage[T1]{fontenc}
\usepackage{amsmath,amssymb}
\usepackage{graphicx}
\usepackage{booktabs}
\usepackage{hyperref}
\usepackage{url}

\usepackage{tabularx}
\usepackage{xcolor}
\usepackage{multirow}
\usepackage{enumitem}

\title{Promises \& Perils of EEG Foundation Models: On the Role of Pretraining}
\author{Tom Maye-Lasserre, Van-Tam Nguyen, Nikola Simidjievski \\
LTCI, T\'el\'ecom Paris, Institut Polytechnique de Paris, France \\
{\small\texttt{\{tom.maye-lasserre,van-tam.nguyen,nikola.simidjievski\}@telecom-paris.fr}}
}

\iclrfinalcopy

\begin{document}
\maketitle
\lhead{Preprint}

\begin{abstract}
Foundation models for electroencephalography (EEG) typically rely on masked prediction, with the promise that pretraining on large unlabeled collections yields general-purpose representations. Yet published evaluations rarely isolate what pretraining itself contributes, leaving open whether reported gains come from representations learned from unlabeled EEG or from the encoder and adaptation protocol. We ask what pretraining contributes, whether its benefits are task-dependent, and whether the pretext task is aligned with all downstream tasks or only with some. Across 23 published encoder configurations and 34 datasets, evaluated with frozen weights, we find that pretraining gains are uneven: substantial for sleep staging and clinical EEG, but small or absent for motor imagery and event-related potentials, and almost entirely diminished once classifiers also receive handcrafted spectral and covariance features. To understand why, we pretrain encoders on transformed EEG that preserves only selected statistics of the original signals. Transformations retaining spectra and cross-spectra alone reproduce the benefits of original-EEG pretraining, and published encoders keep their gains when they read such signals; yet the same transformation substantially impairs supervised motor-imagery and ERP decoding. Finally, adding a supervised term to reconstruction shows these representations to be learnable from the same data at negligible cost to the pretext objective; but removing supervision erodes most of the acquired decoding gains with little change in reconstruction loss. In the configurations studied, good reconstruction therefore does not assure the acquisition or retention of representations useful for motor-imagery and ERP decoding. These findings highlight objective–task alignment as a central challenge for EEG pretraining, motivating objectives better aligned with downstream tasks. \looseness -1

\end{abstract}

\section{Introduction}
\label{sec:introduction}

Foundation models for electroencephalography (EEG) promise to turn a single investment in large-scale pretraining into reusable representations for many downstream tasks. By learning from thousands of hours of unlabeled recordings, one encoder could support diverse EEG applications, including sleep staging, clinical classification, and decoding of motor imagery (MI) and event-related potentials (ERPs), with each new task requiring only a lightweight readout trained on its own labels. A growing body of work pursues this goal through different model architectures, scales, and pretraining data \citep{labram,cbramod,reve,eegpt,luna,csbrain}. 

Whether this promise holds is usually assessed through downstream performance. Such performance, however, reflects an entire pipeline: the architecture, preprocessing, pretrained weights, classifier, and downstream training procedure \citep{eegfmbench,eegbench}. Comparing these pipelines can guide model selection, but does not by itself establish what pretraining contributes. A model may perform well because its architecture already provides useful features, or because supervised adaptation learns what the downstream task requires. In the context of pretraining, many current approaches rely on predicting masked portions of the signal or their representations, on the premise that learning to recover missing information will capture structure useful beyond the pretraining task. To assess the value of pretraining itself, we must isolate the benefit of the pretrained weights and determine how that benefit varies across tasks. For EEG, this also requires an informative reference: spectral and spatial descriptors have long supported decoding \citep{ang2012fbcsp,barachant2012riemannian}. The question is therefore not only whether pretraining helps, but whether it adds predictive value beyond these established representations.

This question matters because, by design, self-supervised learning optimizes a pretext task rather than a particular application task downstream. As such, its usefulness depends on alignment between the information needed to solve that pretext task and the transferable information needed for later downstream prediction \citep{lee2021predicting}. Recovering masked EEG signal may require some properties of a recording without requiring all the structure relevant to its labels. Consequently, success at masked prediction need not imply equally useful representations for every EEG task. Therefore, understanding this relationship requires more than evaluating published checkpoints: it requires interventions on pretraining to examine what the objective needs to learn and whether that matches what downstream tasks need.

In this paper, we investigate what EEG pretraining contributes and what determines the scope of its benefits. We perform a systematic and extensive empirical evaluation of published approaches in a series of controlled experiments, organized around three questions:

\begin{enumerate}[leftmargin=15pt]
\item \textbf{What does pretraining add across EEG tasks?}
We compare 23 published encoder configurations with randomly initialized copies of the same architectures across 34 datasets and tasks (Section~\ref{sec:results}). Frozen encoders and matched readouts (classification heads) isolate the contribution of pretrained weights under a common evaluation protocol. Repeating the comparison with classical spectral and spatial descriptors supplied to both readouts tests whether pretraining provides additional predictive value beyond these descriptors. We find that this is not usually the case: pretraining gains, substantial for sleep staging and clinical classification but small or absent for motor imagery and ERPs, nearly vanish once both readouts receive the descriptors. \looseness -1
\item \textbf{Does pretraining rely on the same signal properties as downstream decoding?}
We intervene on pretraining itself and ask what it needs to solve its pretext task: we ran several pretrainings, on original EEG but also on synthetic transformations of the same recordings (Section~\ref{sec:surrogates}). And pretraining on one transformation that preserves only power spectra and cross-spectra, destroying waveforms and temporal organization, ended up reconstructing masked original EEG nearly as well as pretraining on original recordings, and transferring as well on average across objectives, architectures and model sizes, within $0.007$~$\kappa$ in every task family for REVE-small with masked reconstruction. Yet the same transformation removes information that a supervised EEGNet relies on: ERP decoding falls to chance and motor-imagery decoding loses most of its performance, while sleep, clinical and SSVEP decoding change by less than $0.03$~$\kappa$.

\item \textbf{What role do the encoder, the data, and the objective play?}
To distinguish these possible sources of limited transfer, we train the same encoder architecture on each task's own recordings, comparing self-supervised reconstruction alone with reconstruction augmented by label supervision (Section~\ref{sec:retention}). We then remove the supervised objective and continue reconstruction on the same data. This design tests both whether the encoder can learn task-relevant information from the available recordings and whether reconstruction alone encourages its acquisition and retention. We find that label supervision enables the encoder to learn information relevant to motor imagery and ERPs at little cost in reconstruction error, whereas reconstruction alone yields weak decoding performance and fails to retain most of the gains acquired through supervision.
\end{enumerate}

\section{Isolating the contribution of pretraining across EEG tasks} %
\label{sec:results}

Our first set of analyses focuses on assessing the contribution of pretraining to downstream decoding. To this end, we compare frozen pretrained encoders with randomly initialized variants of the same architectures under matched conditions. Specifically, we evaluate many published encoder configurations across an extensive set of datasets and tasks. Here we aim to assess the gain from pretrained weights across these diverse tasks and datasets. We then extend this analysis, and compare these performances to classical spectral and spatial descriptors supplied to both readouts. We aim with this experiment to measure the remaining performance gain, when such standard, statistical, descriptors are available.

\textbf{Experimental setup.}
We evaluate 23 published encoder configurations spanning masked reconstruction, contrastive learning, self-distillation, autoregressive modeling, and latent prediction (Appendix~\ref{app:s3_models}). The benchmark contains 34 datasets, covering motor imagery, ERP, SSVEP, sleep staging, clinical classification, emotion recognition, mental workload, and imagined speech (Appendix~\ref{app:s3_data}). Each dataset contributes at most 6,000 windows, and subjects are disjoint across training, validation, and test splits. Because EEG foundation models are meant to support downstream learning with few labels, we also repeat the main analyses with 32 and 128 labeled windows per class (Appendix~\ref{app:low_label}).

For each encoder, we compare the pretrained representation with a randomly initialized copy of the same architecture. Specifically, let $Z_P$ denote the tokens of a frozen pretrained encoder, and $Z_0$ those of its frozen, randomly initialized counterpart. We additionally compute classical descriptors, denoted by $H_d$, from the same normalized window. These include channel power spectra, covariance matrices in six frequency bands computed over the whole window, its halves, and its quarters, and coherency between channels. Because these descriptors have substantially higher dimensionality than the pooled encoder features, we use PCA to limit the dimensional imbalance between the two feature blocks, retaining at most $d$ components, where $d$ is the pooled encoder feature width. The projection is fitted without labels on the training and validation windows (Appendix~\ref{app:representations}). Repeating the analysis without PCA preserves the main conclusions, indicating that they are not driven by dimensionality reduction (Appendix~\ref{app:s3_width}, Table~\ref{tab:s3_width}).

For each representation, we freeze the encoder and train a small head on top of it. The head applies attention pooling to collapse the encoder's tokens into a single vector $Z$, then feeds $Z$ to a linear classifier. Pooling keeps the number of head parameters independent of the token count, which varies widely across architectures. When $H_d$ is used, it is concatenated with $Z$ before the classifier. The only trained parameters are the pooling query and the linear layer (Table~\ref{tab:pipelines} lists what is trained in each experiment); alternative pooling and readout choices are examined in Appendices~\ref{app:s3_matched_readouts} and~\ref{app:s3_reader_controls}. Encoders remain frozen by design: once target labels update the encoder, its representation mixes what pretraining produced with what supervision induced, and a gain can no longer be attributed to the pretrained weights (Appendix~\ref{app:why_frozen}). Full fine-tuning instead measures the value of pretraining as an initialization; a targeted comparison on REVE-base shows that, at the full label budget, it raises the scores of pretrained and random encoders by similar amounts ($+0.05$ and $+0.06$~$\kappa$), leaving a pretrained--random gap comparable to the frozen one (Appendix~\ref{app:full_finetuning}). LoRA and LP-FT leave a similar gap. Fine-tuning converges faster from pretrained weights only on the tasks where frozen features already benefit, and the fine-tuned model does not exceed EEGNet on motor imagery or ERPs.

Within each pretrained--random pair, both heads see the same labeled windows and test subjects and follow the same hyperparameter search, with hyperparameters chosen separately for each head (Appendix~\ref{app:s3_readouts}).

With $s(X)$ denoting the test score of a head trained on features $X$, we measure
\begin{align}
    \Delta_{\mathrm{pre}} &= s(Z_P) - s(Z_0), \label{eq:delta_pre}\\
    \Delta_H &= s([H_d, Z_P]) - s([H_d, Z_0]). \label{eq:delta_h}
\end{align}

$\Delta_{\mathrm{pre}}$ measures the gain obtained from the pretrained weights over the same randomly initialized architecture, while $\Delta_H$ measures how much of this gain remains when both heads are also provided with the classical descriptors. Comparing absolute scores across models would not isolate this contribution, since differences between architectures can already be present with random weights (Appendix~\ref{app:s3_absolute_results}). The performance is measured as Cohen's $\kappa$ ($0$ at chance), averaged over three head initializations. A family score averages all its model--dataset pairs, and the overall score weights the five main families equally (Appendix~\ref{app:s3_statistics}).

\subsection{Pretraining gains concentrate in sleep and clinical tasks}
\label{sec:results_where}

\begin{figure}[h]
    \centering
    \includegraphics[width=\linewidth]{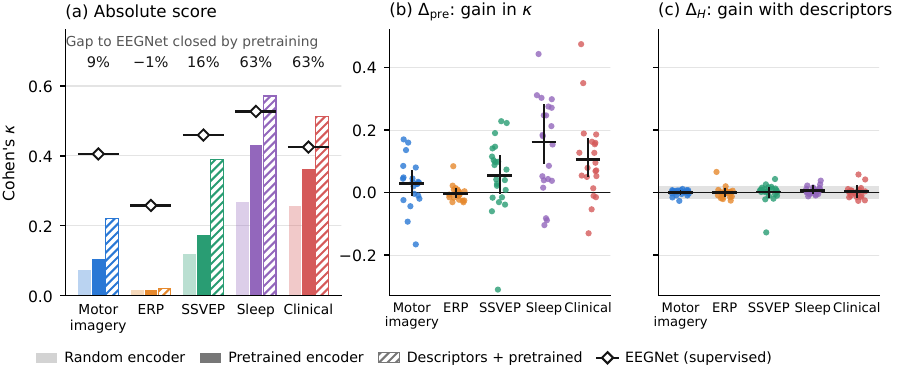}
    \caption{\textbf{Pretraining gains are uneven, nearly vanish once classical descriptors are available, and leave motor imagery and ERPs far from a supervised reference.} Frozen encoders with matched attention-pooling heads. \textbf{(a)} Absolute Cohen's $\kappa$ of random encoders, pretrained encoders and pretrained encoders with classical descriptors, family means over model--dataset pairs. Diamonds: EEGNetv4 trained from scratch on the same windows and splits (own output, mean of three training seeds). Percentages: share of the distance from the random encoder to EEGNetv4 covered by pretraining (Table~\ref{tab:stats_benchmark}). \textbf{(b)} Gain of each pretrained encoder over its randomly initialized copy ($\Delta_{\mathrm{pre}}$). \textbf{(c)} The same gain when both heads also receive the descriptors ($\Delta_H$); the grey band is the $\pm 0.02~\kappa$ equivalence margin. In (b, c) each point averages one encoder over the datasets of a family; black bars show the mean over all model--dataset pairs with 95\% intervals from a bootstrap over models and datasets (Appendix~\ref{app:s3_statistics}). Results for the remaining three families are in Appendix~\ref{app:s3_absolute_results}.}
    \label{fig:benchmark_gains}
\end{figure}

Averaged equally across the five main task families, pretrained weights improve Cohen's $\kappa$ by $0.070$, although the gains vary substantially across tasks (Figure~\ref{fig:benchmark_gains}b). The largest gains are observed for sleep staging ($+0.162$) and clinical classification ($+0.105$), which are also the only two task families whose 95\% intervals exclude zero. Gains are smaller for SSVEP ($+0.055$) and motor imagery ($+0.029$), while almost no improvement is observed for ERP ($-0.002$). We find this task dependence to be significant ($p<0.001$; Table~\ref{tab:stats_benchmark}). It is not a ceiling effect: on motor imagery and ERPs, a supervised EEGNetv4 reaches $0.41$ and $0.26$, far above the pretrained encoders. On motor imagery, pretraining raises the score only from $0.07$ to $0.10$, and on ERPs not at all ($0.02$), whereas on sleep staging it raises it from $0.27$ to $0.43$, most of the way to EEGNetv4 ($0.53$; Figure~\ref{fig:benchmark_gains}a).

The experiments also show that this pattern extends across architectures. For 14 of the 19 encoders evaluated in all five families, the gains on both sleep and clinical tasks exceed those on both motor imagery and ERP (Appendix~\ref{app:s3_corpus_model_results}; 16 of 19 when comparing the average of each pair of families, Wilcoxon $p=0.001$). The difference between absolute performance and the contribution of pretraining is also evident across individual models. For example, on MI~\citep{stieger2021}, pretrained REVE outscores pretrained EEG-DINO (medium) by $0.18$~$\kappa$, although both gain about $0.05$ from pretraining and the difference between them is already present with random weights (Appendix~\ref{app:s3_statistics}).

\subsection{Pretraining adds little beyond classical descriptors}
\label{sec:results_overlap}

When the same classical descriptors are provided to both heads, the mean pretrained--random gap decreases from $0.070$ to $0.002$~$\kappa$ (Figure~\ref{fig:benchmark_gains}c), and is equivalent to zero within $\pm0.02$~$\kappa$ in all five main families (Table~\ref{tab:stats_benchmark}). The reduction is particularly pronounced for sleep staging and clinical classification, where the gains from pretraining are the largest when using encoder features alone, but fall below $0.01$~$\kappa$ once the descriptors are included. Power spectra alone account for most of this reduction in these two families (Appendix~\ref{app:s3_descriptor_results}). At the same time, adding the classical descriptors to pretrained features improves absolute performance by $0.12$ to $0.21$~$\kappa$ in every main family except ERP (Appendix~\ref{app:s3_absolute_results}).

These results indicate substantial overlap in predictive value between the pretrained representations and classical EEG descriptors under the evaluated readouts. They do not, however, establish that pretrained representations contain no additional information. Some model-specific exceptions remain. Specifically, REVE retains a mean ERP gain of $0.066$, while LEAD retains a mean clinical gain of $0.058$ (Appendix~\ref{app:s3_corpus_model_results}). The same general reduction is observed with mean pooling, electrode-wise pooling, and without PCA. Although the residual gain depends on these choices, it remains below $0.04$~$\kappa$ in every main family (Appendices~\ref{app:s3_matched_readouts}--\ref{app:s3_reader_controls}). The same pattern is observed in the three additional families (Appendix~\ref{app:s3_absolute_results}).

The near-zero gain from pretraining does not imply that classical descriptors exhaust the predictive information available for these tasks. EEGNet trained with supervision on the same tasks achieves substantial further improvements (Appendix~\ref{sec:results_headroom}). This shows that additional predictive information can be learned, even when the evaluated pretrained representations provide little additional benefit beyond classical descriptors.

\section{Signal properties that support pretraining and downstream decoding}
\label{sec:surrogates}

We now ask whether the signal properties that suffice for effective pretraining are the same ones that downstream decoding relies on. To answer this, we transform EEG windows so that selected statistics are preserved while other signal structure is disrupted, and use each transformation in two complementary experiments. First, pretraining on transformed EEG and evaluating on original EEG tests whether the preserved statistics are sufficient to recover reconstruction and transfer performance. Second, training a supervised decoder on transformed EEG tests whether the task labels remain predictable from those statistics alone. If a transformation leaves pretraining performance intact but impairs supervised decoding, then the structure it removes is needed by the downstream task but not by pretraining, revealing a mismatch between the two.

\textbf{Experimental setup.}
We compare original EEG with four transformations of the same four-second windows. \textit{PSD + cross-spectra} applies the same random Fourier phase rotation to all channels at each frequency, preserving whole-window power spectra and complex cross-spectra, including relative phases between channels, while altering waveforms and event timing. \textit{PSD} uses independent rotations across channels, preserving their power spectra while disrupting cross-channel phase relationships. \textit{Covariance} generates temporally white Gaussian signals matching the window's channel means and regularized spatial covariance in expectation. \textit{Mean + variance} generates independent Gaussian noise matching each channel's mean and variance in expectation. These transformations test different combinations of signal properties; they do not form a single ordered sequence of degradations. Spectral preservation concerns the full window, not shorter intervals. Transformed training windows are generated afresh at every presentation (Appendix~\ref{app:surrogate_construction}).

To test what supports pretraining, we train REVE-small~\citep{reve} from scratch on original TUEG recordings~\citep{obeid2016tuh} or their transformed versions, keeping the architecture, optimization settings, and training budget fixed. The main experiment uses masked reconstruction, hiding 75\% of the one-second electrode tokens. Both the visible input and the reconstruction target come from the same original or transformed window. We then evaluate every model on \emph{original EEG}: masked reconstruction is measured by $R_0^2$, one minus squared error divided by the energy of the masked samples, and transfer by a trained attention-pooling and linear classification head on the frozen encoder. Both fitting and evaluating this head use original EEG.

To test what supports downstream decoding, we apply the same transformations to each task's own windows and train a supervised decoder from scratch for each signal condition, retaining the original labels. Training, validation, and testing all use the corresponding signal version and the same subject-disjoint splits across conditions. We use EEGNetv4~\citep{eegnet} as a compact decoder to assess how much label-predictive information remains accessible after each transformation. The relevant comparison is the change from original to transformed EEG within each experiment: whether a transformation affects pretraining and supervised decoding in the same way.

Both experiments use the same downstream windows and subject partitions across six dataset--task combinations: BCI IV-2a~\citep{tangermann2012bcicomp} and PhysioNet MI~\citep{physionet_eegmmidb} for motor imagery, ERP~\cite{lee2019openbmi} and SSVEP~\citep{lee2019openbmi}, ISRUC~\citep{khalighi2016isruc} for sleep staging, and TUAB~\citep{obeid2016tuh} for abnormality detection. Classification is measured by Cohen's $\kappa$; aggregate scores weight the six combinations equally (Appendix~\ref{app:surrogates_setup}).

Additional controls test whether the pretraining comparison extends to JEPA-style latent prediction~\citep{assran2023ijepa}, alone or combined with reconstruction, alternative pooling strategies, a larger encoder, a fivefold training budget, and LaBraM with its native discrete prediction objective (Appendices~\ref{app:s4_objective_controls}, \ref{app:s4_budget}, and~\ref{app:s4_labram_short}), as well as other architectures and contrastive objectives (Appendix~\ref{app:beyond_reve}).

\begin{table}[t]
\centering
\small
\setlength{\tabcolsep}{3pt}
\caption{\textbf{Pretraining and supervised decoding respond differently to the same signal transformations.} Pretrained REVE-small with MAE pretrained on different transformed signal versions and evaluated on original EEG data; EEGNetv4 is trained from scratch and tested on the corresponding version. Scores are averaged equally across six datasets (mean ± SD over three seeds; one pretraining run for Mean + variance). The random baseline uses three encoder initializations.}
\label{tab:pretraining_statistics}
\begin{tabular}{lcc|>{}c}
\toprule
\textbf{Signal version}
& \textbf{Recon. $R_0^2$}
& \textbf{Transfer $\kappa$}
& \textbf{EEGNetv4 $\kappa$} \\
\midrule
Original EEG
& $0.657 \pm 0.003$
& $0.347 \pm 0.005$
& $0.523 \pm 0.008$ \\
PSD + cross-spectra
& $0.648 \pm 0.003$
& $0.349 \pm 0.005$
& $0.374 \pm 0.003$ \\
Covariance
& $0.570 \pm 0.008$
& $0.262 \pm 0.006$
& $0.118 \pm 0.003$ \\
PSD
& $0.053 \pm 0.001$
& $0.283 \pm 0.006$
& $0.284 \pm 0.005$ \\
Mean + variance
& $0.000$
& $0.154$
& $0.000 \pm 0.002$ \\
\midrule
Random initialization
& ---
& $0.146 \pm 0.008$
& --- \\
\bottomrule
\end{tabular}
\end{table}

\subsection{Preserving spectra and cross-spectra recovers pretraining performance}
\label{sec:surrogates_transfer}

\begin{figure}[t]
\centering
\includegraphics[width=\linewidth]{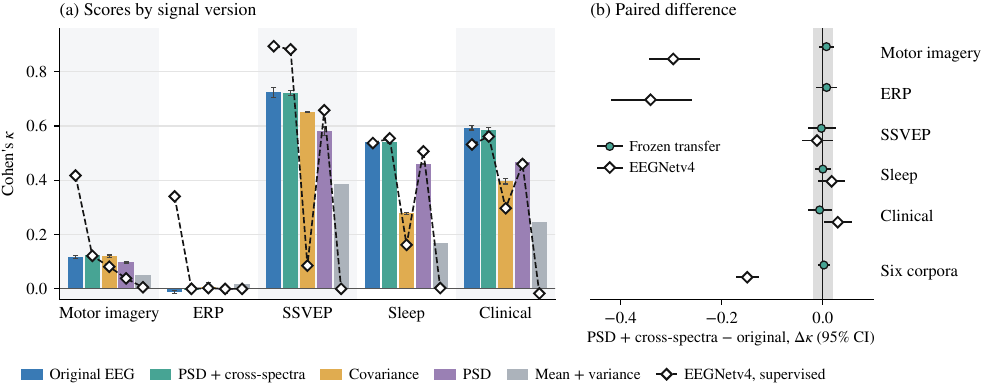}
\caption{\textbf{Preserving pretraining transfer does not preserve downstream decoding.} \textbf{(a)} REVE-small MAE encoders pretrained on each signal version, frozen and evaluated on original EEG (bars); EEGNetv4 trained from scratch and tested on the corresponding signal version (diamonds). Bars are mean $\pm$ SD over three pretraining seeds (Mean + variance: one run). \textbf{(b)} Paired difference between PSD + cross-spectra and original EEG, for frozen transfer (circles) and EEGNetv4 (diamonds), with 95\% intervals from a bootstrap over test subjects and seeds. The grey band is the $\pm 0.02~\kappa$ equivalence margin, 10\% of the mean gain from pretraining. On ERP both transfer scores are near zero, so agreement there reflects a shared failure, not successful transfer. All configurations: Table~\ref{tab:paired_configs}.}
\label{fig:synthetic_transfer}
\vspace{-10pt}
\end{figure}

The results show that REVE-small pretrained on PSD + cross-spectra signals achieves reconstruction and downstream scores close to those obtained after pretraining on original EEG data: $0.648$ vs. $0.657$~$R_0^2$ and $0.349$ vs. $0.347$~$\kappa$ (Table~\ref{tab:pretraining_statistics}). Their paired difference, $+0.002$~$\kappa$ [$-0.008$, $+0.012$], is equivalent to zero within $\pm0.02$~$\kappa$, 10\% of the pretraining gain (TOST $p<0.001$; Figure~\ref{fig:synthetic_transfer}b); by family, equivalence holds on motor imagery and sleep, and the other differences ($\leq 0.007$~$\kappa$) are inconclusive (Table~\ref{tab:paired_configs}). Both pretrained encoders substantially outperform the randomly initialized encoder on average ($0.146$~$\kappa$). The similarity between the two pretraining conditions is also evident across task families (Figure~\ref{fig:synthetic_transfer}) and individual dataset–task combinations (Appendix~\ref{app:s4_corpus_results}). In both conditions, gains over random initialization are much larger on sleep staging, abnormality detection and SSVEP than on motor imagery and ERPs.

In contrast, the other transformations yield lower mean downstream performance. Pretraining on features that capture only channel means and variances gives a reconstruction score close to zero on original EEG data and downstream performance close to that of the randomly initialized encoder. PSD and Covariance both improve downstream performance relative to random initialization, but the gains are smaller than those obtained with real-EEG data pretraining. The comparison also shows that reconstruction and downstream performance do not follow the same ordering. Covariance reconstructs much better than PSD ($0.570$ vs. $0.053$~$R_0^2$), but leads to lower downstream performance ($0.262$ vs. $0.283$~$\kappa$). These findings indicate that better reconstruction does not necessarily correspond to better downstream performance. \looseness -1

\subsection{The result extends across objectives and training configurations}
\label{sec:surrogates_scope}

\begin{table}[b]
\centering
\footnotesize
\setlength{\tabcolsep}{3pt}
\caption{\textbf{Real and PSD + cross-spectra pretraining yield similar transfer across objectives and encoder sizes.} Frozen transfer to six real-EEG corpora, weighted equally, with attention pooling ($\kappa$). REVE-small entries are mean $\pm$ SD over three pretraining seeds; REVE-base uses one run per signal. $\Delta$ is synthetic minus real, computed from unrounded scores within corresponding pretraining-run indices before aggregation.}
\label{tab:synthetic_objective_controls}
\begin{tabular}{llccc}
\toprule
\textbf{Encoder}
& \textbf{Objective}
& \textbf{Original EEG}
& \textbf{PSD + cross-spectra}
& $\boldsymbol{\Delta}$ \\
\midrule
REVE-small
& MAE
& $0.347 \pm 0.005$
& $0.349 \pm 0.005$
& $+0.002 \pm 0.008$ \\
& JEPA
& $0.354 \pm 0.002$
& $0.349 \pm 0.003$
& $-0.005 \pm 0.005$ \\
& JEPA + recon.
& $0.352 \pm 0.004$
& $0.349 \pm 0.003$
& $-0.003 \pm 0.004$ \\
\midrule
REVE-base
& MAE
& $0.368$
& $0.364$
& $-0.003$ \\
\bottomrule
\end{tabular}
\end{table}

Our results demonstrate that models pretrained on original EEG or PSD + cross-spectra signals achieve similar mean downstream performance under all three objectives (Table~\ref{tab:synthetic_objective_controls}). The similarity observed with masked reconstruction therefore also extends to the evaluated latent-prediction JEPA objectives. The downstream scores remain close across these comparisons, regardless of pooling strategies, encoder sizes and training budgets. At the larger budget, attention-pooling transfer reaches $0.371$ for original EEG data and $0.368$ for PSD + cross-spectra. The baselines trained with PSD and Covariance yield lower mean transfer throughout the objective and encoder-size comparisons (Appendices~\ref{app:s4_objective_controls}--\ref{app:s4_budget}). The same holds beyond REVE: CBraMod and LUNA (masked reconstruction) and REVE-small (contrastive) transfer as well after PSD + cross-spectra pretraining as after original-EEG pretraining (within $0.015$~$\kappa$ over 27 corpora), and the gains of the 23 published encoders of Section~\ref{sec:results} persist, on average in every task family, when they read PSD + cross-spectra versions of the downstream windows (Appendix~\ref{app:beyond_reve}).

To validate our pretraining setup, we compared our own pretraining of REVE-base against the official pretrained weights under the same downstream evaluation. Despite differences in pretraining data and compute, our model matches the published one ($\kappa = 0.368$ for ours vs. $0.347$ for the official weights; Appendix~\ref{app:s4_objective_controls}, Table~\ref{tab:s4_official_references}), confirming that our setup is a reasonable basis for studying the effects of pretraining. These findings show that the original EEG waveforms are not required in the pretraining data to obtain similar performance under the evaluated configurations. However, they do not establish that the encoders rely exclusively on the retained statistics.

\subsection{The same transformation impairs motor-imagery and ERP decoding}
\label{sec:surrogates_tasks}

The supervised experiment shows that the effect of the same transformation depends strongly on the downstream task. PSD + cross-spectra leaves EEGNetv4 performance largely unchanged on sleep staging, abnormality detection and SSVEP decoding (Figure~\ref{fig:synthetic_transfer}). In contrast, ERP~\cite{lee2019openbmi} decoding falls to chance, from $0.341$ to $0$~$\kappa$, and motor-imagery decoding drops from $0.440$ to $0.168$ on BCI~IV-2a and from $0.395$ to $0.075$ on PhysioNet~MI (Appendix~\ref{app:s4_corpus_results}). For ERPs, similar real--synthetic transfer thus reflects a shared limitation: neither pretraining condition improves over random initialization, although the real signal supports supervised decoding. The agreement between these conditions therefore does not indicate successful transfer to ERP decoding.  Together, these results support the alignment hypothesis in Section~\ref{sec:introduction}: the transformation preserves the observed pretraining benefit while disrupting structure useful for decoding the tasks that pretraining helps least. Note that these findings do not establish complete information loss. %

\section{Reconstruction does not retain most of the supervised gain}
\label{sec:retention}

The previous analyses leave open whether the limited transfer to motor-imagery and ERP tasks is due to the clinical pretraining data, the encoder of choice, or the training objective. To separate these factors, we pretrain the same encoder architecture directly on the recordings of each downstream task, adding a supervised loss on the task labels to its usual reconstruction objective. This first stage shows what the labels contribute when data and architecture are held fixed. We then drop the label loss and continue training on the same windows with reconstruction alone, as in label-free pretraining, to test whether the encoder retains the decoding gains that supervision had produced. This lets us examine both the acquisition and the retention of representations useful for decoding.

\textbf{Experimental setup.}
We train REVE-small masked autoencoders on four tasks, treating each corpus separately: motor imagery (MI)~\citep{stieger2021}, event-related potentials (ERP)~\citep{lee2019openbmi}, sleep staging~\citep{khalighi2016isruc}, and abnormality detection~\citep{obeid2016tuh}. For each data set and  seed, we initialize two models with the same random weights and train them on the same windows. One minimizes the masked-patch L1 loss $L_{\mathrm{recon}}$; the other minimizes $L_{\mathrm{recon}}+\lambda L_{\mathrm{label}}$, with $\lambda=0.02$. The label loss is the cross-entropy of a linear classifier applied to the mean-pooled encoder features of the complete, unmasked window.

After 4.096 million window samples (the \emph{split}), we copy the jointly trained model, including its optimizer state, into two branches. One branch keeps both losses; the other drops the label loss and continues with reconstruction alone. Both branches then train on another 8.192 million samples, seen in the same order and with the same masks. The reconstruction-only reference also keeps training, up to the same total of 12.288 million samples. Each dataset has about 24,000 training windows, so windows are seen many times. These windows are cue-aligned trial epochs; generic windows cut from the continuous recordings, or up to 7.5 times more aligned epochs, barely change decoding (Appendix~\ref{app:s5_task_epochs}). All runs use the same number of updates and samples; supervision only adds computation, because the label loss needs an extra pass on the complete, unmasked window (Appendices~\ref{app:s5_data}--\ref{app:s5_training}).

Encoder training, classification-head development, and testing use disjoint subjects. At every checkpoint, we freeze the encoder and fit a fresh attention-pooling readout, independent of the classifier used during encoder training. The head takes either the encoder features $Z$ alone or their concatenation with $H_d$. The latter forms $[H_{512},Z]$, which contains the classical descriptors of Section~\ref{sec:results} reduced to $d=512$ principal components. For absolute performance, we pool test windows and average the resulting scores first over three head initializations and then over three encoder seeds. The gains from supervision when descriptors are available and passed to both heads are calculated by averaging within-subject differences against the reconstruction-only reference at the same training budget, with identical descriptors passed to both heads (Appendices~\ref{app:s5_readouts}--\ref{app:s5_statistics}).

\begin{figure}[h]
    \centering
    \includegraphics[width=\linewidth]{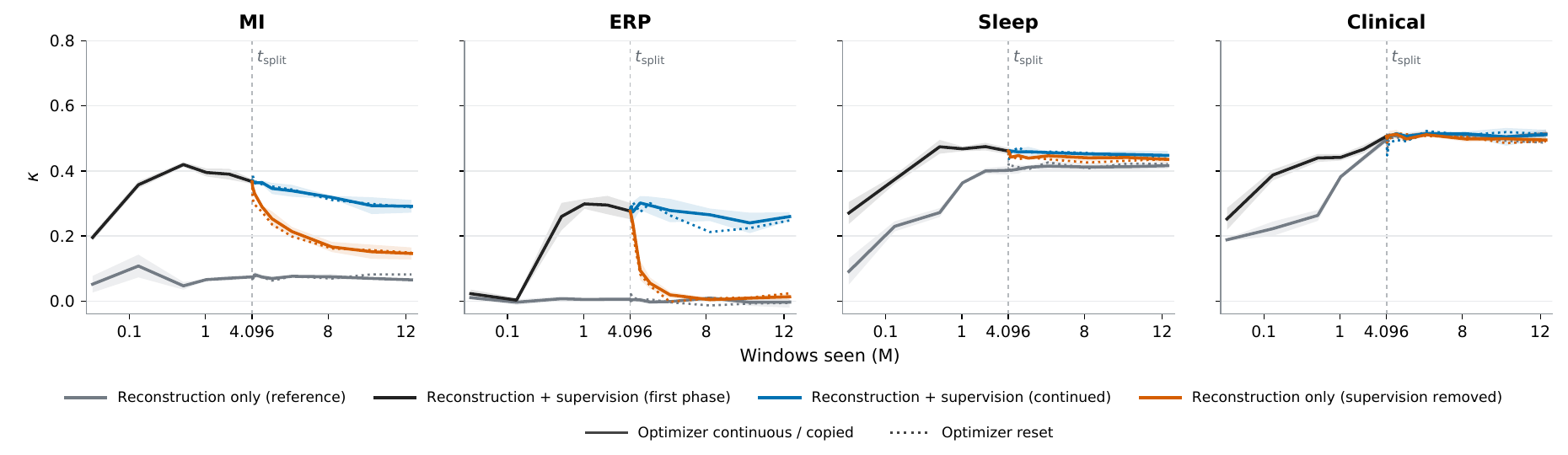}
    \caption{\textbf{Continuing with reconstruction alone loses most of the decoding gain on motor imagery and ERPs.} Test $\kappa$ of an attention-pooling readout refitted at every checkpoint on frozen encoder features $Z$, on subjects unseen during encoder training. Until the split (4.096M samples), encoders are trained with reconstruction alone (grey) or with reconstruction plus label supervision (black). The jointly trained encoder is then copied and continues with (blue) or without (orange) the label loss, on the same windows, in the same order, with the same masks. Dotted lines rerun both continuations and the reference with a fresh optimizer. Curves show mean $\pm$ SD over three seeds, each averaged over three head initializations. x-axis is logarithmic before the split and linear after. \looseness -1}
    \label{fig:retention_trajectories}
\end{figure}

\subsection{Labels reveal predictive gains that reconstruction does not obtain}
\label{sec:retention_learn}

The results on task-specific recordings show the same differences between tasks observed in the preceding experiments. Reconstruction alone produces representations useful for sleep staging and abnormality detection, but leads to much weaker decoding performance on motor imagery and ERPs (Figure~\ref{fig:retention_trajectories}, grey). At the split, leveraging label supervision increases pooled test $\kappa$ from $0.074$ to $0.367$ on MI~\citep{stieger2021} and from $0.005$ to $0.277$ on ERP. These gains remain when both heads receive $H_{512}$: paired subject-level gains are $+0.203$ and $+0.258$, respectively ($+0.159$ and $+0.223$ with the full descriptors; Appendix~\ref{app:s5_controls}). The same architecture and recordings thus support predictive gains that reconstruction-only training does not obtain. A mismatch with the clinical pretraining corpus cannot, by itself, explain this result.

We find that the benefit of label supervision is much smaller on ISRUC and TUAB, particularly when the heads also receive the classical descriptors. As these tasks provide little additional gain to assess, we focus further analyses on MI and ERP (Appendix~\ref{app:s5_main_results}).

\subsection{The decoding gain has little cost in reconstruction error}
\label{sec:retention_recon}

We next examine whether the decoding gains obtained with supervision are accompanied by an increase in reconstruction loss. We compare masked-patch L1 losses at the split, using identical masks on a fixed evaluation set of subjects excluded from encoder training. On MI and ERP, the jointly trained models’ reconstruction losses exceed those of the reconstruction-only reference by up to $0.74\%$ across the three encoder seeds. The upper bounds of all paired 95\% batch-bootstrap intervals are below $1\%$ (Table~\ref{tab:s5_recon}; Appendix~\ref{app:s5_statistics}). Substantial decoding gains can therefore coexist with almost unchanged mean reconstruction loss. Although this does not completely rule out competition between the two objectives during training, it shows that the observed decoding gains require little increase in the measured reconstruction loss.

\subsection{Reconstruction-only continuation loses most of the acquired gain}
\label{sec:retention_keep}

The decoding gains decline when training resumes with reconstruction alone. Without labels, pooled test $\kappa$ falls to $0.146$ on MI and $0.013$ on ERP, compared with $0.291$ and $0.260$ under continued supervised training (Figure~\ref{fig:retention_trajectories}). Across encoder seeds, the reconstruction-only continuation retains 22--32\% of the initial advantage over the reference on MI and 2--14\% on ERP. On MI, we find performance also declines under continued supervision, although a larger proportion of the gain is retained. \looseness -1

Across both tasks and seeds, the performance of the variant trained without supervision degrades compared to its counterpart, with paired 95\% subject-bootstrap intervals (Table~\ref{tab:s5_contrasts}). Moreover, the supervision-induced advantage with descriptors supplied to both heads also shrinks: final performance gains drop to $+0.028$ on MI and $+0.002$ on ERP without labels, compared to $+0.150$ and $+0.205$ with labels (Table~\ref{tab:s5_conditional}). Since every checkpoint receives a newly fitted head, this decline is not simply a mismatch with an outdated classifier. It measures a loss of predictive value accessible to the classification head, not necessarily the disappearance of all task information.

At the endpoint, both continuations still satisfy the reconstruction criterion on MI and ERP: their loss excesses over the reference have upper interval bounds below $1\%$. Losing most of the decoding gain is thus not necessary to remain within this tolerance. The decline also occurs with a fresh optimizer and a fivefold larger label weight (one encoder seed per control; Appendix~\ref{app:s5_controls}). Restricting the labels available to the fresh readouts preserves the retention gap without changing encoder training (Appendix~\ref{app:low_label_retention}). This mirrors the dissociation reported by \citet{zuna11}, whose downstream accuracy fell during training, on some datasets to near chance, while its reconstruction kept improving. CBraMod (motor imagery and ERP) and LUNA (motor imagery) show the same acquisition and the same loss (Appendix~\ref{app:beyond_reve}).

\section{Related Work}
\label{sec:related_work}

Several benchmarks compare EEG foundation models under shared protocols \citep{eegfmbench,eegbench,neuralbench,neuroatlas}. While they are extensive and useful for model selection, they focus on complete pipelines, with a very limited analysis on the contribution of pretrained weights. The closest work to ours is the one of \citet{yang2026steegformer}, who find that linear probing performs poorly and that, when data are scarce, fine-tuned models often fail to significantly outperform compact networks or classical non-neural decoders. We instead isolate the contribution of the pretrained weights. We compare each checkpoint with a randomly initialized copy of the same architecture \citep{saxe2011random,wieting2019random}, and we give the same classical descriptors to the readout in both cases. This extends the from-scratch or random-initialization baselines used in some benchmarks \citep{eegfmbench,neuroatlas}. These results implicate objective–task alignment as a limiting factor for MI and ERP decoding under the tested training setups.

The usefulness of pretraining depends in part on the alignment between the pretext and downstream tasks \citep{lee2021predicting}. Reconstruction objectives may not meet this condition. In vision, they favor high-variance components that can be irrelevant for perception \citep{balestriero2024reconstruction}. In EEG, masked-reconstruction gradients are only weakly aligned with classification gradients \citep{eegfmbench}, and both reconstruction losses and embeddings are dominated by low-frequency, aperiodic activity \citep{yu2026fame,kommineni2026aperiodic}. These studies diagnose the problem with the objective; we go further and intervene on pretraining, either by changing the signal the model sees or by adding label supervision. Prior work has used synthetic signals as EEG pretraining data \citep{grieger2025frequency,gnassounou2025generic}; we instead use transformed recordings as a probe, in the spirit of surrogate-data testing \citep{theiler1992surrogate,prichard1994multivariate}. Lastly, \citet{zuna11} observe that reconstruction quality and downstream performance can diverge during training. Our continuation experiments test this with matched data exposure, reconstruction masks, and optimization steps, which lets us separate whether downstream gains are acquired or retained.

\section{Conclusion}
\label{sec:conclusion}
In this work we perform an extensive empirical analysis of the benefits of EEG foundation models and the contribution of pretraining across datasets and tasks. Across 23 published encoders, pretraining gains are substantial for sleep staging and clinical EEG, small or absent for motor imagery and ERPs, and nearly vanish once both readouts also receive classical descriptors. Our findings show that pretraining on transformed EEG that preserves spectra and cross-spectra can almost reproduce the performance obtained by pretraining on raw EEG data, while the same transformation impairs supervised motor imagery and ERP decoding. We find that supervised objective functions enable the same encoders to learn these tasks, but continued reconstruction alone erodes most of the gains. These results identify the pretraining objective as a key bottleneck: It does not ensure that an encoder learns or retains representations useful across EEG tasks. As such, our work focuses on diagnosing these limitations -- identifying and validating self-supervised objectives that recover and retain these predictive gains is left to future work. These conclusions concern the four objective families we tested (masked reconstruction, latent and discrete-token prediction, contrastive learning) and scales up to 69M parameters, 23{,}000 hours of distinct data and 43 passes. Within them, nothing closed the gap to supervised decoding on motor imagery and ERPs, not even pretraining on the target recordings themselves (Appendices~\ref{app:scaling} and~\ref{app:s5_task_epochs}). Larger scales remain untested.
 
Our findings point to two directions for improvement. The first is an objective that spectra and cross-spectra alone cannot solve, aimed at the parts of the signal that downstream tasks use. These statistics are enough for current objectives, but not for tasks of motor imagery and ERPs (Section~\ref{sec:surrogates}). The second direction considers leveraging weak supervision. Most task recordings include event markers, and a small label term can be sufficient to better train an encoder at almost no cost to reconstruction (Section~\ref{sec:retention}). In both cases, a candidate objective should be evaluated as we evaluated pretraining here: by its gain over a random copy, by tasks family or against a supervised baseline. \looseness -1

\newpage

\subsection*{AI Use Statement}
The scientific ideas, methods, experimental design, analyses, and conclusions originate from the authors. Generative AI tools were used solely for assistance with translation and wording, preparation of some plots from the authors’ results, and organization of existing code for release. They were not used to develop or implement the experimental methods or to conduct experiments.

\subsection*{Ethics Statement}
This study uses existing EEG datasets made available to the research community and cited in the paper. No new participants were recruited and no new recordings were collected for this work. Our experiments investigate representation learning using previously collected data. The reported results do not establish the clinical safety or effectiveness of the evaluated models and should not be interpreted as validating them for clinical deployment.

\subsection*{Reproducibility Statement}
Our code will be made publicly available to support reproduction of the reported experiments. The experimental protocols are described in the paper and its appendices, including data preprocessing, subject-disjoint splits, encoder configurations, descriptor construction, training and evaluation procedures, surrogate generation, and continuation experiments.

\bibliography{references,refs_revision}

@inproceedings{cbramod,
  title     = {{CBraMod}: A Criss-Cross Brain Foundation Model for {EEG} Decoding},
  author    = {Wang, Jiquan and Zhao, Sha and Luo, Zhiling and Zhou, Yangxuan and Jiang, Haiteng and Li, Shijian and Li, Tao and Pan, Gang},
  booktitle = {International Conference on Learning Representations},
  year      = {2025},
  url       = {https://arxiv.org/abs/2412.07236}
}

@misc{reve,
  title         = {{REVE}: A Foundation Model for {EEG} -- Adapting to Any Setup with Large-Scale Pretraining on 25,000 Subjects},
  author        = {El Ouahidi, Yassine and Lys, Jonathan and Th{\"o}lke, Philipp and Farrugia, Nicolas and Pasdeloup, Bastien and Gripon, Vincent and Jerbi, Karim and Lioi, Giulia},
  year          = {2025},
  eprint        = {2510.21585},
  archivePrefix = {arXiv},
  url           = {https://arxiv.org/abs/2510.21585},
  howpublished  = {arXiv preprint arXiv:2510.21585}
}

@misc{labram,
  title         = {Large Brain Model for Learning Generic Representations with Tremendous {EEG} Data in {BCI}},
  author        = {Jiang, Wei-Bang and Zhao, Li-Ming and Lu, Bao-Liang},
  year          = {2024},
  eprint        = {2405.18765},
  archivePrefix = {arXiv},
  url           = {https://arxiv.org/abs/2405.18765},
  howpublished  = {arXiv preprint arXiv:2405.18765}
}

@misc{bendr,
  title         = {{BENDR}: Using Transformers and a Contrastive Self-Supervised Learning Task to Learn from Massive Amounts of {EEG} Data},
  author        = {Kostas, Demetres and Aroca-Ouellette, Stephane and Rudzicz, Frank},
  year          = {2021},
  eprint        = {2101.12037},
  archivePrefix = {arXiv},
  url           = {https://arxiv.org/abs/2101.12037},
  howpublished  = {arXiv preprint arXiv:2101.12037}
}

@inproceedings{eegpt,
  title     = {{EEGPT}: Pretrained Transformer for Universal and Reliable Representation of {EEG} Signals},
  author    = {Wang, Guangyu and Liu, Wenchao and He, Yuhong and Xu, Cong and Ma, Lin and Li, Haifeng},
  booktitle = {Advances in Neural Information Processing Systems},
  volume    = {37},
  year      = {2024},
  url       = {https://proceedings.neurips.cc/paper_files/paper/2024/hash/4540d267eeec4e5dbd9dae9448f0b739-Abstract-Conference.html}
}

@misc{eegfmbench,
  title         = {{EEG-FM-Bench}: A Comprehensive Benchmark for the Systematic Evaluation and Diagnostic Analyses of {EEG} Foundation Models},
  author        = {Xiong, Wei and Li, Jiangtong and Li, Jie and Zhu, Kun and Jiang, Changjun},
  year          = {2026},
  eprint        = {2508.17742v3},
  archivePrefix = {arXiv},
  note          = {Version 3, revised 15 June 2026; first posted in 2025},
  url           = {https://arxiv.org/abs/2508.17742v3},
  howpublished  = {arXiv preprint arXiv:2508.17742v3}
}

@misc{eegbench,
  title         = {{EEG-Bench}: A Benchmark for {EEG} Foundation Models in Clinical Applications},
  author        = {Kastrati, Ard and B{\"u}rki, Josua and Lauer, Jonas and Xuan, Cheng and Iaquinto, Raffaele and Wattenhofer, Roger},
  year          = {2025},
  eprint        = {2512.08959},
  archivePrefix = {arXiv},
  url           = {https://arxiv.org/abs/2512.08959},
  howpublished  = {arXiv preprint arXiv:2512.08959}
}

@article{ang2012fbcsp,
  title   = {Filter Bank Common Spatial Pattern Algorithm on {BCI} Competition {IV} Datasets 2a and 2b},
  author  = {Ang, Kai Keng and Chin, Zheng Yang and Wang, Chuanchu and Guan, Cuntai and Zhang, Haihong},
  journal = {Frontiers in Neuroscience},
  volume  = {6},
  pages   = {39},
  year    = {2012},
  doi     = {10.3389/fnins.2012.00039},
  url     = {https://www.frontiersin.org/journals/neuroscience/articles/10.3389/fnins.2012.00039/full}
}

@article{barachant2012riemannian,
  title   = {Multiclass Brain-Computer Interface Classification by {Riemannian} Geometry},
  author  = {Barachant, Alexandre and Bonnet, St{\'e}phane and Congedo, Marco and Jutten, Christian},
  journal = {IEEE Transactions on Biomedical Engineering},
  volume  = {59},
  number  = {4},
  pages   = {920--928},
  year    = {2012},
  doi     = {10.1109/TBME.2011.2172210},
  url     = {https://hal.science/hal-00681328}
}

@inproceedings{yang2026steegformer,
  title     = {Are {EEG} Foundation Models Worth It? Comparative Evaluation with Traditional Decoders in Diverse {BCI} Tasks},
  author    = {Yang, Liuyin and Sun, Qiang and Li, Ang and Van Hulle, Marc M.},
  booktitle = {International Conference on Learning Representations},
  year      = {2026},
  url       = {https://proceedings.iclr.cc/paper_files/paper/2026/hash/f0f39b7686634fc81ca0112566b2c05f-Abstract-Conference.html}
}

@misc{luna,
  title         = {{LUNA}: Efficient and Topology-Agnostic Foundation Model for {EEG} Signal Analysis},
  author        = {D{\"o}ner, Berkay and Ingolfsson, Thorir Mar and Benini, Luca and Li, Yawei},
  year          = {2025},
  eprint        = {2510.22257},
  archivePrefix = {arXiv},
  url           = {https://arxiv.org/abs/2510.22257},
  howpublished  = {arXiv preprint arXiv:2510.22257}
}

@misc{csbrain,
  title         = {{CSBrain}: A Cross-scale Spatiotemporal Brain Foundation Model for {EEG} Decoding},
  author        = {Zhou, Yuchen and Wu, Jiamin and Ren, Zichen and Yao, Zhouheng and Lu, Weiheng and Peng, Kunyu and Zheng, Qihao and Song, Chunfeng and Ouyang, Wanli and Gou, Chao},
  year          = {2025},
  eprint        = {2506.23075},
  archivePrefix = {arXiv},
  url           = {https://arxiv.org/abs/2506.23075},
  howpublished  = {arXiv preprint arXiv:2506.23075}
}

@misc{femba,
  title         = {{FEMBA}: Efficient and Scalable {EEG} Analysis with a Bidirectional Mamba Foundation Model},
  author        = {Tegon, Anna and Ingolfsson, Thorir Mar and Wang, Xiaying and Benini, Luca and Li, Yawei},
  year          = {2025},
  eprint        = {2502.06438},
  archivePrefix = {arXiv},
  url           = {https://arxiv.org/abs/2502.06438},
  howpublished  = {arXiv preprint arXiv:2502.06438}
}

@inproceedings{biot,
  title     = {{BIOT}: Biosignal Transformer for Cross-data Learning in the Wild},
  author    = {Yang, Chaoqi and Westover, M. Brandon and Sun, Jimeng},
  booktitle = {Advances in Neural Information Processing Systems},
  volume    = {36},
  year      = {2023},
  doi       = {10.52202/075280-3420},
  url       = {https://proceedings.neurips.cc/paper_files/paper/2023/hash/f6b30f3e2dd9cb53bbf2024402d02295-Abstract.html}
}

@inproceedings{neurogpt,
  title         = {{Neuro-GPT: Towards A Foundation Model For EEG}},
  author        = {Cui, Wenhui and Jeong, Woojae and Th{\"{o}}lke, Philipp and Medani, Takfarinas and Jerbi, Karim and Joshi, Anand A. and Leahy, Richard M.},
  booktitle     = {2024 IEEE International Symposium on Biomedical Imaging (ISBI)},
  pages         = {1--5},
  year          = {2024},
  doi           = {10.1109/isbi56570.2024.10635453},
  url           = {https://doi.org/10.1109/isbi56570.2024.10635453}
}

@inproceedings{neurolm,
  title         = {{{NeuroLM}: A Universal Multi-task Foundation Model for Bridging the Gap between Language and {EEG} Signals}},
  author        = {Jiang, Wei-Bang and Wang, Yansen and Lu, Bao-Liang and Li, Dongsheng},
  booktitle     = {International Conference on Learning Representations},
  year          = {2025},
  url           = {https://proceedings.iclr.cc/paper_files/paper/2025/hash/8b4add8b0aa8749d80a34ca5d941c355-Abstract-Conference.html}
}

@misc{neurorvq,
  title         = {{NeuroRVQ}: Multi-Scale {EEG} Tokenization for Generative Large Brainwave Models},
  author        = {Barmpas, Konstantinos and Lee, Na and Koliousis, Alexandros and Panagakis, Yannis and Adamos, Dimitrios A. and Laskaris, Nikolaos and Zafeiriou, Stefanos},
  note          = {Version 1},
  year          = {2025},
  eprint        = {2510.13068v1},
  archivePrefix = {arXiv},
  url           = {https://arxiv.org/abs/2510.13068v1},
  howpublished  = {arXiv preprint arXiv:2510.13068v1}
}

@misc{codebrain,
  title         = {{CodeBrain}: Bridging Decoupled Tokenizer and Multi-Scale Architecture for {EEG} Foundation Model},
  author        = {Ma, Jingying and Wu, Feng and Lin, Qika and Xing, Yucheng and Liu, Chenyu and Jia, Ziyu and Feng, Mengling},
  year          = {2025},
  eprint        = {2506.09110},
  archivePrefix = {arXiv},
  url           = {https://arxiv.org/abs/2506.09110},
  howpublished  = {arXiv preprint arXiv:2506.09110}
}

@misc{brainomni,
  title         = {{BrainOmni}: A Brain Foundation Model for Unified {EEG} and {MEG} Signals},
  author        = {Xiao, Qinfan and Cui, Ziyun and Zhang, Chi and Chen, Siqi and Wu, Wen and Thwaites, Andrew and Woolgar, Alexandra and Zhou, Bowen and Zhang, Chao},
  year          = {2025},
  eprint        = {2505.18185},
  archivePrefix = {arXiv},
  url           = {https://arxiv.org/abs/2505.18185},
  howpublished  = {arXiv preprint arXiv:2505.18185}
}

@misc{signaljepa,
  title         = {{S-JEPA}: Towards Seamless Cross-Dataset Transfer Through Dynamic Spatial Attention},
  author        = {Guetschel, Pierre and Moreau, Thomas and Tangermann, Michael},
  year          = {2024},
  eprint        = {2403.11772},
  archivePrefix = {arXiv},
  url           = {https://arxiv.org/abs/2403.11772},
  howpublished  = {arXiv preprint arXiv:2403.11772}
}

@misc{osf,
  title         = {{OSF}: On Pre-training and Scaling of Sleep Foundation Models},
  author        = {Shuai, Zitao and Xu, Zongzhe and Yang, David and Wang, Wei and Yang, Yuzhe},
  year          = {2026},
  eprint        = {2603.00190},
  archivePrefix = {arXiv},
  url           = {https://arxiv.org/abs/2603.00190},
  howpublished  = {arXiv preprint arXiv:2603.00190}
}

@misc{lead,
  title         = {{LEAD}: An {EEG} Foundation Model for Alzheimer's Disease Detection},
  author        = {Wang, Yihe and Huang, Nan and Mammone, Nadia and Cecchi, Marco and Zhang, Xiang},
  year          = {2025},
  eprint        = {2502.01678},
  archivePrefix = {arXiv},
  url           = {https://arxiv.org/abs/2502.01678},
  howpublished  = {arXiv preprint arXiv:2502.01678}
}

@article{eegnet,
  title   = {{EEGNet}: A Compact Convolutional Neural Network for {EEG}-Based Brain--Computer Interfaces},
  author  = {Lawhern, Vernon J. and Solon, Amelia J. and Waytowich, Nicholas R. and Gordon, Stephen M. and Hung, Chou P. and Lance, Brent J.},
  journal = {Journal of Neural Engineering},
  volume  = {15},
  number  = {5},
  pages   = {056013},
  year    = {2018},
  doi     = {10.1088/1741-2552/aace8c},
  url     = {https://doi.org/10.1088/1741-2552/aace8c}
}

@article{welch1967,
  title   = {The Use of Fast Fourier Transform for the Estimation of Power Spectra: A Method Based on Time Averaging over Short, Modified Periodograms},
  author  = {Welch, Peter D.},
  journal = {IEEE Transactions on Audio and Electroacoustics},
  volume  = {15},
  number  = {2},
  pages   = {70--73},
  year    = {1967},
  doi     = {10.1109/TAU.1967.1161901},
  url     = {https://doi.org/10.1109/TAU.1967.1161901}
}

@article{prichard1994multivariate,
  title   = {Generating Surrogate Data for Time Series with Several Simultaneously Measured Variables},
  author  = {Prichard, Dean and Theiler, James},
  journal = {Physical Review Letters},
  volume  = {73},
  number  = {7},
  pages   = {951--954},
  year    = {1994},
  doi     = {10.1103/PhysRevLett.73.951},
  url     = {https://doi.org/10.1103/PhysRevLett.73.951}
}

@article{theiler1992surrogate,
  title   = {Testing for Nonlinearity in Time Series: The Method of Surrogate Data},
  author  = {Theiler, James and Eubank, Stephen and Longtin, Andr{\'e} and Galdrikian, Bryan and Farmer, J. Doyne},
  journal = {Physica D: Nonlinear Phenomena},
  volume  = {58},
  number  = {1--4},
  pages   = {77--94},
  year    = {1992},
  doi     = {10.1016/0167-2789(92)90102-S},
  url     = {https://doi.org/10.1016/0167-2789(92)90102-S}
}

@article{eegmamba,
  title   = {{EEGMamba}: An {EEG} Foundation Model with Mamba},
  author  = {Wang, Jiquan and Zhao, Sha and Luo, Zhiling and Zhou, Yangxuan and Li, Shijian and Pan, Gang},
  journal = {Neural Networks},
  volume  = {192},
  pages   = {107816},
  year    = {2025},
  doi     = {10.1016/j.neunet.2025.107816},
  url     = {https://doi.org/10.1016/j.neunet.2025.107816}
}

@inproceedings{eegdino,
  title     = {{EEG-DINO}: Learning {EEG} Foundation Models via Hierarchical Self-Distillation},
  author    = {Wang, Xujia and Liu, Xuhui and Liu, Xi and Si, Qian and Xu, Zhaoliang and Li, Yang and Zhen, Xiantong},
  booktitle = {Medical Image Computing and Computer Assisted Intervention -- MICCAI 2025},
  series    = {Lecture Notes in Computer Science},
  volume    = {15960},
  pages     = {196--205},
  year      = {2025},
  publisher = {Springer Nature Switzerland},
  doi       = {10.1007/978-3-032-04927-8_19},
  url       = {https://doi.org/10.1007/978-3-032-04927-8_19}
}

@article{stieger2021,
  title   = {Continuous Sensorimotor Rhythm Based Brain Computer Interface Learning in a Large Population},
  author  = {Stieger, James R. and Engel, Stephen A. and He, Bin},
  journal = {Scientific Data},
  volume  = {8},
  pages   = {98},
  year    = {2021},
  doi     = {10.1038/s41597-021-00883-1},
  url     = {https://doi.org/10.1038/s41597-021-00883-1}
}

@article{tangermann2012bcicomp,
  author  = {Tangermann, Michael and M{\"u}ller, Klaus-Robert and
             Aertsen, Ad and Birbaumer, Niels and Braun, Christoph and
             Brunner, Clemens and Leeb, Robert and Mehring, Carsten and
             Miller, Kai J. and M{\"u}ller-Putz, Gernot R. and Nolte, Guido and
             Pfurtscheller, Gert and Preissl, Hubert and Schalk, Gerwin and
             Schl{\"o}gl, Alois and Vidaurre, Carmen and Waldert, Stephan and
             Blankertz, Benjamin},
  title   = {Review of the {BCI} Competition {IV}},
  journal = {Frontiers in Neuroscience},
  year    = {2012},
  volume  = {6},
  pages   = {55},
  doi     = {10.3389/fnins.2012.00055},
  url     = {https://doi.org/10.3389/fnins.2012.00055}
}

@article{cho2017mi,
  author  = {Cho, Hohyun and Ahn, Minkyu and Ahn, Sangtae and
             Kwon, Moonyoung and Jun, Sung Chan},
  title   = {{EEG} datasets for motor imagery brain--computer interface},
  journal = {GigaScience},
  year    = {2017},
  volume  = {6},
  number  = {7},
  pages   = {gix034},
  doi     = {10.1093/gigascience/gix034},
  url     = {https://doi.org/10.1093/gigascience/gix034}
}

@article{dreyer2023mi,
  author  = {Dreyer, Pauline and Roc, Aline and Pillette, L{\'e}a and
             Rimbert, S{\'e}bastien and Lotte, Fabien},
  title   = {A large {EEG} database with users' profile information for
             motor imagery brain-computer interface research},
  journal = {Scientific Data},
  year    = {2023},
  volume  = {10},
  number  = {1},
  pages   = {580},
  doi     = {10.1038/s41597-023-02445-z},
  url     = {https://doi.org/10.1038/s41597-023-02445-z}
}

@article{lee2019openbmi,
  author  = {Lee, Min-Ho and Kwon, O-Yeon and Kim, Yong-Jeong and
             Kim, Hong-Kyung and Lee, Young-Eun and Williamson, John and
             Fazli, Siamac and Lee, Seong-Whan},
  title   = {{EEG} dataset and {OpenBMI} toolbox for three {BCI} paradigms:
             an investigation into {BCI} illiteracy},
  journal = {GigaScience},
  year    = {2019},
  volume  = {8},
  number  = {5},
  pages   = {giz002},
  doi     = {10.1093/gigascience/giz002},
  url     = {https://doi.org/10.1093/gigascience/giz002}
}

@article{physionet_eegmmidb,
  author  = {Schalk, Gerwin},
  title   = {{EEG} Motor Movement/Imagery Dataset},
  journal = {PhysioNet},
  year    = {2009},
  month   = sep,
  note    = {Version 1.0.0},
  doi     = {10.13026/C28G6P},
  url     = {https://doi.org/10.13026/C28G6P}
}

@article{goldberger2000physionet,
  author  = {Goldberger, Ary L. and Amaral, Luis A. N. and Glass, Leon and
             Hausdorff, Jeffrey M. and Ivanov, Plamen Ch. and Mark, Roger G. and
             Mietus, Joseph E. and Moody, George B. and Peng, Chung-Kang and
             Stanley, H. Eugene},
  title   = {{PhysioBank}, {PhysioToolkit}, and {PhysioNet}: Components of a
             New Research Resource for Complex Physiologic Signals},
  journal = {Circulation},
  year    = {2000},
  volume  = {101},
  number  = {23},
  pages   = {e215--e220},
  doi     = {10.1161/01.CIR.101.23.e215},
  url     = {https://doi.org/10.1161/01.CIR.101.23.e215}
}

@article{schirrmeister2017deep,
  author  = {Schirrmeister, Robin Tibor and Springenberg, Jost Tobias and
             Fiederer, Lukas Dominique Josef and Glasstetter, Martin and
             Eggensperger, Katharina and Tangermann, Michael and
             Hutter, Frank and Burgard, Wolfram and Ball, Tonio},
  title   = {Deep learning with convolutional neural networks for {EEG}
             decoding and visualization},
  journal = {Human Brain Mapping},
  year    = {2017},
  volume  = {38},
  number  = {11},
  pages   = {5391--5420},
  doi     = {10.1002/hbm.23730},
  url     = {https://doi.org/10.1002/hbm.23730}
}

@article{kappenman2021erpcore,
  author  = {Kappenman, Emily S. and Farrens, Jaclyn L. and Zhang, Wendy and
             Stewart, Andrew X. and Luck, Steven J.},
  title   = {{ERP CORE}: An open resource for human event-related potential
             research},
  journal = {NeuroImage},
  year    = {2021},
  volume  = {225},
  pages   = {117465},
  doi     = {10.1016/j.neuroimage.2020.117465},
  url     = {https://doi.org/10.1016/j.neuroimage.2020.117465}
}

@article{liu2020beta,
  author  = {Liu, Bingchuan and Huang, Xiaoshan and Wang, Yijun and
             Chen, Xiaogang and Gao, Xiaorong},
  title   = {{BETA}: A Large Benchmark Database Toward {SSVEP-BCI} Application},
  journal = {Frontiers in Neuroscience},
  year    = {2020},
  volume  = {14},
  pages   = {627},
  doi     = {10.3389/fnins.2020.00627},
  url     = {https://doi.org/10.3389/fnins.2020.00627}
}

@article{kalunga2016ssvep,
  author  = {Kalunga, Emmanuel K. and Chevallier, Sylvain and
             Barth{\'e}lemy, Quentin and Djouani, Karim and
             Monacelli, Eric and Hamam, Yskandar},
  title   = {Online {SSVEP}-based {BCI} using Riemannian geometry},
  journal = {Neurocomputing},
  year    = {2016},
  volume  = {191},
  pages   = {55--68},
  doi     = {10.1016/j.neucom.2016.01.007},
  url     = {https://doi.org/10.1016/j.neucom.2016.01.007}
}

@article{nakanishi2015ssvep,
  author  = {Nakanishi, Masaki and Wang, Yijun and Wang, Yu-Te and
             Jung, Tzyy-Ping},
  title   = {A Comparison Study of Canonical Correlation Analysis Based
             Methods for Detecting Steady-State Visual Evoked Potentials},
  journal = {PLOS ONE},
  year    = {2015},
  volume  = {10},
  number  = {10},
  pages   = {e0140703},
  doi     = {10.1371/journal.pone.0140703},
  url     = {https://doi.org/10.1371/journal.pone.0140703}
}

@article{khalighi2016isruc,
  author  = {Khalighi, Sirvan and Sousa, Teresa and
             Santos, Jos{\'e} Moutinho and Nunes, Urbano},
  title   = {{ISRUC-Sleep}: A comprehensive public dataset for sleep researchers},
  journal = {Computer Methods and Programs in Biomedicine},
  year    = {2016},
  volume  = {124},
  pages   = {180--192},
  doi     = {10.1016/j.cmpb.2015.10.013},
  url     = {https://doi.org/10.1016/j.cmpb.2015.10.013}
}

@article{kemp2000sleepedf,
  author  = {Kemp, B. and Zwinderman, A. H. and Tuk, B. and
             Kamphuisen, H. A. C. and Oberye, J. J. L.},
  title   = {Analysis of a sleep-dependent neuronal feedback loop:
             the slow-wave microcontinuity of the {EEG}},
  journal = {IEEE Transactions on Biomedical Engineering},
  year    = {2000},
  volume  = {47},
  number  = {9},
  pages   = {1185--1194},
  doi     = {10.1109/10.867928},
  url     = {https://doi.org/10.1109/10.867928}
}

@article{miltiadous2023adftd,
  author  = {Miltiadous, Andreas and Tzimourta, Katerina D. and
             Afrantou, Theodora and Ioannidis, Panagiotis and
             Grigoriadis, Nikolaos and Tsalikakis, Dimitrios G. and
             Angelidis, Pantelis and Tsipouras, Markos G. and
             Glavas, Euripidis and Giannakeas, Nikolaos and
             Tzallas, Alexandros T.},
  title   = {A Dataset of Scalp {EEG} Recordings of Alzheimer's Disease,
             Frontotemporal Dementia and Healthy Subjects from Routine {EEG}},
  journal = {Data},
  year    = {2023},
  volume  = {8},
  number  = {6},
  pages   = {95},
  doi     = {10.3390/data8060095},
  url     = {https://doi.org/10.3390/data8060095}
}

@article{obeid2016tuh,
  author  = {Obeid, Iyad and Picone, Joseph},
  title   = {The {Temple University Hospital EEG Data Corpus}},
  journal = {Frontiers in Neuroscience},
  year    = {2016},
  volume  = {10},
  pages   = {196},
  doi     = {10.3389/fnins.2016.00196},
  url     = {https://doi.org/10.3389/fnins.2016.00196}
}

@article{katsigiannis2018dreamer,
  author  = {Katsigiannis, Stamos and Ramzan, Naeem},
  title   = {{DREAMER}: A Database for Emotion Recognition Through {EEG} and
             {ECG} Signals From Wireless Low-cost Off-the-Shelf Devices},
  journal = {IEEE Journal of Biomedical and Health Informatics},
  year    = {2018},
  volume  = {22},
  number  = {1},
  pages   = {98--107},
  doi     = {10.1109/JBHI.2017.2688239},
  url     = {https://doi.org/10.1109/JBHI.2017.2688239}
}

@article{faced2023,
  author  = {Chen, Jingjing and Wang, Xiaobin and Huang, Chen and
             Hu, Xin and Shen, Xinke and Zhang, Dan},
  title   = {A Large Finer-grained Affective Computing {EEG} Dataset},
  journal = {Scientific Data},
  year    = {2023},
  volume  = {10},
  number  = {1},
  pages   = {740},
  doi     = {10.1038/s41597-023-02650-w},
  url     = {https://doi.org/10.1038/s41597-023-02650-w}
}

@article{zyma2019eegmat,
  author  = {Zyma, Igor and Tukaev, Sergii and Seleznov, Ivan and
             Kiyono, Ken and Popov, Anton and Chernykh, Mariia and
             Shpenkov, Oleksii},
  title   = {Electroencephalograms during Mental Arithmetic Task Performance},
  journal = {Data},
  year    = {2019},
  volume  = {4},
  number  = {1},
  pages   = {14},
  doi     = {10.3390/data4010014},
  url     = {https://doi.org/10.3390/data4010014}
}

@article{shin2018eegnirs,
  author  = {Shin, Jaeyoung and von L{\"u}hmann, Alexander and
             Blankertz, Benjamin and Kim, Do-Won and Jeong, Jichai and
             Hwang, Han-Jeong and M{\"u}ller, Klaus-Robert},
  title   = {Open Access Dataset for {EEG+NIRS} Single-Trial Classification},
  journal = {IEEE Transactions on Neural Systems and Rehabilitation Engineering},
  year    = {2017},
  volume  = {25},
  number  = {10},
  pages   = {1735--1745},
  doi     = {10.1109/TNSRE.2016.2628057},
  url     = {https://doi.org/10.1109/TNSRE.2016.2628057}
}

@article{zheng2017vigilance,
  author  = {Zheng, Wei-Long and Lu, Bao-Liang},
  title   = {A multimodal approach to estimating vigilance using {EEG} and
             forehead {EOG}},
  journal = {Journal of Neural Engineering},
  year    = {2017},
  volume  = {14},
  number  = {2},
  pages   = {026017},
  doi     = {10.1088/1741-2552/aa5a98},
  url     = {https://doi.org/10.1088/1741-2552/aa5a98}
}

@inproceedings{lee2021predicting,
  author    = {Lee, Jason D. and Lei, Qi and Saunshi, Nikunj
               and Zhuo, Jiacheng},
  title     = {Predicting What You Already Know Helps: Provable Self-Supervised Learning},
  booktitle = {Advances in Neural Information Processing Systems},
  volume    = {34},
  pages     = {309--323},
  year      = {2021},
  url       = {https://proceedings.nips.cc/paper_files/paper/2021/hash/02e656adee09f8394b402d9958389b7d-Abstract.html}
}

@misc{neuralbench,
  title         = {{NeuralBench: A Unifying Framework to Benchmark NeuroAI Models}},
  author        = {Banville, Hubert and d'Ascoli, St{\'{e}}phane and Dahan, Simon and Rapin, J{\'{e}}r{\'{e}}my and Careil, Marl{\`{e}}ne and Benchetrit, Yohann and L{\'{e}}vy, Jarod and Panchavati, Saarang and Ratouchniak, Antoine and Zhang, Mingfang (Lucy) and Cascardi, Elisa and Begany, Katelyn and Brooks, Teon and King, Jean-R{\'{e}}mi},
  year          = {2026},
  howpublished  = {arXiv preprint arXiv:2605.08495},
  eprint        = {2605.08495},
  archivePrefix = {arXiv},
  url           = {https://arxiv.org/abs/2605.08495}
}

@misc{neuroatlas,
  title         = {{NeuroAtlas: Benchmarking Foundation Models for Clinical EEG and Brain-Computer Interfaces}},
  author        = {Kontras, Konstantinos and Osselaer, Trui and Mouslech, Stylianos G. and Karaiskou, Angeliki-Ilektra and Gagliardi, Guido and Strypsteen, Thomas and Badiei, Mohammad Hossein and Rani, Anku and Vanmarcke, Maarten and Bhagubai, Miguel and Ekbote, Chanakya and Hwang, Jaedong and Chatzichristos, Christos and Liang, Paul Pu and De Vos, Maarten},
  year          = {2026},
  howpublished  = {arXiv preprint arXiv:2605.14698},
  eprint        = {2605.14698},
  archivePrefix = {arXiv},
  url           = {https://arxiv.org/abs/2605.14698}
}

@misc{yu2026fame,
  title         = {{Understanding and Correcting Low-Frequency Bias in EEG Foundation Model}},
  author        = {Yu, Junjie and Deng, Zihan and Zhang, Jianyu and Mu, Junrong and An, Jiahui and Ma, Wenxiao and Lu, Ziling and Wang, Yue and Zhu, Yan and Lou, Kexin and Liu, Quanying},
  year          = {2026},
  howpublished  = {arXiv preprint arXiv:2608.01898},
  eprint        = {2608.01898},
  archivePrefix = {arXiv},
  url           = {https://arxiv.org/abs/2608.01898}
}

@misc{gnassounou2025generic,
  title         = {{Leveraging Generic Time Series Foundation Models for EEG Classification}},
  author        = {Gnassounou, Th{\'{e}}o and Moakher, Yessin and Xie, Shifeng and Feofanov, Vasilii and Redko, Ievgen},
  year          = {2025},
  howpublished  = {arXiv preprint arXiv:2510.27522},
  eprint        = {2510.27522},
  archivePrefix = {arXiv},
  url           = {https://arxiv.org/abs/2510.27522}
}

@misc{zuna11,
  title         = {{ZUNA1.1: A more flexible EEG foundation model for Denoising and Super-resolution}},
  author        = {Warner, Christopher and Mago, Jonas and Huml, JR and Millidge, Beren},
  year          = {2026},
  howpublished  = {arXiv preprint arXiv:2607.27308},
  eprint        = {2607.27308},
  archivePrefix = {arXiv},
  url           = {https://arxiv.org/abs/2607.27308}
}

@article{grieger2025frequency,
  title         = {{Data-Efficient Sleep Staging with Synthetic Time Series Pretraining}},
  author        = {Grieger, Niklas and Mehrkanoon, Siamak and Bialonski, Stephan},
  journal       = {Algorithms},
  volume        = {18},
  number        = {9},
  pages         = {580},
  year          = {2025},
  doi           = {10.3390/a18090580},
  url           = {https://doi.org/10.3390/a18090580}
}

@article{arico2014p300,
  title         = {{Influence of P300 latency jitter on event related potential-based brain--computer interface performance}},
  author        = {Aric{\`{o}}, P and Aloise, F and Schettini, F and Salinari, S and Mattia, D and Cincotti, F},
  journal       = {Journal of Neural Engineering},
  volume        = {11},
  number        = {3},
  pages         = {035008},
  year          = {2014},
  doi           = {10.1088/1741-2560/11/3/035008},
  url           = {https://doi.org/10.1088/1741-2560/11/3/035008}
}

@article{guger2009p300,
  title         = {{How many people are able to control a P300-based brain--computer interface (BCI)?}},
  author        = {Guger, Christoph and Daban, Shahab and Sellers, Eric and Holzner, Clemens and Krausz, Gunther and Carabalona, Roberta and Gramatica, Furio and Edlinger, Guenter},
  journal       = {Neuroscience Letters},
  volume        = {462},
  number        = {1},
  pages         = {94--98},
  year          = {2009},
  doi           = {10.1016/j.neulet.2009.06.045},
  url           = {https://doi.org/10.1016/j.neulet.2009.06.045}
}

@article{alvarez2021hmc,
  title         = {{Inter-database validation of a deep learning approach for automatic sleep scoring}},
  author        = {Alvarez-Estevez, Diego and Rijsman, Roselyne M.},
  journal       = {PLOS ONE},
  volume        = {16},
  number        = {8},
  pages         = {e0256111},
  year          = {2021},
  doi           = {10.1371/journal.pone.0256111},
  url           = {https://doi.org/10.1371/journal.pone.0256111}
}

@article{anjum2024iowa,
  title         = {{Resting-state EEG measures cognitive impairment in Parkinson's disease}},
  author        = {Anjum, Md Fahim and Espinoza, Arturo I. and Cole, Rachel C. and Singh, Arun and May, Patrick and Uc, Ergun Y. and Dasgupta, Soura and Narayanan, Nandakumar S.},
  journal       = {npj Parkinson's Disease},
  volume        = {10},
  number        = {1},
  pages         = {6},
  year          = {2024},
  doi           = {10.1038/s41531-023-00602-0},
  url           = {https://doi.org/10.1038/s41531-023-00602-0}
}

@article{liu2022seedv,
  title         = {{Comparing Recognition Performance and Robustness of Multimodal Deep Learning Models for Multimodal Emotion Recognition}},
  author        = {Liu, Wei and Qiu, Jie-Lin and Zheng, Wei-Long and Lu, Bao-Liang},
  journal       = {IEEE Transactions on Cognitive and Developmental Systems},
  volume        = {14},
  number        = {2},
  pages         = {715--729},
  year          = {2022},
  doi           = {10.1109/tcds.2021.3071170},
  url           = {https://doi.org/10.1109/tcds.2021.3071170}
}

@article{jeong2022competition,
  title         = {{2020 International brain--computer interface competition: A review}},
  author        = {Jeong, Ji-Hoon and Cho, Jeong-Hyun and Lee, Young-Eun and Lee, Seo-Hyun and Shin, Gi-Hwan and Kweon, Young-Seok and Mill{\'{a}}n, Jos{\'{e}} del R. and M{\"{u}}ller, Klaus-Robert and Lee, Seong-Whan},
  journal       = {Frontiers in Human Neuroscience},
  volume        = {16},
  pages         = {898300},
  year          = {2022},
  doi           = {10.3389/fnhum.2022.898300},
  url           = {https://doi.org/10.3389/fnhum.2022.898300}
}

@inproceedings{saxe2011random,
  title         = {{On Random Weights and Unsupervised Feature Learning}},
  author        = {Saxe, Andrew M. and Koh, Pang Wei and Chen, Zhenghao and Bhand, Maneesh and Suresh, Bipin and Ng, Andrew Y.},
  booktitle     = {Proceedings of the 28th International Conference on Machine Learning},
  year          = {2011},
  url           = {https://ai.stanford.edu/~ang/papers/icml11-RandomWeights.pdf}
}

@inproceedings{wieting2019random,
  title         = {{No Training Required: Exploring Random Encoders for Sentence Classification}},
  author        = {Wieting, John and Kiela, Douwe},
  booktitle     = {International Conference on Learning Representations},
  year          = {2019},
  url           = {https://openreview.net/forum?id=BkgPajAcY7}
}

@inproceedings{balestriero2024reconstruction,
  title         = {{How Learning by Reconstruction Produces Uninformative Features For Perception}},
  author        = {Balestriero, Randall and LeCun, Yann},
  booktitle     = {Proceedings of the 41st International Conference on Machine Learning},
  series        = {Proceedings of Machine Learning Research},
  volume        = {235},
  pages         = {2566--2585},
  year          = {2024},
  publisher     = {PMLR},
  url           = {https://proceedings.mlr.press/v235/balestriero24b.html}
}

@inproceedings{assran2023ijepa,
  title         = {{Self-Supervised Learning From Images With a Joint-Embedding Predictive Architecture}},
  author        = {Assran, Mahmoud and Duval, Quentin and Misra, Ishan and Bojanowski, Piotr and Vincent, Pascal and Rabbat, Michael and LeCun, Yann and Ballas, Nicolas},
  booktitle     = {Proceedings of the IEEE/CVF Conference on Computer Vision and Pattern Recognition},
  year          = {2023},
  pages         = {15619--15629},
  url           = {https://openaccess.thecvf.com/content/CVPR2023/html/Assran_Self-Supervised_Learning_From_Images_With_a_Joint-Embedding_Predictive_Architecture_CVPR_2023_paper.html}
}

@misc{nikolopoulos2021mamem3,
  title         = {{MAMEM EEG SSVEP Dataset III (14 channels, 11 subjects, 5 frequencies presented simultaneously)}},
  author        = {Nikolopoulos, Spiros},
  year          = {2021},
  howpublished  = {Figshare dataset},
  note          = {Version 3},
  doi           = {10.6084/m9.figshare.3413851.v3},
  url           = {https://doi.org/10.6084/m9.figshare.3413851.v3}
}

@misc{mumtaz2016mdd,
  title         = {{MDD Patients and Healthy Controls EEG Data (New)}},
  author        = {Mumtaz, Wajid},
  year          = {2016},
  howpublished  = {Figshare dataset},
  note          = {Version 2},
  doi           = {10.6084/m9.figshare.4244171.v2},
  url           = {https://doi.org/10.6084/m9.figshare.4244171.v2}
}

@misc{guttag2010chbmit,
  title         = {{CHB-MIT Scalp EEG Database}},
  author        = {Guttag, John},
  year          = {2010},
  howpublished  = {PhysioNet dataset},
  note          = {Version 1.0.0},
  doi           = {10.13026/C2K01R},
  url           = {https://physionet.org/content/chbmit/1.0.0/}
}

@misc{kommineni2026aperiodic,
  title         = {{Aperiodic and Low-Frequency Spectral Bias in Reconstruction based EEG Foundation Models}},
  author        = {Kommineni, Aditya and Zhou, Emily and Avramidis, Kleanthis and Segaard, Simon Bock and M{\"u}nster, Jeppe Roden and Hansen, Andreas Peter Juhl and Medani, Takfarinas and Feng, Tiantian and Leahy, Richard and Narayanan, Shrikanth},
  year          = {2026},
  howpublished  = {arXiv preprint arXiv:2605.26434},
  eprint        = {2605.26434},
  archivePrefix = {arXiv},
  url           = {https://arxiv.org/abs/2605.26434}
}

@inproceedings{simclr,
  title     = {A Simple Framework for Contrastive Learning of Visual Representations},
  author    = {Chen, Ting and Kornblith, Simon and Norouzi, Mohammad and Hinton, Geoffrey},
  booktitle = {International Conference on Machine Learning},
  year      = {2020}
}

@inproceedings{seqclr,
  title     = {Contrastive Representation Learning for Electroencephalogram Classification},
  author    = {Mohsenvand, Mostafa Neo and Izadi, Mohammad Rasool and Maes, Pattie},
  booktitle = {Machine Learning for Health (ML4H)},
  year      = {2020}
}

@inproceedings{hu2022lora,
  title     = {{LoRA}: Low-Rank Adaptation of Large Language Models},
  author    = {Hu, Edward J. and Shen, Yelong and Wallis, Phillip and Allen-Zhu, Zeyuan and Li, Yuanzhi and Wang, Shean and Wang, Lu and Chen, Weizhu},
  booktitle = {International Conference on Learning Representations},
  year      = {2022}
}
\bibliographystyle{iclr2027_conference}

\clearpage
\appendix
\section{Experimental Setup for Section~\ref{sec:results}}
\label{app:datasets}

This appendix describes the encoder benchmark in Section~\ref{sec:results}. The main results use attention pooling and classical descriptors reduced by PCA. We report the other readouts and the experiments without PCA separately, so that each result is tied to the protocol that produced it.

\subsection{Data and subject splits}
\label{app:s3_data}

We evaluate 34 datasets from eight families (Table~\ref{tab:section2_corpora}). The five main families are motor imagery, ERP, SSVEP, sleep, and clinical tasks, covering 27 entries. Emotion, workload, and imagined speech provide seven additional entries. Each corpus is capped at 6{,}000 windows, sampled without replacement with seed 20260906. All representations compared within a corpus use the same retained windows and subject partitions.

The common preprocessing uses EEG channels, a 0.5--45~Hz band-pass, resampling to 200~Hz, and four-second windows. Each channel is standardized within each window by subtracting its temporal mean and dividing by its population standard deviation, with a floor of $10^{-8}$. The classical descriptors and encoder inputs therefore start from the same normalized signal.

We use the common 19-channel montage whenever all its channels are available; otherwise, we use the native montage or a predefined subset. The common montage is FP1, FP2, F7, F3, FZ, F4, F8, T7, C3, CZ, C4, T8, P7, P3, PZ, P4, P8, O1, and O2. Channel counts for the six intervention datasets are reported in Table~\ref{tab:s4_data}.

All windows and recordings from one person remain in the same partition. Datasets with at least ten subjects use one fixed split of approximately 45\% training, 15\% validation, and 40\% test subjects; otherwise, we use leave-one-subject-out evaluation (LOSO). The split seed is 20260908. For subject-level classification, splits are stratified by the subject's label; TUAB uses the majority label of each person's recordings. BNCI2014-001 and Nakanishi2015 each use nine leave-one-subject-out folds, with six training, two validation, and one test subject per fold. The full annotation budget means all training and validation labels in the retained subset, not all recordings in the original dataset.

\begin{table}[!htbp]
\centering
\small
\setlength{\tabcolsep}{4pt}
\caption{Data retained for Section~\ref{sec:results}. Counts describe the
evaluated subsets after preprocessing and subsampling.}
\label{tab:section2_corpora}
\begin{tabular}{llrrr}
\toprule
Family & Dataset / task & Windows & Subjects & Classes \\
\midrule
Motor imagery
  & BNCI2014-001 \citep{tangermann2012bcicomp}
  & 5184 & 9 & 4 \\
  & Cho2017 \citep{cho2017mi}
  & 5997 & 52 & 2 \\
  & Dreyer2023 \citep{dreyer2023mi}
  & 5978 & 87 & 2 \\
  & Lee2019 MI \citep{lee2019openbmi}
  & 5994 & 54 & 2 \\
  & PhysioNet MI \citep{physionet_eegmmidb,goldberger2000physionet}
  & 6000 & 109 & 5 \\
  & Schirrmeister2017 \citep{schirrmeister2017deep}
  & 6000 & 14 & 4 \\
  & Stieger2021 \citep{stieger2021}
  & 6000 & 62 & 4 \\
\midrule
ERP
  & BNCI2014-009 \citep{arico2014p300}
  & 6000 & 10 & 2 \\
  & BNCI2015-003 \citep{guger2009p300}
  & 6000 & 10 & 2 \\
  & ERP CORE MMN \citep{kappenman2021erpcore}
  & 6000 & 40 & 2 \\
  & ERP CORE N170 \citep{kappenman2021erpcore}
  & 6000 & 40 & 2 \\
  & ERP CORE N2pc \citep{kappenman2021erpcore}
  & 6000 & 40 & 2 \\
  & Lee2019 ERP \citep{lee2019openbmi}
  & 5994 & 54 & 2 \\
\midrule
SSVEP
  & BETA \citep{liu2020beta}
  & 5995 & 55 & 40 \\
  & Kalunga2016 \citep{kalunga2016ssvep}
  & 960 & 12 & 4 \\
  & Lee2019 SSVEP \citep{lee2019openbmi}
  & 5994 & 54 & 4 \\
  & MAMEM3 \citep{nikolopoulos2021mamem3}
  & 1375 & 11 & 5 \\
  & Nakanishi2015 \citep{nakanishi2015ssvep}
  & 1620 & 9 & 12 \\
\midrule
Sleep
  & HMC \citep{alvarez2021hmc}
  & 6000 & 151 & 5 \\
  & ISRUC \citep{khalighi2016isruc}
  & 6000 & 98 & 5 \\
  & Sleep-EDF \citep{kemp2000sleepedf}
  & 6000 & 78 & 5 \\
\midrule
Clinical
  & ADFTD \citep{miltiadous2023adftd}
  & 6000 & 88 & 3 \\
  & CHB-MIT \citep{guttag2010chbmit}
  & 5432 & 23 & 2 \\
  & Mumtaz \citep{mumtaz2016mdd}
  & 4329 & 58 & 2 \\
  & PD Iowa \citep{anjum2024iowa}
  & 5904 & 149 & 2 \\
  & TUAB \citep{obeid2016tuh}
  & 6000 & 2326 & 2 \\
  & TUEV \citep{obeid2016tuh}
  & 6000 & 370 & 6 \\
\midrule
Emotion
  & DREAMER \citep{katsigiannis2018dreamer}
  & 6000 & 23 & 2 \\
  & FACED \citep{faced2023}
  & 6000 & 123 & 9 \\
  & SEED-V \citep{liu2022seedv}
  & 6000 & 16 & 5 \\
\midrule
Workload
  & EEGMAT \citep{zyma2019eegmat}
  & 2134 & 36 & 2 \\
  & SEED-VIG \citep{zheng2017vigilance}
  & 6000 & 21 & 3 \\
  & Shin2017B \citep{shin2018eegnirs}
  & 1740 & 29 & 2 \\
\midrule
Imagined speech
  & BCIC2020-3 \citep{jeong2022competition}
  & 6000 & 15 & 5 \\
\bottomrule
\end{tabular}
\end{table}

\subsection{Frozen encoders and coverage}
\label{app:s3_models}

Table~\ref{tab:s3_models} lists the 23 encoder configurations. We use published checkpoints of REVE-base \citep{reve}, CBraMod-base \citep{cbramod}, LUNA-base \citep{luna}, EEGPT-large \citep{eegpt}, LaBraM-base \citep{labram}, CSBrain-base \citep{csbrain}, EEGMamba-base \citep{eegmamba}, FEMBA-base \citep{femba}, ST-EEGFormer-base \citep{yang2026steegformer}, BIOT-six18 \citep{biot}, BENDR-base \citep{bendr}, Neuro-GPT-base \citep{neurogpt}, NeuroLM-base \citep{neurolm}, NeuroRVQ-base \citep{neurorvq}, CodeBrain \citep{codebrain}, BrainOmni-base \citep{brainomni}, EEG-DINO-small and EEG-DINO-medium \citep{eegdino}, Signal-JEPA \citep{signaljepa}, LEAD \citep{lead}, OSF-base \citep{osf}, and ZUNA-1.1 \citep{zuna11}.
Each published checkpoint is paired with one randomly initialized instance of the same architecture. Both encoders remain frozen. This design isolates the contribution of pretraining to the learned representations: fine-tuning would modify these representations through task-specific adaptation, obscuring which properties were acquired during pretraining. Our objective is to characterize what pretraining has learned rather than to maximize downstream performance. The pipeline checks that parameter names, parameter shapes, and output-grid shapes agree between the two versions. The comparisons therefore change the encoder weights while keeping the architecture and downstream training procedure fixed.

Model-specific adapters receive the same windows, resampled with a polyphase FIR filter when required.

\begin{table}[!htbp]
\centering
\small
\setlength{\tabcolsep}{4pt}
\caption{Encoder configurations and available model--corpus pairs. $D$ is the token feature width. A, M, and E denote attention, global mean, and electrode-wise pooling. The maximum coverage is 34 corpora per configuration.}
\label{tab:s3_models}
\begin{tabular}{llrrrr}
\toprule
Encoder & Configuration & $D$ & A & M & E \\
\midrule
REVE & base & 512 & 32 & 32 & 32 \\
CBraMod & base & 200 & 34 & 34 & 34 \\
LUNA & base & 256 & 32 & 32 & 0 \\
EEGPT & large & 512 & 34 & 34 & 0 \\
LaBraM & base & 200 & 31 & 31 & 31 \\
CSBrain & base & 200 & 34 & 34 & 34 \\
EEGMamba & base & 200 & 34 & 34 & 34 \\
FEMBA & base & 385 & 21 & 21 & 0 \\
ST-EEGFormer & base & 768 & 5 & 31 & 31 \\
ZUNA & 1.1 output & 32 & 32 & 32 & 32 \\
BIOT & six18 & 256 & 26 & 26 & 0 \\
BENDR & base & 512 & 32 & 32 & 0 \\
Neuro-GPT & base & 1024 & 32 & 32 & 0 \\
NeuroLM & base & 768 & 31 & 31 & 31 \\
NeuroRVQ & base & 800 & 31 & 31 & 31 \\
CodeBrain & released & 200 & 34 & 34 & 34 \\
BrainOmni & base & 512 & 32 & 32 & 0 \\
EEG-DINO (small) & small & 200 & 28 & 28 & 28 \\
EEG-DINO (medium) & medium & 512 & 28 & 28 & 28 \\
Signal-JEPA & released & 64 & 21 & 21 & 21 \\
LEAD & released & 128 & 32 & 32 & 32 \\
OSF & base & 768 & 34 & 34 & 0 \\
ZUNA (hidden) & 1.1 hidden & 1024 & 2 & 32 & 32 \\
\bottomrule
\end{tabular}
\end{table}

\subsection{Classical signal descriptors}
\label{app:representations}

Descriptors are computed separately for each normalized window. Band-based descriptors use six frequency bands: $[1,4)$, $[4,8)$, $[8,13)$, $[13,20)$, $[20,30)$, and $[30,45)$~Hz. We evaluate the following groups.

\paragraph{Band power.}
For each channel, we compute the full-window FFT, sum its power within each band, and take the natural logarithm after adding $10^{-12}$. This gives six values per channel.

\paragraph{Spectrum.}
We concatenate log Welch power spectra~\citep{welch1967} estimated with one-, two-, and four-second Hann segments, 50\% overlap, and mean removal within each segment. Frequencies from 0.5 to 45~Hz are retained. Power is floored at $10^{-20}$ before taking the natural logarithm.

\paragraph{Covariance.}
We filter each window in each band with a fourth-order forward--backward Butterworth filter. We then compute $XX^\top/T$ on the whole window, its two halves, and its four quarters, where $T$ is the number of samples in the relevant interval. These are second-moment matrices; the filtered subwindows are not centered again. Each matrix is shrunk toward the trace-scaled identity with coefficient 0.001, then mapped by the matrix logarithm, with eigenvalues floored at $10^{-12}$. We retain the upper triangle and multiply off-diagonal entries by $\sqrt{2}$.

\paragraph{Coherence.}
We estimate complex coherency, $S_{ij}/\sqrt{S_{ii}S_{jj}}$, using one-second Hann segments, 50\% overlap, and segment-wise mean removal. We average within each frequency band and retain the real and imaginary parts for every distinct channel pair. This descriptor is complex coherency, rather than magnitude-squared coherence alone.

The five descriptor variants are \texttt{bp} (band power), \texttt{spec} (spectrum), \texttt{cov} (covariance), \texttt{speccov} (spectrum and covariance), and \texttt{full} (spectrum, covariance, and coherence). The main results use \texttt{full}.

For $C$ channels, the spectrum, covariance, and coherence groups contain $314C$, $42C(C+1)/2$, and $12C(C-1)/2$ values, respectively. The full descriptor therefore contains 15{,}998 values for 19 channels and 2{,}946 values for six channels. No subject-specific reference or alignment is fitted when computing these per-window descriptors.

\paragraph{Standardization and PCA.}
For each fold and descriptor variant, columns are standardized on the training and validation windows, without labels. Columns with standard deviation at most $10^{-6}$ are set to zero. Each group is divided by the square root of its number of active columns times the number of groups, giving the groups equal total weight. PCA is fitted to this representation on the same non-test windows.

If the encoder readout has width $d$, $H_d$ retains the leading $\min(d,r)$ components, where $r$ is the numerical rank. This dimensionality cap limits the imbalance between the handcrafted and encoder feature blocks and reduces downstream classifier training costs. The rank threshold is $10^{-8}$ times the largest eigenvalue of the centered Gram matrix. PCA scores are not whitened. Validation windows thus contribute to this label-free preprocessing, but test windows contribute to neither standardization nor PCA. Crucially, repeating the comparisons without PCA preserves the substantial reduction in the average pretraining advantage, although individual residual gains differ (Appendix~\ref{app:s3_width}). Dimensionality reduction therefore improves computational efficiency while preserving the main qualitative conclusion.

\subsection{Readouts and classifier fitting}
\label{app:s3_readouts}

We compare four arms: $Z_0$, $Z_P$, $[H_d,Z_0]$, and $[H_d,Z_P]$. The first two use features from the random and pretrained encoders. The other two also receive the same classical descriptors. Each arm selects its own hyperparameters on validation performance, using the same search space as its paired arm.

\paragraph{Main readout: attention pooling.}
We use single-query attention pooling as the main readout to accommodate the different token-grid sizes produced by the evaluated encoders while keeping the downstream head compact. Concatenating all tokens would make classifier size grow with the number of tokens, introducing substantial variation in downstream capacity across architectures and complicating the interpretation of their scores. Attention pooling instead produces one $D$-dimensional vector per window and allows task-dependent weighting of tokens using only $D$ additional query parameters. This choice addresses the sensitivity to classifier design highlighted by EEG-FM-Bench \citep{eegfmbench}. It removes dependence of head size on token count, while the pretrained and random versions of each encoder retain exactly matched heads; feature widths and head sizes can still differ across architectures.

For tokens $z_i\in\mathbb{R}^{D}$, a single learned query $q$ gives
\begin{equation}
a_i=\frac{\exp(q^\top z_i/\sqrt{D})}{\sum_j\exp(q^\top z_j/\sqrt{D})}, \qquad z_{\mathrm{pool}}=\sum_i a_i z_i.
\end{equation}
A linear classifier receives $z_{\mathrm{pool}}$, or its concatenation with $H_d$. Here $d=D$. Only the query and classifier are trained; the encoder remains frozen. The query starts at zero, giving uniform attention at initialization.

Token features are standardized per dimension over the fitting windows and tokens. The PCA scores are centered on the fitting windows and scaled by one common scalar to unit average component variance. This scaling does not whiten individual components.

We use full-batch Adam, learning rate 0.01, and cross-entropy loss. The L2 penalty on the query and classification weights is selected from $\{0.001,0.01,0.1\}$. Validation $\kappa$ is evaluated every five steps, up to 300 steps, to select the penalty and training duration. Ties between penalties favor stronger regularization; ties between steps retain the earlier step. The query and classifier are then reinitialized and refitted on training plus validation windows for the selected number of steps. We repeat this procedure with head seeds 0, 1, and 2 and average their test scores, rather than voting over their predictions.

Attention produces one vector per window, regardless of the number of tokens. It does not give every model the same feature width or head size. For $K$ classes, the encoder-only head has $D+K(D+1)$ trainable parameters. With $d_H$ retained classical components, it has $D+K(D+d_H+1)$. Head sizes are matched between pretrained and random versions of each model, not across different models.

\paragraph{Global mean and electrode-wise pooling.}
Global mean pooling averages all tokens, giving width $d=D$. Electrode-wise pooling averages over time within each electrode and concatenates the results, giving width $d=CD$ when an electrode axis is available. In the experiments accompanying the main PCA results, these two readouts use linear ridge regression to one-hot targets. Predictions are the class with the largest score.

Encoder features are standardized on the fitting data and divided by the square root of their number of active dimensions. PCA scores are centered without component-wise rescaling. Ridge uses centered linear kernels and regularization $n_{\mathrm{fit}}\alpha$, with $\alpha\in\{10,1,10^{-1},10^{-2},10^{-3},10^{-4},10^{-5}\}$. For combined arms, the kernel is $(1-w)K_H+wK_Z$, with $w\in\{0,0.1,0.5,0.9,1\}$. The endpoints allow the classifier to ignore either block. Validation $\kappa$ selects the hyperparameters; ties favor larger $\alpha$, then smaller $w$. The selected classifier is refitted on training plus validation windows.

These comparisons change the pooling, classifier fitting procedure, and, for electrode-wise pooling, the width of $H_d$. They therefore do not isolate the effect of pooling alone. We address this with uniform-pooling controls and a further experiment using cross-entropy classifiers for all three readouts (Appendix~\ref{app:s3_reader_controls}).

\subsection{Scores, averages, and uncertainty}
\label{app:s3_statistics}

We report Cohen's $\kappa$ over all held-out test windows in a model--corpus pair. For leave-one-subject-out evaluation, out-of-fold predictions are concatenated before scoring. Differences use the same test observations for the compared arms. The two contrasts are $\Delta_{\mathrm{pre}}=s(Z_P)-s(Z_0)$ and $\Delta_H=s([H_d,Z_P])-s([H_d,Z_0])$, as in Equations~\ref{eq:delta_pre} and~\ref{eq:delta_h}.

For each pair, 95\% confidence intervals use 2{,}000 paired bootstrap resamples of test subjects. All windows from a sampled subject stay together, and the same resample is used for both arms. Attention scores are averaged across head seeds within each resample.

Each small model-level point in the main figure averages that model's available corpora within a family. The family summary instead averages all available model--corpus pairs. Its 95\% interval uses 2{,}000 bootstrap resamples of the rows and columns of the model-by-corpus matrix, sampled independently with replacement, and averages the non-missing entries. This family bootstrap does not resample subjects again. The overall summary gives equal weight to the five main family means; it is not an average of every pair across all families.

\subsection{What is trained in each experiment}
\label{app:pipelines}
Table~\ref{tab:pipelines} summarizes, for every experiment, which weights are pretrained, which are trained on target labels, and whether the encoder stays frozen.

\begin{table}[!htbp]
\centering
\footnotesize
\setlength{\tabcolsep}{3pt}
\caption{\textbf{What is pretrained, what is trained on labels, and on whom, in each experiment.} Downstream splits are always subject-disjoint.}
\label{tab:pipelines}
\begin{tabular}{p{2.2cm}p{3.4cm}p{3.6cm}p{3.6cm}}
\toprule
Experiment & Encoder weights & Trained on target labels & Encoder during readout \\
\midrule
Section~\ref{sec:results} & 23 published checkpoints, and a random copy of each & attention query and linear layer (with $H_d$ concatenated when stated) & frozen \\
Section~\ref{sec:surrogates} & REVE-small (and the controls of Table~\ref{tab:synthetic_objective_controls}) pretrained by us on original or transformed TUEG & attention query and linear layer; EEGNetv4 is trained end to end on each signal version & frozen \\
Section~\ref{sec:retention} & REVE-small trained from scratch on one corpus, by reconstruction with or without a label term, on encoder-training subjects only & a fresh attention query and linear layer at each checkpoint, on separate subjects & frozen \\
Appendix~\ref{app:full_finetuning} & published REVE-base, and a random copy & the whole encoder and head (FT), the head first and then everything (LP-FT), or rank-8 LoRA adapters and the head & fine-tuned \\
\bottomrule
\end{tabular}
\end{table}

\subsection{Other preprocessings}
\label{app:preprocessing}

All encoders read one common preprocessing (Appendix~\ref{app:s3_data}), but several were pretrained differently: CBraMod on 0.3--75~Hz with a line-noise notch, LaBraM on 0.1--75~Hz, and most on signals in microvolts, which keep amplitude differences between windows and electrodes. We re-read 20 encoders of Figure~\ref{fig:benchmark_gains} and their random copies on the 7 motor-imagery and 4 ERP corpora with 0.3--75~Hz recordings, under \emph{(i)} a 0.3--75~Hz band-pass with a notch at the local line frequency, \emph{(ii)} one divisor for the whole corpus (the median channel standard deviation) instead of one per window and channel, and both. Everything else follows Figure~\ref{fig:benchmark_gains}, and EEGNet is retrained on each version.

Neither change, alone or combined, alters what pretraining adds (Table~\ref{tab:preprocessing}): every change in the gain is 0.01~$\kappa$ or less, with a 95\% interval that contains zero, and the gains of individual encoders stay correlated across preprocessings ($r$ = 0.95--0.98 on motor imagery, 0.73--0.81 on ERP). EEGNet stays at 0.39--0.42 on motor imagery and 0.23--0.26 on ERP: the information is present under every preprocessing, and pretraining does not extract it under any.

\begin{table}[!ht]
\centering
\scriptsize
\setlength{\tabcolsep}{3pt}
\caption{\textbf{Pretraining gain under four preprocessings.} Mean over the same encoder--corpus pairs in every row (MI: motor imagery; number of corpora in parentheses); z-scored: per window and channel; gain = pretrained $-$ random; change = gain $-$ gain under the paper's preprocessing, on the same pairs. In brackets, 95\% intervals from resampling encoders and corpora.}
\label{tab:preprocessing}
\begin{tabular}{llrrrrr}
\toprule
Family & Preprocessing & Random & Pretrained & Gain & Change vs.\ paper & EEGNet \\
\midrule
MI (7) & 0.5--45~Hz, z-scored (paper) & $0.076$ & $0.108$ & $0.032$ [-0.006, +0.074] & --- & $0.415$ \\
 & 0.3--75~Hz + notch, z-scored & $0.081$ & $0.114$ & $0.033$ [-0.006, +0.074] & $0.001$ [-0.008, +0.010] & $0.400$ \\
 & 0.5--45~Hz, amplitude kept & $0.078$ & $0.100$ & $0.023$ [-0.014, +0.061] & $-0.009$ [-0.022, +0.003] & $0.394$ \\
 & 0.3--75~Hz + notch, amplitude kept & $0.086$ & $0.108$ & $0.022$ [-0.013, +0.060] & $-0.010$ [-0.026, +0.004] & $0.418$ \\
\midrule
ERP (4) & 0.5--45~Hz, z-scored (paper) & $0.014$ & $0.010$ & $-0.004$ [-0.016, +0.008] & --- & $0.261$ \\
 & 0.3--75~Hz + notch, z-scored & $0.012$ & $0.010$ & $-0.002$ [-0.012, +0.012] & $0.002$ [-0.004, +0.009] & $0.234$ \\
 & 0.5--45~Hz, amplitude kept & $0.008$ & $0.004$ & $-0.004$ [-0.016, +0.007] & $-0.001$ [-0.009, +0.008] & $0.256$ \\
 & 0.3--75~Hz + notch, amplitude kept & $0.008$ & $0.006$ & $-0.002$ [-0.010, +0.008] & $0.002$ [-0.006, +0.010] & $0.240$ \\
\bottomrule
\end{tabular}

\end{table}

\section{Additional Results for Section~\ref{sec:results}}
\label{app:section3_results}

\subsection{Absolute scores and pretraining gains}
\label{app:s3_absolute_results}

Tables~\ref{tab:s3_absolute} and~\ref{tab:stats_benchmark} give the absolute scores and paired gains for the main attention readout. Equal weighting of the five main families gives $\Delta_{\mathrm{pre}}=+0.0698$ and $\Delta_H=+0.0024$, reported as $+0.070$ and $+0.002$ in the main text.

The absolute scores matter when interpreting a small difference. On ERP, both combined arms have mean $\kappa=0.021$. Their similar scores do not establish that the classical descriptors capture everything needed for the task. Conversely, small average differences can hide positive and negative results: among the 652 attention pairs, 89 have a subject-bootstrap interval for $\Delta_H$ entirely above zero, and 67 entirely below zero. These are descriptive counts without multiple-comparison correction.

\begin{table}[!htbp]
\centering
\small
\setlength{\tabcolsep}{5pt}
\caption{Absolute test $\kappa$ with attention pooling and the full descriptor $H_d$. Values average the available pairs within each family. $n$ is the number of model--evaluation dataset pairs.}
\label{tab:s3_absolute}
\begin{tabular}{lrrrrr}
\toprule
Family & $n$ & $Z_0$ & $Z_P$ & $[H_d,Z_0]$ & $[H_d,Z_P]$ \\
\midrule
Motor imagery & 138 & 0.075 & 0.104 & 0.220 & 0.221 \\
ERP & 113 & 0.018 & 0.015 & 0.021 & 0.021 \\
SSVEP & 100 & 0.120 & 0.175 & 0.387 & 0.389 \\
Sleep & 50 & 0.269 & 0.431 & 0.564 & 0.571 \\
Clinical & 115 & 0.258 & 0.363 & 0.511 & 0.514 \\
Emotion & 62 & 0.020 & 0.037 & 0.112 & 0.113 \\
Workload & 53 & 0.054 & 0.134 & 0.176 & 0.190 \\
Imagined speech & 21 & -0.001 & 0.005 & 0.014 & 0.016 \\
\bottomrule
\end{tabular}
\end{table}

Table~\ref{tab:stats_benchmark} also reports Holm-corrected tests of each family's gain, the share of the difference between a random encoder and EEGNetv4 that pretraining covers, and equivalence tests of $\Delta_H$ ($\pm0.02~\kappa$). Over the 27 datasets, sleep and clinical gains exceed motor-imagery and ERP gains by $+0.120~\kappa$ [$0.066$, $0.188$]; this survives a permutation of task labels over datasets ($p<0.001$), holds for 16 of 19 encoders (Wilcoxon $p=0.001$), stays $+0.113$ without the encoder--dataset pairs known to overlap with pretraining, and vanishes beyond descriptors ($+0.005$ [$-0.007$, $+0.017$]).

\begin{table}[!ht]
\centering
\small
\setlength{\tabcolsep}{2.2pt}
\caption{\textbf{Statistics of the benchmark.} $\Delta_{\mathrm{pre}}$ and $\Delta_H$ with the 95\% family bootstrap intervals of Appendix~\ref{app:s3_statistics}. $p_{\mathrm{Holm}}$: two-sided bootstrap test against zero, Holm-corrected over the five main families. Gap closed: share of the distance from the random encoder to EEGNetv4 that the pretrained encoder covers, $(Z_P-Z_0)/(\mathrm{EEGNet}-Z_0)$. $p_{\mathrm{TOST}}$: equivalence of $\Delta_H$ to zero within $\pm 0.02~\kappa$, the 90\% interval lying inside the margin at $p<0.05$. Bounds very close to zero retain an extra decimal place to show their sign; the SSVEP lower bound is zero within Monte Carlo error. The last three families are additional evaluations.}
\label{tab:stats_benchmark}
\begin{tabular}{lrrrrrrr}
\toprule
Family & $\Delta_{\mathrm{pre}}$ & 95\% interval & $p_{\mathrm{Holm}}$ & Gap closed & $\Delta_H$ & 95\% interval & $p_{\mathrm{TOST}}$ \\
\midrule
Motor imagery & $+0.029$ & $[-0.007, +0.069]$ & $0.270$ & $9\%$ & $+0.000$ & $[-0.005, +0.006]$ & $<0.001$ \\
ERP & $-0.002$ & $[-0.015, +0.012]$ & $0.689$ & $-1\%$ & $+0.000$ & $[-0.010, +0.013]$ & $0.003$ \\
SSVEP & $+0.055$ & $[-0.0003, +0.118]$ & $0.145$ & $16\%$ & $+0.002$ & $[-0.013, +0.016]$ & $0.008$ \\
Sleep & $+0.162$ & $[+0.094, +0.281]$ & $<0.001$ & $63\%$ & $+0.007$ & $[-0.0001, +0.021]$ & $0.025$ \\
Clinical & $+0.105$ & $[+0.051, +0.171]$ & $0.001$ & $63\%$ & $+0.002$ & $[-0.016, +0.020]$ & $0.028$ \\
\addlinespace[2pt]
Emotion & $+0.018$ & $[+0.007, +0.030]$ & --- & --- & $+0.001$ & $[-0.002, +0.004]$ & $<0.001$ \\
Workload & $+0.080$ & $[+0.036, +0.120]$ & --- & --- & $+0.014$ & $[+0.003, +0.026]$ & $0.151$ \\
Imagined speech & $+0.006$ & $[+0.002, +0.010]$ & --- & --- & $+0.002$ & $[-0.002, +0.005]$ & $<0.001$ \\
\bottomrule
\end{tabular}
\end{table}

\paragraph{Absolute rankings do not isolate the contribution of pretraining.}
Absolute downstream scores and gains from pretrained weights can lead to different conclusions about the same models. Table~\ref{tab:s3_absolute_rankings} illustrates this distinction on four corpora using the main attention-pooling protocol, encoder features only, and the full annotation budget. Within each corpus, all entries use the same labeled windows and subject partitions. Both random and pretrained encoders remain frozen; only the attention query and linear classifier are trained. Thus, ``random'' refers to the encoder weights, not to an untrained downstream classifier.

\begin{table}[t]
\centering
\small
\setlength{\tabcolsep}{4pt}
\caption{Selected examples contrasting absolute performance and the contribution of pretraining. Scores are test Cohen's $\kappa$, averaged over three downstream-head seeds. $D$ is the pooled feature width. Each gain compares pretrained and random weights of the same architecture; its interval uses 2{,}000 paired test-subject bootstrap resamples (Appendix~\ref{app:s3_statistics}). Differences are computed before rounding. These intervals concern within-architecture gains, not cross-architecture score differences, and do not include variation across encoder initializations or pretraining runs.}
\label{tab:s3_absolute_rankings}
\begin{tabular}{llrrrl}
\toprule
Corpus & Encoder & $D$ & $s(Z_0)$ & $s(Z_P)$ & $\Delta_{\mathrm{pre}}$ [95\% interval] \\
\midrule
Stieger2021 & REVE & 512 & 0.256 & 0.304 & $+0.047\;[+0.019,+0.080]$ \\
 & EEG-DINO (medium) & 512 & 0.077 & 0.125 & $+0.048\;[+0.024,+0.074]$ \\
\midrule
Lee2019 SSVEP & BIOT & 256 & 0.866 & 0.645 & $-0.220\;[-0.280,-0.165]$ \\
 & REVE & 512 & 0.583 & 0.816 & $+0.233\;[+0.186,+0.281]$ \\
 & CBraMod & 200 & 0.063 & 0.674 & $+0.611\;[+0.537,+0.680]$ \\
\midrule
ISRUC & EEGPT & 512 & 0.448 & 0.494 & $+0.046\;[+0.020,+0.072]$ \\
 & CSBrain & 200 & 0.171 & 0.417 & $+0.246\;[+0.209,+0.285]$ \\
\midrule
TUAB & BIOT & 256 & 0.477 & 0.549 & $+0.071\;[+0.037,+0.106]$ \\
 & CSBrain & 200 & 0.188 & 0.415 & $+0.227\;[+0.184,+0.269]$ \\
\bottomrule
\end{tabular}
\end{table}

On Stieger2021, randomly initialized REVE scores $0.256$, above pretrained EEG-DINO (medium) at $0.125$. Yet both architectures benefit from pretraining by almost the same amount: $+0.047$ and $+0.048$, respectively. Both produce 512-dimensional features and use 2{,}564 trainable readout parameters, so this example does not arise from a difference in head size. The much higher absolute score of pretrained REVE ($0.304$ versus $0.125$) coexists with a similar within-architecture pretraining gain.

On Lee2019 SSVEP, random BIOT reaches $0.866$, above pretrained REVE ($0.816$) and CBraMod ($0.674$). Nevertheless, pretraining improves REVE by $+0.233$ and CBraMod by $+0.611$ relative to their own random encoders. REVE therefore has the higher pretrained score, while CBraMod has the larger gain from pretrained weights. BIOT provides the converse example: its pretrained score of $0.645$ accompanies a decrease of $0.220$ from its random reference. A high absolute score can therefore coexist with either a substantial positive or a negative pretraining gain.

The distinction also appears in sleep and clinical evaluation. On ISRUC, random EEGPT scores $0.448$, above pretrained CSBrain at $0.417$. Pretrained EEGPT also ranks higher ($0.494$), although its gain is smaller: $+0.046$ versus $+0.246$ for CSBrain. On TUAB, random BIOT scores $0.477$, above pretrained CSBrain at $0.415$, while their respective pretraining gains are $+0.071$ and $+0.227$. These comparisons show that a pretrained encoder can benefit substantially from pretraining even when a different architecture with random weights scores higher. The cross-architecture orderings reported here are descriptive comparisons of observed scores, not claims of statistically significant superiority.

For two architectures $A$ and $B$ evaluated on the same corpus, the distinction can be written exactly as
\begin{equation}
\begin{aligned}
s(Z_P^A)-s(Z_P^B)
={}&\bigl[s(Z_0^A)-s(Z_0^B)\bigr]
&+\bigl[\Delta_{\mathrm{pre}}^A-\Delta_{\mathrm{pre}}^B\bigr].
\label{eq:s3_ranking_decomposition}
\end{aligned}
\end{equation}
An absolute ranking therefore combines differences already present with random encoders and differences in the gains from pretraining. Random-encoder performance reflects the architecture, input representation, initialization, and supervised readout together; it is not a measure of architecture alone. Tokenization, channel handling, positional information, feature width, and compatibility with the readout can all affect the resulting pipeline.

Absolute scores remain useful for selecting a pipeline under a specified evaluation protocol, but cannot by themselves identify what pretraining contributes. We therefore report them alongside paired pretrained--random contrasts with the architecture and downstream training procedure held fixed. The same reasoning motivates $\Delta_H$: once both classifiers receive identical classical descriptors, their paired difference measures the remaining predictive benefit of pretrained weights. Neither contrast ranks pretraining objectives in isolation, since the released models also differ in their pretraining data and training recipes.
\subsection{Which classical descriptors reduce the gain?}
\label{app:s3_descriptor_results}

Table~\ref{tab:s3_ablation} reports each descriptor variant under each readout. With attention, spectra alone reduce the sleep and clinical gains to about $+0.016$. Adding covariance reduces them to $+0.006$ and $+0.001$, respectively. Adding coherence gives $+0.007$ and $+0.002$, so it does not reduce the remaining gain further in these two families.

These comparisons need not decrease monotonically. Adding a group changes the relative weight of the existing groups and can change which directions PCA retains. The variants are finite descriptions of the signal, not an exhaustive representation of every spectral and spatial property.

\begin{table}[!htbp]
\centering
\scriptsize
\setlength{\tabcolsep}{3pt}
\caption{Descriptor ablation. Without $H$, values are $\Delta_{\mathrm{pre}}$; the other columns give $\Delta_H$ for the indicated descriptor reduced to the readout width. Each row averages all available pairs for that readout and family. A dash means the combination was not evaluated. Global mean and electrode-wise readouts in this table use ridge; attention uses cross-entropy.}

\label{tab:s3_ablation}
\begin{tabular}{llrrrrrr}
\toprule
Readout & Family & Without $H$ & bp & spec & cov & speccov & full \\
\midrule
Attention & Motor imagery & $+0.029$ & --- & $+0.006$ & --- & $-0.001$ & $+0.000$ \\
 & ERP & $-0.002$ & --- & $+0.002$ & --- & $+0.000$ & $+0.000$ \\
 & SSVEP & $+0.055$ & --- & $-0.002$ & --- & $-0.001$ & $+0.002$ \\
 & Sleep & $+0.162$ & --- & $+0.016$ & --- & $+0.006$ & $+0.007$ \\
 & Clinical & $+0.105$ & --- & $+0.016$ & --- & $+0.001$ & $+0.002$ \\
\midrule
Global mean & Motor imagery & $+0.015$ & $-0.001$ & $+0.003$ & $-0.002$ & $+0.002$ & $-0.003$ \\
 & ERP & $-0.006$ & $-0.004$ & $-0.002$ & $-0.005$ & $-0.004$ & $-0.004$ \\
 & SSVEP & $+0.020$ & $-0.002$ & $+0.001$ & $-0.007$ & $-0.005$ & $+0.006$ \\
 & Sleep & $+0.149$ & $+0.033$ & $+0.034$ & $+0.024$ & $+0.013$ & $+0.018$ \\
 & Clinical & $+0.090$ & $+0.043$ & $+0.034$ & $+0.040$ & $+0.028$ & $+0.030$ \\
\midrule
Electrode-wise & Motor imagery & $+0.047$ & $+0.024$ & $+0.029$ & $-0.002$ & $+0.003$ & $+0.005$ \\
 & ERP & $-0.011$ & $-0.010$ & $-0.010$ & $-0.010$ & $-0.011$ & $-0.012$ \\
 & SSVEP & $+0.016$ & $+0.004$ & $+0.003$ & $+0.004$ & $+0.001$ & $+0.007$ \\
 & Sleep & $+0.185$ & $+0.039$ & $+0.039$ & $+0.026$ & $+0.020$ & $+0.020$ \\
 & Clinical & $+0.119$ & $+0.041$ & $+0.025$ & $+0.047$ & $+0.003$ & $+0.035$ \\
\bottomrule
\end{tabular}
\end{table}

\subsection{Results by corpus and encoder}
\label{app:s3_corpus_model_results}

Table~\ref{tab:s3_corpus_scores} gives the four absolute scores and two contrasts for every corpus, averaging available encoders. Tables~\ref{tab:s3_model_raw} and~\ref{tab:s3_model_conditional} give encoder-level means in the five main families. Coverage differs across encoders, so these means should not be treated as a ranking on a common set of datasets. A missing entry indicates no available result in that family.

\begin{table}[!htbp]
\centering
\scriptsize
\setlength{\tabcolsep}{3.5pt}
\caption{Results by corpus with attention pooling and $H_d$ from the full descriptor. Each row averages $n$ encoder configurations. Contrasts are computed before rounding.}
\label{tab:s3_corpus_scores}
\begin{tabular}{lrrrrrrr}
\toprule
Corpus / task & $n$ & $Z_0$ & $Z_P$ & $[H_d,Z_0]$ & $[H_d,Z_P]$ & $\Delta_{\mathrm{pre}}$ & $\Delta_H$ \\
\midrule
BNCI2014-001 & 16 & 0.056 & 0.095 & 0.187 & 0.193 & +0.039 & +0.006 \\
Cho2017 & 21 & 0.066 & 0.080 & 0.164 & 0.162 & +0.014 & -0.002 \\
Dreyer2023 & 17 & 0.052 & 0.076 & 0.233 & 0.233 & +0.024 & +0.000 \\
Lee2019 MI & 21 & 0.039 & 0.056 & 0.237 & 0.239 & +0.017 & +0.003 \\
PhysioNet MI & 21 & 0.077 & 0.081 & 0.190 & 0.186 & +0.005 & -0.004 \\
Schirrmeister2017 & 21 & 0.130 & 0.208 & 0.315 & 0.320 & +0.079 & +0.004 \\
Stieger2021 & 21 & 0.095 & 0.123 & 0.213 & 0.210 & +0.027 & -0.003 \\
\midrule
BNCI2014-009 & 19 & 0.057 & 0.053 & 0.054 & 0.052 & -0.003 & -0.002 \\
BNCI2015-003 & 19 & -0.002 & 0.000 & 0.003 & 0.004 & +0.002 & +0.001 \\
ERP CORE MMN & 18 & 0.002 & 0.001 & -0.004 & -0.009 & -0.001 & -0.004 \\
ERP CORE N170 & 18 & 0.018 & 0.022 & 0.018 & 0.028 & +0.004 & +0.010 \\
ERP CORE N2pc & 18 & 0.026 & 0.013 & 0.052 & 0.051 & -0.013 & -0.001 \\
Lee2019 ERP & 21 & 0.006 & 0.005 & 0.004 & 0.001 & -0.002 & -0.003 \\
\midrule
BETA & 21 & 0.049 & 0.069 & 0.221 & 0.220 & +0.020 & -0.002 \\
Kalunga2016 & 19 & 0.072 & 0.168 & 0.362 & 0.382 & +0.096 & +0.020 \\
Lee2019 SSVEP & 21 & 0.271 & 0.377 & 0.654 & 0.653 & +0.106 & -0.001 \\
MAMEM3 & 20 & 0.050 & 0.051 & 0.109 & 0.107 & +0.000 & -0.002 \\
Nakanishi2015 & 19 & 0.152 & 0.207 & 0.593 & 0.588 & +0.055 & -0.005 \\
\midrule
HMC & 21 & 0.218 & 0.361 & 0.516 & 0.522 & +0.143 & +0.006 \\
ISRUC & 21 & 0.298 & 0.432 & 0.591 & 0.595 & +0.134 & +0.004 \\
Sleep-EDF & 8 & 0.327 & 0.610 & 0.619 & 0.640 & +0.284 & +0.021 \\
\midrule
ADFTD & 21 & 0.070 & 0.161 & 0.314 & 0.321 & +0.092 & +0.007 \\
CHB-MIT & 10 & 0.257 & 0.345 & 0.497 & 0.523 & +0.088 & +0.026 \\
Mumtaz & 21 & 0.528 & 0.651 & 0.847 & 0.830 & +0.123 & -0.018 \\
PD Iowa & 21 & 0.132 & 0.175 & 0.332 & 0.311 & +0.044 & -0.021 \\
TUAB & 21 & 0.298 & 0.437 & 0.568 & 0.578 & +0.139 & +0.010 \\
TUEV & 21 & 0.262 & 0.398 & 0.501 & 0.524 & +0.136 & +0.023 \\
\midrule
DREAMER & 20 & 0.007 & 0.022 & 0.106 & 0.107 & +0.015 & +0.001 \\
FACED & 21 & 0.027 & 0.042 & 0.112 & 0.112 & +0.015 & +0.000 \\
SEED-V & 21 & 0.024 & 0.048 & 0.119 & 0.120 & +0.023 & +0.001 \\
\midrule
EEGMAT & 21 & 0.069 & 0.165 & 0.181 & 0.197 & +0.096 & +0.016 \\
SEED-VIG & 18 & 0.066 & 0.145 & 0.169 & 0.180 & +0.078 & +0.011 \\
Shin2017B & 14 & 0.015 & 0.071 & 0.178 & 0.192 & +0.056 & +0.014 \\
\midrule
BCIC2020-3 & 21 & -0.001 & 0.005 & 0.014 & 0.016 & +0.006 & +0.002 \\
\bottomrule
\end{tabular}
\end{table}

\begin{table}[!htbp]
\centering
\small
\setlength{\tabcolsep}{4pt}
\caption{Gain from pretraining without classical descriptors, $\Delta_{\mathrm{pre}}$, by encoder and family. Attention pooling; each entry averages that encoder's available corpora.}
\label{tab:s3_model_raw}
\begin{tabular}{lrrrrr}
\toprule
Encoder & Motor imagery & ERP & SSVEP & Sleep & Clinical \\
\midrule
REVE & $+0.160$ & $+0.084$ & $+0.088$ & $+0.247$ & $+0.189$ \\
CBraMod & $+0.085$ & $+0.002$ & $+0.190$ & $+0.271$ & $+0.163$ \\
LUNA & $+0.049$ & $+0.000$ & $-0.031$ & $-0.082$ & $-0.010$ \\
EEGPT & $-0.020$ & $+0.003$ & $+0.074$ & $+0.038$ & $-0.015$ \\
LaBraM & $+0.044$ & $-0.015$ & $+0.022$ & $+0.039$ & $+0.071$ \\
CSBrain & $+0.032$ & $+0.004$ & $+0.141$ & $+0.312$ & $+0.186$ \\
EEGMamba & $+0.033$ & $-0.011$ & $+0.098$ & $+0.181$ & $+0.053$ \\
FEMBA & $+0.028$ & $+0.010$ & $+0.115$ & --- & $+0.054$ \\
ST-EEGFormer & --- & $+0.022$ & $+0.228$ & $+0.086$ & --- \\
ZUNA & $-0.024$ & $+0.010$ & $+0.042$ & $+0.052$ & $+0.050$ \\
BIOT & $-0.093$ & $-0.009$ & $-0.310$ & --- & $+0.069$ \\
BENDR & $-0.004$ & $-0.006$ & $+0.028$ & $-0.088$ & $-0.053$ \\
Neuro-GPT & $-0.166$ & $-0.031$ & $-0.038$ & $+0.154$ & $+0.079$ \\
NeuroLM & $+0.136$ & $+0.014$ & $+0.147$ & $+0.299$ & $+0.350$ \\
NeuroRVQ & $-0.009$ & $-0.026$ & $-0.060$ & $+0.016$ & $+0.014$ \\
CodeBrain & $+0.028$ & $-0.016$ & $-0.015$ & $+0.303$ & $+0.096$ \\
BrainOmni & $+0.170$ & $+0.004$ & $+0.223$ & $+0.443$ & $+0.475$ \\
EEG-DINO (small) & $+0.020$ & $-0.025$ & $+0.008$ & $+0.247$ & $+0.128$ \\
EEG-DINO (medium) & $+0.080$ & $-0.023$ & $+0.040$ & $+0.213$ & $+0.127$ \\
Signal-JEPA & $-0.044$ & $+0.002$ & $-0.016$ & $-0.104$ & $-0.130$ \\
LEAD & $+0.034$ & $-0.006$ & $+0.104$ & $+0.184$ & $+0.155$ \\
OSF & $+0.045$ & $-0.031$ & $+0.099$ & $+0.275$ & $+0.159$ \\
ZUNA (hidden) & --- & --- & --- & $+0.043$ & --- \\
\bottomrule
\end{tabular}
\end{table}

\begin{table}[p]
\centering
\small
\setlength{\tabcolsep}{4pt}
\caption{Remaining pretraining gain, $\Delta_H$, by encoder and family. Attention pooling with the full descriptor reduced to $H_d$; each entry averages that encoder's available corpora.}
\label{tab:s3_model_conditional}
\begin{tabular}{lrrrrr}
\toprule
Encoder & Motor imagery & ERP & SSVEP & Sleep & Clinical \\
\midrule
REVE & $+0.006$ & $+0.066$ & $+0.013$ & $+0.012$ & $-0.003$ \\
CBraMod & $+0.012$ & $+0.007$ & $+0.018$ & $+0.015$ & $+0.011$ \\
LUNA & $+0.007$ & $-0.002$ & $-0.010$ & $-0.009$ & $-0.025$ \\
EEGPT & $+0.003$ & $+0.000$ & $+0.009$ & $+0.006$ & $+0.008$ \\
LaBraM & $+0.008$ & $-0.016$ & $+0.002$ & $-0.012$ & $+0.005$ \\
CSBrain & $+0.002$ & $+0.020$ & $+0.012$ & $-0.006$ & $+0.004$ \\
EEGMamba & $+0.008$ & $-0.014$ & $-0.005$ & $+0.005$ & $+0.007$ \\
FEMBA & $+0.005$ & $+0.006$ & $+0.003$ & --- & $+0.007$ \\
ST-EEGFormer & --- & $-0.007$ & $+0.026$ & $+0.023$ & --- \\
ZUNA & $+0.001$ & $+0.002$ & $+0.002$ & $+0.002$ & $-0.014$ \\
BIOT & $-0.015$ & $+0.001$ & $-0.127$ & --- & $+0.042$ \\
BENDR & $+0.008$ & $-0.009$ & $+0.007$ & $-0.005$ & $+0.003$ \\
Neuro-GPT & $-0.027$ & $-0.026$ & $-0.019$ & $+0.001$ & $-0.015$ \\
NeuroLM & $-0.003$ & $+0.009$ & $-0.022$ & $+0.000$ & $-0.014$ \\
NeuroRVQ & $-0.004$ & $-0.021$ & $+0.011$ & $-0.002$ & $+0.015$ \\
CodeBrain & $+0.008$ & $+0.002$ & $+0.002$ & $-0.005$ & $-0.010$ \\
BrainOmni & $+0.002$ & $+0.002$ & $+0.002$ & $+0.001$ & $+0.003$ \\
EEG-DINO (small) & $-0.002$ & $-0.018$ & $+0.008$ & $+0.038$ & $-0.026$ \\
EEG-DINO (medium) & $+0.001$ & $-0.009$ & $+0.016$ & $+0.026$ & $+0.000$ \\
Signal-JEPA & $-0.009$ & $-0.004$ & $-0.010$ & $+0.005$ & $+0.001$ \\
LEAD & $+0.004$ & $+0.005$ & $+0.017$ & $+0.012$ & $+0.058$ \\
OSF & $-0.009$ & $-0.012$ & $+0.043$ & $+0.013$ & $-0.010$ \\
ZUNA (hidden) & --- & --- & --- & $+0.016$ & --- \\
\bottomrule
\end{tabular}
\end{table}

\subsection{Comparing readouts on the same model--dataset pairs}
\label{app:s3_matched_readouts}

The three readouts have different coverage. We therefore repeat the summary using only the 409 model--dataset pairs available under all three, including 325 pairs in the five main families (Table~\ref{tab:s3_common_readouts}). The reduction in gain remains visible on this common set. The residual still depends on the readout: in clinical tasks it is $+0.002$ with attention, $+0.036$ with global mean pooling, and $+0.037$ with electrode-wise pooling. Matching coverage removes differences in which pairs are evaluated; it does not remove differences in classifier fitting or in $H_d$'s width.

\begin{table}[!htbp]
\centering
\scriptsize
\setlength{\tabcolsep}{3pt}
\caption{Readout comparison on the same pairs. Each readout uses the main PCA protocol, including its original classifier fitting procedure. $n$ is identical across readouts within a row.}
\label{tab:s3_common_readouts}
\begin{tabular}{lrrrrrrr}
\toprule
& & \multicolumn{2}{c}{Attention} & \multicolumn{2}{c}{Global mean} & \multicolumn{2}{c}{Electrode-wise} \\
\cmidrule(lr){3-4}\cmidrule(lr){5-6}\cmidrule(lr){7-8}
Family & $n$ & $\Delta_{\mathrm{pre}}$ & $\Delta_H$ & $\Delta_{\mathrm{pre}}$ & $\Delta_H$ & $\Delta_{\mathrm{pre}}$ & $\Delta_H$ \\
\midrule
Motor imagery & 85 & $+0.046$ & $+0.003$ & $+0.026$ & $+0.001$ & $+0.061$ & $+0.010$ \\
ERP & 68 & $+0.002$ & $+0.004$ & $-0.003$ & $-0.002$ & $-0.008$ & $-0.009$ \\
SSVEP & 65 & $+0.068$ & $+0.006$ & $+0.027$ & $+0.010$ & $+0.024$ & $+0.009$ \\
Sleep & 36 & $+0.175$ & $+0.009$ & $+0.169$ & $+0.020$ & $+0.185$ & $+0.020$ \\
Clinical & 71 & $+0.113$ & $+0.002$ & $+0.110$ & $+0.036$ & $+0.130$ & $+0.037$ \\
Emotion & 39 & $+0.024$ & $+0.002$ & $+0.026$ & $+0.000$ & $+0.047$ & $-0.001$ \\
Workload & 32 & $+0.103$ & $+0.019$ & $+0.105$ & $+0.012$ & $+0.102$ & $+0.011$ \\
Imagined speech & 13 & $+0.007$ & $+0.004$ & $+0.010$ & $+0.012$ & $+0.006$ & $+0.007$ \\
\bottomrule
\end{tabular}
\end{table}

\subsection{Using classical descriptors without PCA}
\label{app:s3_width}

Because we used PCA to reduce compute time and to place the descriptor groups on an equal footing, this section checks that it did not introduce any evaluation artifact. To that end, we repeat the combined-arm comparison on the full descriptor matrix without reducing its dimension, and show that the results are unchanged. Two choices are involved here and should not be confused: ``full descriptor'' refers to the set of descriptor groups used (spectrum, covariance, and coherence), while ``without PCA'' refers to keeping all of their columns rather than projecting them onto principal components.

Groups are standardized and given equal total weight as above. In these no-PCA experiments, this preprocessing is fitted on training windows during selection and on training plus validation windows for the final fit. For attention, the concatenated descriptor matrix is scaled to total variance $D$ before entering the classifier. This matches the total variance of standardized token dimensions, but does not guarantee equal variance for the descriptor block and the pooled encoder vector.

Table~\ref{tab:s3_width} compares only pairs available in both versions. With global mean pooling and the same ridge search, the mean change in $\Delta_H$ over 230 common pairs is $-0.0010$. The attention comparison contains 611 common pairs. Its clinical residual increases from $+0.002$ to $+0.013$, while the sleep residual changes from $+0.007$ to $+0.009$. Thus, removing PCA preserves the broad reduction in pretraining gain, but does not give identical residuals. These results also include the stated change in how descriptor preprocessing is fitted. 

\begin{table}[!htbp]
\centering
\small
\setlength{\tabcolsep}{5pt}
\caption{Pretraining gain conditional on classical descriptors, with and without PCA. Each row uses exactly the same model--corpus pairs on both sides. The global mean comparison uses ridge in both versions; the attention comparison uses cross-entropy in both.}
\label{tab:s3_width}
\begin{tabular}{llrrr}
\toprule
Readout & Family & $n$ & $\Delta_H$, PCA & $\Delta_H$, no PCA \\
\midrule
Attention & Motor imagery & 137 & $+0.000$ & $-0.002$ \\
 & ERP & 113 & $+0.000$ & $-0.002$ \\
 & SSVEP & 60 & $+0.005$ & $+0.002$ \\
 & Sleep & 50 & $+0.007$ & $+0.009$ \\
 & Clinical & 115 & $+0.002$ & $+0.013$ \\
 & Emotion & 62 & $+0.001$ & $-0.001$ \\
 & Workload & 53 & $+0.014$ & $+0.007$ \\
 & Imagined speech & 21 & $+0.002$ & $+0.000$ \\
\midrule
Global mean & Motor imagery & 44 & $-0.006$ & $-0.002$ \\
 & ERP & 37 & $-0.004$ & $-0.007$ \\
 & SSVEP & 21 & $-0.006$ & $-0.006$ \\
 & Sleep & 48 & $+0.016$ & $+0.016$ \\
 & Clinical & 39 & $+0.017$ & $+0.013$ \\
 & Emotion & 41 & $+0.000$ & $-0.002$ \\
\bottomrule
\end{tabular}
\end{table}

\subsection{Separating pooling from classifier fitting}
\label{app:s3_reader_controls}

\paragraph{Uniform pooling with the attention training procedure.}
We replace learned attention by fixed uniform token weights and keep the same linear head, normalization, optimizer, regularization search, and validation selection. This control is evaluated without $H$. For pretrained encoders, learned attention improves average $\kappa$ over this control by $0.031$ on motor imagery and $0.051$ on SSVEP; the gains on ERP, sleep, and clinical tasks are $0.002$, $0.006$, and $0.007$. For random encoders, the corresponding gains are $0.016$, $0.030$, $-0.001$, $-0.004$, and $0.014$, respectively. These results show that pooling can affect decoding, but do not isolate the effect of learned attention when $H$ is also supplied.

\paragraph{A common classifier fitting procedure for all three readouts.}
In a further evaluation without PCA, global mean and electrode-wise features are read by the same cross-entropy training procedure as attention, with no learned query for these deterministic poolings. We use the same learning rate, L2 search, validation selection, refitting procedure, and three head seeds. Descriptor groups have equal weight and are scaled to total variance equal to the encoder readout width.

Table~\ref{tab:s3_common_ce} reports the 340 pairs available under all three readouts in this evaluation, including 263 in the five main families. Classical descriptors again reduce the average pretraining gain, while some residuals remain, especially on SSVEP and workload. This comparison supports the overall reduction under a common classifier fitting procedure; it does not require the residual to be zero for every task or readout.

\begin{table}[!htbp]
\centering
\scriptsize
\setlength{\tabcolsep}{3pt}
\caption{Common-pair comparison using cross-entropy classifiers for all three readouts and classical descriptors without PCA. Each row uses the same pairs across its six score columns. This is a separate evaluation from the main PCA results.}
\label{tab:s3_common_ce}
\begin{tabular}{lrrrrrrr}
\toprule
& & \multicolumn{2}{c}{Attention} & \multicolumn{2}{c}{Global mean} & \multicolumn{2}{c}{Electrode-wise} \\
\cmidrule(lr){3-4}\cmidrule(lr){5-6}\cmidrule(lr){7-8}
Family & $n$ & $\Delta_{\mathrm{pre}}$ & $\Delta_H$ & $\Delta_{\mathrm{pre}}$ & $\Delta_H$ & $\Delta_{\mathrm{pre}}$ & $\Delta_H$ \\
\midrule
Motor imagery & 76 & $+0.044$ & $-0.001$ & $+0.023$ & $+0.003$ & $+0.058$ & $+0.010$ \\
ERP & 62 & $-0.005$ & $-0.003$ & $-0.003$ & $-0.005$ & $-0.002$ & $-0.006$ \\
SSVEP & 28 & $+0.081$ & $+0.012$ & $+0.043$ & $+0.035$ & $+0.041$ & $+0.015$ \\
Sleep & 32 & $+0.181$ & $+0.013$ & $+0.161$ & $+0.015$ & $+0.177$ & $+0.015$ \\
Clinical & 65 & $+0.099$ & $+0.010$ & $+0.093$ & $-0.007$ & $+0.098$ & $-0.003$ \\
Emotion & 32 & $+0.022$ & $-0.001$ & $+0.023$ & $+0.001$ & $+0.036$ & $+0.006$ \\
Workload & 32 & $+0.103$ & $+0.017$ & $+0.092$ & $+0.015$ & $+0.091$ & $+0.014$ \\
Imagined speech & 13 & $+0.007$ & $+0.000$ & $+0.002$ & $+0.000$ & $+0.006$ & $-0.001$ \\
\bottomrule
\end{tabular}
\end{table}
\section{Experimental setup for Section~\ref{sec:surrogates}}
\label{app:surrogates_setup}

We use two complementary experiments. First, we change the signal used for self-supervised pretraining, using different synthetic signals, then evaluate the resulting models on real EEG. The main comparison uses masked signal reconstruction; an extension compares masked reconstruction, JEPA-style latent prediction, and their combination while keeping the REVE-small encoder architecture fixed. A separate control increases the encoder size to REVE-base under the reconstruction objective. Second, we train EEGNetv4 from scratch and apply the same signal transformation to its training, validation, and test data. The first experiment measures transfer from pretraining; the second measures supervised decoding of each signal version. They share the downstream windows and subject partitions. The transformations and their numerical validation are described in Appendix~\ref{app:surrogate_construction}; results for the additional objectives appear in Appendix~\ref{app:s4_objective_controls}.

\subsection{Data and subject splits}

Pretraining uses TUEG recordings with the 19-channel montage listed in Appendix~\ref{app:datasets}. Real windows are filtered to 0.5--45~Hz, sampled at 200~Hz, and standardized separately for each channel within each four-second window. Each window contains 800 samples per channel. Synthetic versions are generated from these preprocessed windows. For all TUEG pretraining experiments conducted in this work, we excluded recordings belonging to TUAB or any other downstream evaluation dataset. Both real and synthetic pretraining conditions use this same filtered corpus, ensuring no recording overlap between pretraining and downstream evaluation.

The six downstream corpora and their retained sample sizes are given in Table~\ref{tab:s4_data}. All methods use the same retained windows. BCI IV-2a uses nine leave-one-subject-out folds, with six training, two validation, and one test subject in each fold. The other corpora use one fixed subject split, approximately 45\%/15\%/40\% for training, validation, and testing (seed 20260908). TUAB splits are stratified using each person's majority label. All windows from a person stay in one partition within a fold. The window-sampling seed is 20260906.

\begin{table}[!htbp]
\centering
\small
\setlength{\tabcolsep}{4pt}
\caption{Retained downstream data. The last column counts training, validation, and test subjects. BCI IV-2a uses nine folds; the other corpora use one fixed split.}
\label{tab:s4_data}
\begin{tabular}{lrrrrl}
\toprule
Corpus & Windows & Subjects & Channels & Classes & Subject split \\
\midrule
BCI IV-2a & 5,184 & 9 & 22 & 4 & 6/2/1 (each fold) \\
PhysioNet MI & 6,000 & 109 & 19 & 5 & 49/16/44 \\
Lee2019 ERP & 5,994 & 54 & 19 & 2 & 24/8/22 \\
Lee2019 SSVEP & 5,994 & 54 & 19 & 4 & 24/8/22 \\
ISRUC & 6,000 & 98 & 6 & 5 & 44/15/39 \\
TUAB & 6,000 & 2326 & 19 & 2 & 1046/349/931 \\
\bottomrule
\end{tabular}
\end{table}

We retain 22 channels for BCI IV-2a, six for ISRUC, and the common 19-channel montage for the remaining corpora. The BCI IV-2a channels are Fz, FC3, FC1, FCz, FC2, FC4, C5, C3, C1, Cz, C2, C4, C6, CP3, CP1, CPz, CP2, CP4, P1, Pz, P2, and POz. The ISRUC channels are F3, C3, O1, F4, C4, and O2. TUEG is additionally evaluated on 1,024 fixed real windows for reconstruction only. This TUEG evaluation is a reference on the pretraining corpus, not evidence of transfer to an unseen corpus. Subject separation in downstream evaluation does not, by itself, establish separation from pretraining subjects or recordings.

\subsection{Architecture and masked-reconstruction pretraining}
\label{app:s4_pretraining}

The five MAE signal conditions use the same REVE-small MAE architecture. Its encoder has four Transformer layers, eight attention heads, and 512-dimensional tokens. The decoder has two layers, eight heads, and width 512. A token represents 200 samples from one electrode. Consecutive patches overlap by 20 samples, giving four patches per channel, starting at 0, 0.9, 1.8, and 2.7 seconds. Thus, patch extraction covers the first 740 samples of the 800-sample input; the signal transformation acts on the full window before patch extraction.

Training randomly masks 75\% of electrode--patch tokens. The loss is the mean absolute error (L1) on the masked target patches. It is not the squared error used to report reconstruction scores. For synthetic conditions, a new signal is drawn whenever a window is presented, before masking. Both the visible input and the reconstruction target come from this same transformed window.

We use AdamW with a constant learning rate of $10^{-3}$, weight decay 0.05, $\beta_1=0.9$, $\beta_2=0.95$, $\epsilon=10^{-8}$, and batch size 2,048. The standard-budget comparison uses approximately 20.8 million window presentations per run; checkpoint counts range from 20,819,968 to 20,847,310. Real EEG, PSD + cross-spectra, Covariance, and PSD each use three REVE-small MAE pretraining seeds. Mean + variance uses one pretraining run. Real EEG and PSD + cross-spectra are also evaluated with one run each at 104,236,550 presentations, five times the nominal standard budget of 20,847,310. These counts include repeated presentations of source windows.

\subsection{Additional objectives and encoder-size control}
\label{app:s4_objective_setup}

\paragraph{Objective comparison at fixed encoder architecture.}
We retain the REVE-small encoder, tokenization, positional encoding, 75\% masking rate, and two-layer decoder/predictor described above, and vary the pretraining objective. The three objectives are masked signal reconstruction (MAE), masked latent prediction (JEPA), and their sum (JEPA + reconstruction). The downstream encoder and its token grid are the same architecture in all three cases. The latent objectives add an exponential-moving-average (EMA) teacher and a latent output projection used only during pretraining; neither is used by the downstream readout.

Let $u$ denote the real or transformed input window and $\mathcal M$ its masked token indices. The student encoder, with parameters $\theta$, encodes the visible tokens, and the two-layer predictor produces an output $h_i$ at each masked position. A teacher encoder with parameters $\bar\theta$ receives the complete version of the same window. Its final-layer token $t_i$ is standardized over the feature dimension:
\begin{equation}
\bar t_i=\frac{t_i-\operatorname{mean}_d(t_i)}
{\sqrt{\operatorname{var}_d(t_i)+10^{-6}}},
\qquad
\bar\theta\leftarrow0.999\bar\theta+0.001\theta,
\label{eq:s4_teacher}
\end{equation}
where the variance uses denominator $D$. The teacher is initialized from the student encoder, receives no gradients, and uses clean electrode positions without the student's training-time positional jitter. We predict $\bar t_i$ through a linear latent projection $g_P$, with smooth L1 loss averaged over masked tokens and features:
\begin{equation}
L_P=\frac{1}{|\mathcal M|D}
\sum_{i\in\mathcal M}\sum_{d=1}^{D}
\ell_{\mathrm{SL1}}\bigl([g_P(h_i)]_d-\bar t_{id}\bigr),
\qquad
\ell_{\mathrm{SL1}}(e)=
\begin{cases}
e^2/2,& |e|<1,\\
|e|-1/2,& |e|\geq1.
\end{cases}
\label{eq:s4_latent_loss}
\end{equation}
Writing $L_R$ for the masked-patch L1 reconstruction loss, we train with
\begin{equation}
L_{\mathrm{MAE}}=L_R,\qquad
L_{\mathrm{JEPA}}=L_P,\qquad
L_{\mathrm{JEPA+recon}}=L_P+L_R.
\label{eq:s4_objectives}
\end{equation}
Both terms of the combined objective use the same masked positions and unit coefficients. Pure JEPA uses no signal-reconstruction loss. No additional anti-collapse regularizer is applied. These are controlled implementations of latent prediction and latent prediction combined with reconstruction, rather than reproductions of a complete published JEPA or EEGPT training recipe.

\paragraph{Signal conditions and size control.}
For each objective, we compare real EEG, PSD + cross-spectra, Covariance, and PSD. Synthetic windows are drawn afresh before masking; the student input, teacher input, and any signal target derive from the same draw. At a fixed signal condition, the REVE-small comparison changes the objective and its pretraining-only prediction components while retaining the encoder architecture, data, masking, optimizer settings, and downstream evaluation. At a fixed objective, the real--synthetic comparison changes the pretraining signal.

The separate size control uses REVE-base: 22 Transformer layers, eight attention heads, width 512, and approximately 69.2 million encoder parameters, compared with four layers and 12.7 million parameters for REVE-small. It retains the same patch layout, masking rate, and two-layer decoder and uses masked reconstruction under the same four signal conditions. The REVE-small grid comprises 36 pretraining runs (three objectives times four signals times three seeds); REVE-base uses four runs (one per signal). The mean-and-variance condition is evaluated only for REVE-small MAE, with one pretraining run.

\paragraph{Training budget and initialization.}
All conditions use the TUEG store and AdamW settings above, with batch size 2,048. Each REVE-small objective--signal condition uses three pretraining seeds, indexed 0, 1, and 2; each REVE-base signal condition uses one run. Standard-budget checkpoint counts lie between 20,819,968 and 20,847,310 window presentations, a difference of 0.13\%. The presentation budgets are therefore approximately matched; computational cost differs because the latent objectives require a teacher forward pass and REVE-base has a deeper encoder. Initialization and stochastic training draws vary across pretraining seeds, while the order of windows in the pre-shuffled store remains fixed. The individual random checkpoints A and B are reported as descriptive references; the aggregate random-encoder reference summarizes three encoder initializations. The comparisons control architecture and training recipe without assuming identical starting weights or stochastic trajectories across all signal conditions.

\paragraph{Shared downstream evaluation.}
Every configuration uses the retained windows and subject partitions in Table~\ref{tab:s4_data} and the three frozen readouts in Appendix~\ref{app:s4_readouts}. All stages of classifier fitting and evaluation use complete real EEG windows, without supplying an additional classical-descriptor block. Every pretraining seed is evaluated separately with this same protocol. The MAE results in the main signal comparison and the objective comparison use the same three runs per condition.

We also evaluate the published REVE-base encoder as an external frozen-transfer reference under this same downstream protocol. Its original pretraining data and computational budget are not matched to the controlled runs. It is assigned no reconstruction score in this evaluation.

\subsection{Reconstruction on real EEG}
\label{app:s4_reconstruction}

Every model with a trained signal-reconstruction head reconstructs real EEG, regardless of its pretraining signal. This includes the MAEs, REVE-small with JEPA + reconstruction, and the REVE-base MAEs. Pure JEPA has no trained signal-reconstruction head and is not assigned a reconstruction score. We use a random mask of 75\% of tokens, with the same evaluation mask for every model on a given window. Masks use seed 20260910, incremented by $1009b$ for reconstruction batch $b$, with batches of 128 windows. Predicted patches are mapped back to the signal before scoring, averaging predictions where masked patches overlap. We score only samples covered by a masked patch and by no visible patch; samples outside the extracted patches are excluded.

For each corpus, the reconstruction score is
\begin{equation}
R_{0}^{2}
=1-\frac{\sum_w\sum_{(c,t)\in M_w}(x_{wct}-\hat{x}_{wct})^2}
{\sum_w\sum_{(c,t)\in M_w}x_{wct}^{2}},
\label{eq:s4_reconstruction}
\end{equation}
where $M_w$ is the set of scored hidden samples. This compares the prediction with zero; the denominator is not centered by the mean of the hidden samples. For each pretraining run and corpus, we sum errors and target energies over windows before taking their ratio. For the overall reconstruction score, we average the six corpus scores equally within each pretraining run, then report the mean and sample standard deviation across the three pretraining runs when available. Corpus-specific entries instead summarize the three reconstruction scores for that corpus. The archive also contains the mean and standard deviation of separate per-window ratios; these are different summaries and are not substituted for the pooled score.

\subsection{Frozen-feature and supervised readouts}
\label{app:s4_readouts}

The frozen encoder receives complete real windows without masking. We compare three ways to read its token grid: averaging all tokens (\emph{mean pooling}); averaging time within each electrode and concatenating the electrode vectors (\emph{electrodes kept}); and learning one attention query over all tokens (\emph{attention pooling}). All three are followed by a linear classifier. No encoder parameter is updated during downstream fitting.

For mean pooling and electrodes kept, the classifier is ridge regression on one-hot labels. Feature scaling is fitted on training data, with constant columns removed from the effective feature count. The linear kernel is divided by the number of active features. We select $\alpha\in\{10^{-5},10^{-4},10^{-3},10^{-2},10^{-1},1,10\}$ by validation $\kappa$, using a ridge penalty of $n\alpha$ for $n$ fitting windows. We then refit scaling and the classifier on training and validation windows together. Class prediction uses the largest output.

For attention pooling, feature means and sample standard deviations are fitted over training tokens. Standard deviations at most $10^{-6}$ are replaced by one. A learned query $q$ gives token $z_i$ weight $a_i=\operatorname{softmax}_i(q^\top z_i/\sqrt{D})$; the classifier receives $\sum_i a_i z_i$. Here $D=512$ for all REVE configurations in this appendix. The query is randomly initialized, with each component drawn from $\mathcal{N}(0,1/D)$. We train the query and linear head with unweighted cross-entropy, Adam, learning rate $5\times10^{-4}$, weight decay $10^{-3}$, and batch size 128. Training lasts at most 80 epochs, with patience 10 on validation $\kappa$; an increase greater than $10^{-4}$ updates the retained state. We evaluate the best validation checkpoint without refitting on validation data. The three attention seeds are 20260908, 20260909, and 20260910.

EEGNetv4 is trained from scratch with AdamW, learning rate $10^{-3}$, weight decay $10^{-4}$, batch size 128, unweighted cross-entropy, and gradient-norm clipping at 1. Training lasts at most 100 epochs, with patience 15 on validation $\kappa$. The best validation checkpoint is used for testing. The implementation uses the EEGNetv4 defaults of eight temporal filters, depth multiplier two, 16 output filters, and a 64-sample temporal kernel. We use three seed indices $s\in\{0,1,2\}$; the seed in fold $f$ is $1000s+f$. Synthetic training windows are drawn afresh in every batch; validation and test use fixed draws for each seed and fold.

Table~\ref{tab:s4_classifier_sizes} gives the trainable parameter counts. For $C$ channels and $K$ classes, the three frozen readouts have $513K$, $512CK+K$, and $512+513K$ parameters. These counts exclude the pretrained encoder and apply to both REVE-small and REVE-base because their token widths are equal. The similar sizes of attention pooling and EEGNetv4 do not make their architectures, training procedures, or total computation equivalent.

\begin{table}[!htbp]
\centering
\small
\setlength{\tabcolsep}{4pt}
\caption{Trainable classifier parameters. Frozen-readout counts exclude the encoder; EEGNetv4 counts include the entire supervised network.}
\label{tab:s4_classifier_sizes}
\begin{tabular}{lrrrr}
\toprule
Corpus & Mean pool & Electrodes kept & Attention & EEGNetv4 \\
\midrule
BCI IV-2a & 2,052 & 45,060 & 2,564 & 3,060 \\
PhysioNet MI & 2,565 & 48,645 & 3,077 & 3,413 \\
Lee2019 ERP & 1,026 & 19,458 & 1,538 & 2,210 \\
Lee2019 SSVEP & 2,052 & 38,916 & 2,564 & 3,012 \\
ISRUC & 2,565 & 15,365 & 3,077 & 3,205 \\
TUAB & 1,026 & 19,458 & 1,538 & 2,210 \\
\bottomrule
\end{tabular}
\end{table}

\subsection{Scores and uncertainty}

We compute Cohen's $\kappa$ from the concatenated test predictions within each corpus. For leave-one-subject-out evaluation, each subject contributes predictions from its test fold. For each pretrained encoder and corpus, attention-pooling scores first average the three classifier initializations; we average scores, not predictions. The two ridge readouts yield one score per corpus and pretrained encoder.

Let $\kappa_{s,d,h}$ denote the attention score for pretraining seed $s$, corpus $d$, and classifier seed $h$. The score for one pretraining run is
\begin{equation}
K_s=\frac{1}{6}\sum_{d=1}^{6}\left(\frac{1}{3}\sum_{h=1}^{3}\kappa_{s,d,h}\right).
\end{equation}
For conditions with three pretraining seeds, we report $\bar K\pm s_K$, where
\begin{equation}
\bar K=\frac{1}{3}\sum_{s=0}^{2}K_s,
\qquad
s_K=\sqrt{\frac{1}{2}\sum_{s=0}^{2}(K_s-\bar K)^2}.
\end{equation}
Thus, the reported SD measures variation across pretraining seeds of the equal-weight six-corpus mean. It is not the SD across corpora, a standard error, or a confidence interval. For mean pooling and electrodes kept, the same aggregation applies without a classifier-seed average. Corpus-specific REVE-small entries report mean and sample SD across the three pretraining seeds for that corpus, after averaging classifier seeds where applicable. The three classifier initializations per encoder are not treated as independent pretraining runs. Random-encoder summaries use the same procedure across three encoder initializations.

For EEGNetv4, we compute a six-corpus mean separately for each supervised training seed, then report the mean and sample SD across the three supervised seeds where an SD is shown. In the task-family figure, motor imagery averages BCI IV-2a and PhysioNet MI within each seed; the other families contain one corpus each. Frozen-transfer error bars summarize the three pretraining seeds after averaging classifier seeds, while EEGNetv4 error bars summarize the three supervised training seeds. Conditions with one pretraining run have no estimated pretraining-seed SD.

For the REVE-small real--synthetic contrasts, we compute the synthetic-minus-real difference within each corresponding pretraining-run index, then report the mean and sample SD of the three differences. These SDs are computed from the differences themselves, rather than by adding or averaging the separate conditions' SDs. All contrasts use unrounded scores.

Where subject-bootstrap intervals are retained for supervised EEGNetv4 or single-run controls, they are 95\% percentile intervals from 2,000 resamples of whole test subjects, paired across conditions. Models and predictions remain fixed, and classifier-seed scores are averaged within each resample where applicable. These intervals do not measure variability across pretraining runs. The single-run REVE-base contrast uses bootstrap seed 20260920, with separate random streams for the six corpora and shared subject draws across conditions within each corpus. The two Lee2019 tasks are resampled separately. These intervals condition on the fitted models, readouts, subject partitions, and six retained corpora; they do not estimate uncertainty over a population of corpora.

\section{Construction and validation of the synthetic signals}
\label{app:surrogate_construction}

The transformations act separately on each preprocessed window, before masking. For pretraining, each window has 19 channels and $T=800$ samples at 200~Hz; the supervised controls use the channels available in each downstream corpus. Let $x_c(t)$ be channel $c$, $X_c(k)$ its discrete Fourier transform, and $\widetilde x_c(t)$ the transformed signal. A fresh draw is generated at every training presentation. The visible input and its target always come from the same draw. No filtering or normalization is applied after the transformation.

\subsection{PSD + cross-spectra}
\label{app:s4_joint}

For every positive Fourier frequency except Nyquist, we draw $\phi_k\sim\mathcal U[0,2\pi)$ and apply the same phase rotation to all channels:
\begin{equation}
\widetilde X_c(k)=X_c(k)e^{i\phi_k}.
\end{equation}
The DC and Nyquist components remain unchanged; negative frequencies follow conjugate symmetry, giving a real-valued inverse transform. For every channel pair,
\begin{equation}
\widetilde X_c(k)\widetilde X_d(k)^*=X_c(k)X_d(k)^*.
\label{eq:s4_cross_spectrum}
\end{equation}
The transformation therefore preserves the whole-window power spectra and complex cross-spectra, including relative phases between channels. It also preserves means, variances, and zero-lag covariance, up to numerical precision.

These identities concern the full four-second window. They do not preserve the original waveform, event timing, or spectra estimated within shorter intervals. Phase randomization changes how frequency components combine in time. It does not imply that every localized fluctuation disappears, that energy becomes exactly uniform in time in each draw, or that all higher-order dependencies are removed. Because Fourier amplitudes remain fixed at their source-window values, the transformed signals are not exact Gaussian-process draws. Their preserved cross-spectra also contain more detail than the finite descriptor set in Appendix~\ref{app:representations}.

\subsection{PSD}
\label{app:s4_indep}

Here the phases $\phi_{c,k}$ are drawn independently for every channel and frequency. Whole-window channel power spectra, means, and variances remain unchanged. Cross-channel phase relationships generally change, so the original complex cross-spectra and off-diagonal covariance are no longer preserved. This does not remove every cross-channel statistic: for example, the magnitude $|X_c(k)X_d(k)^*|$ remains fixed by the two preserved amplitudes. DC and Nyquist components are again unchanged.

\subsection{Covariance}
\label{app:s4_cov}

Let $\mu$ be the vector of channel means and $\Sigma=(x-\mu)(x-\mu)^\top/T$ the window covariance. With $L$ the Cholesky factor of $\Sigma+\epsilon I$, where $\epsilon=10^{-3}\operatorname{tr}(\Sigma)/C+10^{-6}$, we draw
\begin{equation}
\widetilde x(t)=\mu+L\eta(t),\qquad \eta(t)\sim\mathcal N(0,I),
\end{equation}
independently at each time point. The output is temporally white and matches the regularized covariance in expectation. Finite windows do not reproduce it exactly. In particular, this control does not preserve the original band-limited spectrum.

\subsection{Mean + variance}
\label{app:s4_white}

We draw $\widetilde x_c(t)=\mu_c+s_c\eta_c(t)$ independently across channels and time, where $s_c$ is the source window's sample standard deviation, using denominator $T-1$. The target means and variances are matched in expectation, without readjustment after sampling. Since the input is already standardized per channel, this condition is close to independent standardized white noise.

\begin{figure}[ht]
\centering
\includegraphics[width=\linewidth]{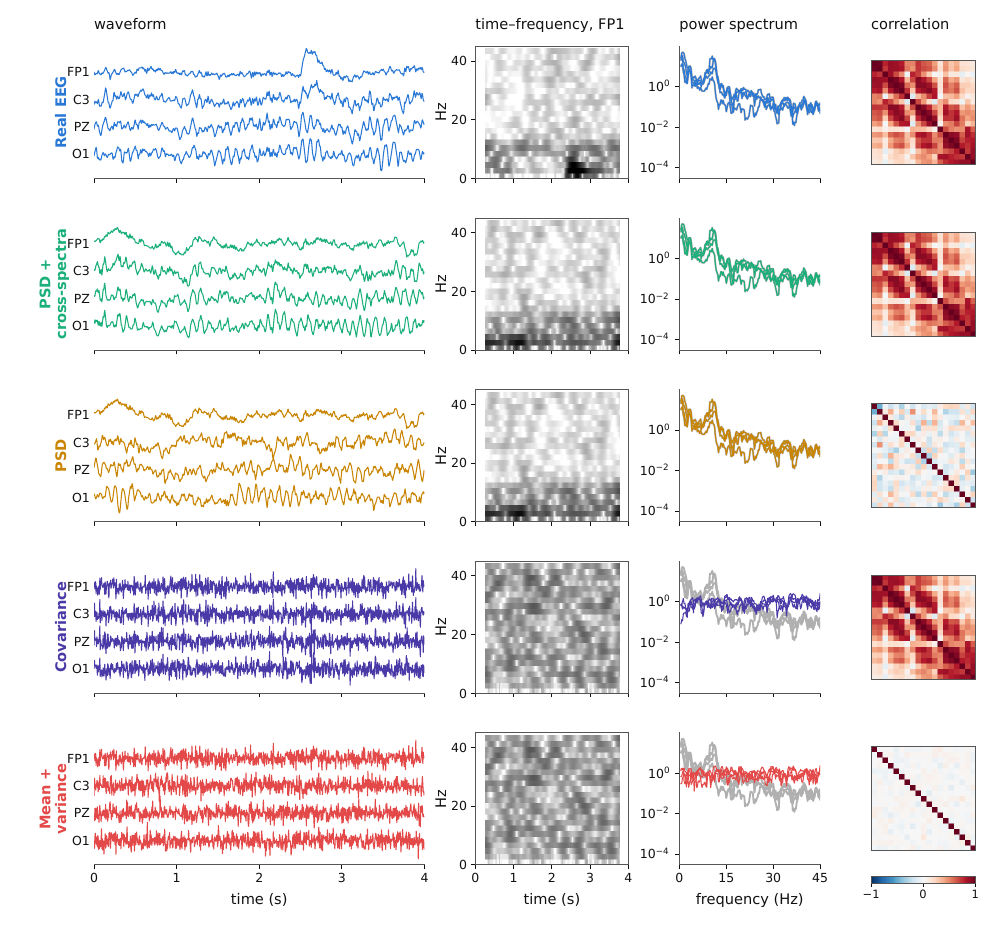}
\caption{\textbf{One real TUEG window and its four synthetic versions.} Rows show the real window and one draw of each transformation. Columns show four channels, time--frequency power on FP1 (0.5-second segments), smoothed whole-window power spectra (real reference in grey), and the 19-channel correlation matrix. The localized frontal transient is altered by phase randomization while the whole-window spectra remain unchanged. Shared phases also preserve the correlation matrix. Gaussian controls target the full covariance or only the channel variances, in expectation.}
\label{fig:synthetic_gallery}
\end{figure}

\begin{figure}[ht]
\centering
\includegraphics[width=\linewidth]{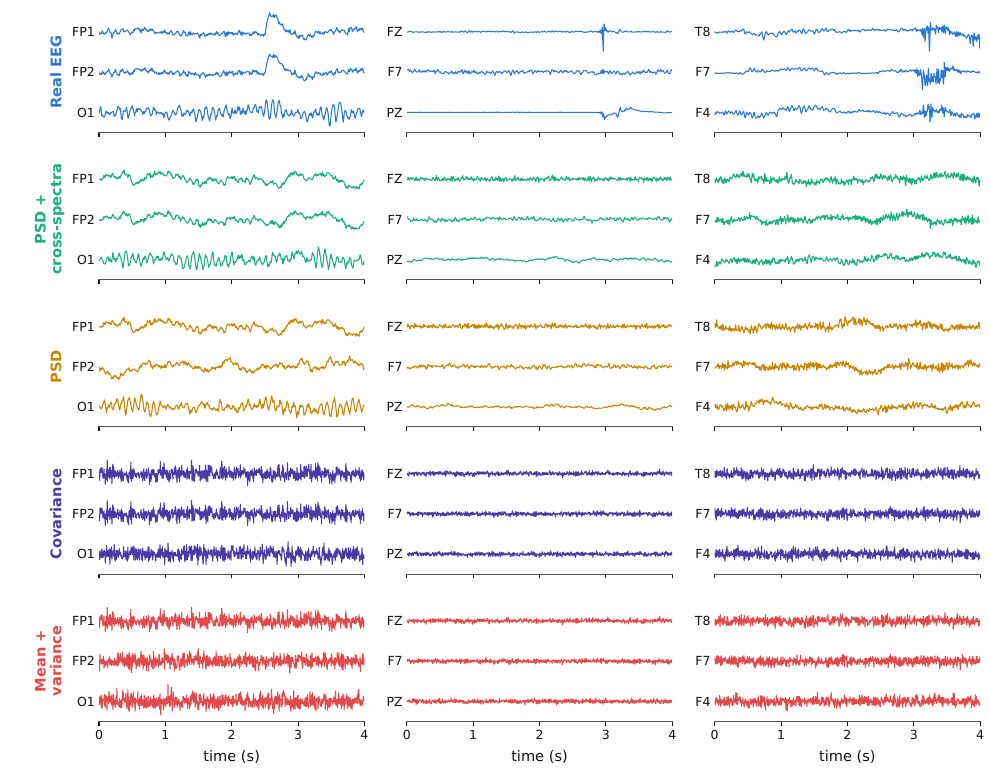}
\caption{\textbf{Examples of changes in waveform and temporal localization.} Columns show a frontal blink-like transient, a sharp transient on FZ (both TUEG), and a burst of high-frequency activity on T8 and F7 (mixed evaluation set). Three channels are displayed per window. Rows show the real signal and one draw of each synthetic version, using the colours of Figure~\ref{fig:synthetic_gallery}. These examples illustrate altered event morphology and timing; they are not a quantitative test that every event or physiological feature is removed.}
\label{fig:synthetic_waveforms}
\end{figure}

\subsection{Numerical preservation checks}

\begin{figure}[h]
\centering
\includegraphics[width=\linewidth]{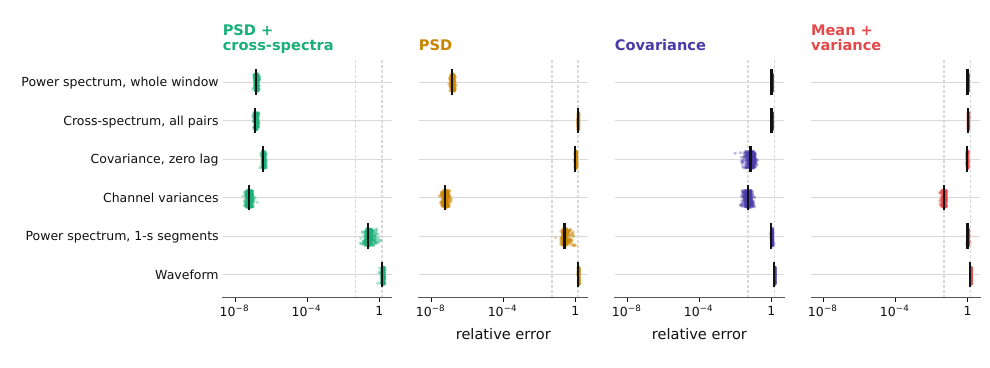}
\caption{\textbf{Per-window preservation errors.} Each point compares a synthetic window with its source (128 windows from each of three sets). Black bars summarize the pooled errors. The reference $\sqrt{2/T}\approx0.05$ gives the scale of relative sampling fluctuations of a Gaussian variance estimate; it is not a general threshold for covariance-matrix error. The reference $\sqrt{2}$ is the relative distance between orthogonal signals of equal energy. The small Fourier errors are consistent with the exact identities in Equation~\ref{eq:s4_cross_spectrum} evaluated at finite precision.}
\label{fig:synthetic_errors}
\end{figure}

We apply the pretraining transformations to 128 archived windows from each of three evaluation sets (TUEG, Stieger, and a mixed set), all with 19 channels and 800 samples. For statistic $s$, the relative error is $\|s(\widetilde x)-s(x)\|_F/\|s(x)\|_F$. Table~\ref{tab:s4_invariants_all} first pools squared norms over the windows of each set, then averages the three set-level errors. Figure~\ref{fig:synthetic_errors} also shows the per-window ratios; a pooled ratio need not equal their mean. Distances between different real windows provide a descriptive reference.

\begin{table}[!htbp]
\centering
\footnotesize
\setlength{\tabcolsep}{3pt}
\caption{Numerical preservation checks. Entries are relative Frobenius errors pooled within each evaluation set and averaged over the three sets. Fourier products use the complete four-second window; Welch uses one-second segments. The last row compares two different real windows from the same set.}
\label{tab:s4_invariants_all}
\begin{tabular}{lrrrrrr}
\toprule
Signal version & Power spec. & Cross-spec. & Covariance & Variances & Welch 1 s & Waveform \\
\midrule
PSD + cross-spectra & $1.4\times10^{-7}$ & $1.3\times10^{-7}$ & $3.5\times10^{-7}$ & $5.8\times10^{-8}$ & 0.246 & 1.425 \\
PSD & $1.4\times10^{-7}$ & 1.359 & 0.975 & $5.9\times10^{-8}$ & 0.250 & 1.414 \\
Covariance & 1.000 & 1.011 & 0.069 & 0.050 & 0.984 & 1.415 \\
Mean + variance & 1.000 & 1.015 & 0.926 & 0.050 & 0.983 & 1.415 \\
\midrule
\textit{Two real windows} & 1.059 & 1.288 & 0.779 & 0.013 & 0.813 & 1.408 \\
\bottomrule
\end{tabular}
\end{table}

The set-level errors for exactly preserved Fourier statistics are below $4\times10^{-7}$. The covariance control has relative covariance error 0.069, and the Gaussian variance errors are approximately 0.050, reflecting finite-sample variation and, for covariance, regularization. Independent phases substantially change the complex cross-spectra and zero-lag covariance; Gaussian inputs substantially change the spectra. Waveform distances are approximately 1.41 in all four conditions.

One-second Welch spectra are not exactly preserved by either phase transformation (relative errors 0.246 and 0.250). The identity applies to the full-window Fourier products, while Welch estimation uses shorter, tapered segments. Similarly, filtering a finite window can introduce boundary effects in band-limited descriptors.
\section{Additional results for Section~\ref{sec:surrogates}}
\label{app:surrogate_results}

\subsection{Results by corpus and readout}
\label{app:s4_corpus_results}

Tables~\ref{tab:s4_results_mi}--\ref{tab:s4_results_sleep_clinical} give the signal conditions, individual initialization references A and B, and extended-budget checkpoints for the six corpora. At the standard budget, the Real EEG, PSD + cross-spectra, Covariance, and PSD MAE entries summarize three REVE-small pretraining seeds. The raw-linear reference applies the same ridge procedure to the flattened real window, without an encoder. EEGNetv4 entries use the signal version indicated by the row for both training and evaluation. All MAE entries use real EEG for evaluation. All frozen-readout results use the protocol of Appendix~\ref{app:s4_readouts}. Appendix~\ref{app:s4_objective_controls} compares the pretraining objectives and the larger REVE encoder.

\begin{table}[!htbp]
\centering
\footnotesize
\setlength{\tabcolsep}{3pt}
\caption{Motor imagery. Reconstruction is pooled $R_0^2$ on real EEG; the three frozen-readout columns report $\kappa$ on real EEG. For the four standard-budget conditions Real EEG, PSD + cross-spectra, Covariance, and PSD, these entries give mean $\pm$ sample SD across three REVE-small pretraining seeds. Attention first averages three classifier seeds within each pretraining run and corpus. Mean + variance, each individual initialization reference, and each $5\times$ condition use one encoder run and are reported as point estimates. EEGNetv4 entries remain means across three supervised training seeds on the row's signal version. Dashes denote inapplicable evaluations, not zero scores.}
\label{tab:s4_results_mi}
\begin{tabular}{lrrrrr}
\toprule
Condition & $R_0^2$ & Mean pool & Electrodes kept & Attention & EEGNetv4 \\
\midrule
\multicolumn{6}{l}{\textit{BCI IV-2a}} \\
Real EEG & \shortstack{$0.890$\\$\pm 0.001$} & \shortstack{$0.155$\\$\pm 0.008$} & \shortstack{$0.305$\\$\pm 0.009$} & \shortstack{$0.111$\\$\pm 0.012$} & 0.440 \\
PSD + cross-spectra & \shortstack{$0.890$\\$\pm 0.001$} & \shortstack{$0.160$\\$\pm 0.012$} & \shortstack{$0.296$\\$\pm 0.011$} & \shortstack{$0.134$\\$\pm 0.023$} & 0.168 \\
Covariance & \shortstack{$0.874$\\$\pm 0.004$} & \shortstack{$0.139$\\$\pm 0.011$} & \shortstack{$0.244$\\$\pm 0.004$} & \shortstack{$0.144$\\$\pm 0.006$} & 0.089 \\
PSD & \shortstack{$0.058$\\$\pm 0.000$} & \shortstack{$0.094$\\$\pm 0.006$} & \shortstack{$0.242$\\$\pm 0.008$} & \shortstack{$0.082$\\$\pm 0.006$} & 0.037 \\
Mean + variance & 0.001 & 0.036 & 0.126 & 0.063 & 0.000 \\
\addlinespace[2pt]
Initialization A & -0.315 & 0.111 & 0.181 & 0.111 & --- \\
Initialization B & -0.325 & 0.105 & 0.199 & 0.105 & --- \\
\addlinespace[2pt]
Real EEG, $5\times$ & 0.887 & 0.169 & 0.320 & 0.129 & --- \\
PSD + cross-spectra, $5\times$ & 0.888 & 0.171 & 0.298 & 0.160 & --- \\
\addlinespace[2pt]
Raw linear & --- & \multicolumn{3}{c}{0.270 (raw-window ridge)} & --- \\
\midrule
\multicolumn{6}{l}{\textit{PhysioNet MI}} \\
Real EEG & \shortstack{$0.715$\\$\pm 0.001$} & \shortstack{$0.162$\\$\pm 0.031$} & \shortstack{$0.267$\\$\pm 0.016$} & \shortstack{$0.124$\\$\pm 0.002$} & 0.395 \\
PSD + cross-spectra & \shortstack{$0.707$\\$\pm 0.000$} & \shortstack{$0.135$\\$\pm 0.012$} & \shortstack{$0.240$\\$\pm 0.007$} & \shortstack{$0.115$\\$\pm 0.003$} & 0.075 \\
Covariance & \shortstack{$0.627$\\$\pm 0.003$} & \shortstack{$0.151$\\$\pm 0.014$} & \shortstack{$0.221$\\$\pm 0.017$} & \shortstack{$0.095$\\$\pm 0.006$} & 0.075 \\
PSD & \shortstack{$0.062$\\$\pm 0.001$} & \shortstack{$0.105$\\$\pm 0.020$} & \shortstack{$0.267$\\$\pm 0.009$} & \shortstack{$0.113$\\$\pm 0.003$} & 0.040 \\
Mean + variance & 0.000 & 0.037 & 0.147 & 0.042 & 0.011 \\
\addlinespace[2pt]
Initialization A & -0.319 & 0.135 & 0.207 & 0.032 & --- \\
Initialization B & -0.344 & 0.131 & 0.214 & 0.035 & --- \\
\addlinespace[2pt]
Real EEG, $5\times$ & 0.721 & 0.148 & 0.273 & 0.135 & --- \\
PSD + cross-spectra, $5\times$ & 0.712 & 0.187 & 0.242 & 0.127 & --- \\
\addlinespace[2pt]
Raw linear & --- & \multicolumn{3}{c}{0.272 (raw-window ridge)} & --- \\
\bottomrule
\end{tabular}
\end{table}

\begin{table}[!htbp]
\centering
\footnotesize
\setlength{\tabcolsep}{3pt}
\caption{ERP and SSVEP. Reconstruction is pooled $R_0^2$ on real EEG; the three frozen-readout columns report $\kappa$ on real EEG. For the four standard-budget conditions Real EEG, PSD + cross-spectra, Covariance, and PSD, these entries give mean $\pm$ sample SD across three REVE-small pretraining seeds. Attention first averages three classifier seeds within each pretraining run and corpus. Mean + variance, each individual initialization reference, and each $5\times$ condition use one encoder run and are reported as point estimates. EEGNetv4 entries remain means across three supervised training seeds on the row's signal version. Dashes denote inapplicable evaluations, not zero scores.}
\label{tab:s4_results_erp_ssvep}
\begin{tabular}{lrrrrr}
\toprule
Condition & $R_0^2$ & Mean pool & Electrodes kept & Attention & EEGNetv4 \\
\midrule
\multicolumn{6}{l}{\textit{Lee2019 ERP}} \\
Real EEG & \shortstack{$0.726$\\$\pm 0.000$} & \shortstack{$0.003$\\$\pm 0.010$} & \shortstack{$0.037$\\$\pm 0.017$} & \shortstack{$-0.013$\\$\pm 0.006$} & 0.341 \\
PSD + cross-spectra & \shortstack{$0.712$\\$\pm 0.002$} & \shortstack{$0.005$\\$\pm 0.009$} & \shortstack{$0.032$\\$\pm 0.023$} & \shortstack{$-0.005$\\$\pm 0.006$} & 0.000 \\
Covariance & \shortstack{$0.608$\\$\pm 0.012$} & \shortstack{$0.002$\\$\pm 0.013$} & \shortstack{$0.073$\\$\pm 0.047$} & \shortstack{$0.008$\\$\pm 0.014$} & 0.003 \\
PSD & \shortstack{$0.061$\\$\pm 0.001$} & \shortstack{$0.012$\\$\pm 0.003$} & \shortstack{$0.092$\\$\pm 0.031$} & \shortstack{$-0.003$\\$\pm 0.014$} & 0.000 \\
Mean + variance & 0.000 & 0.000 & 0.068 & 0.020 & 0.000 \\
\addlinespace[2pt]
Initialization A & -0.328 & -0.002 & 0.081 & 0.000 & --- \\
Initialization B & -0.349 & 0.000 & 0.099 & 0.012 & --- \\
\addlinespace[2pt]
Real EEG, $5\times$ & 0.733 & 0.033 & 0.130 & 0.003 & --- \\
PSD + cross-spectra, $5\times$ & 0.717 & -0.007 & 0.035 & -0.011 & --- \\
\addlinespace[2pt]
\midrule
\multicolumn{6}{l}{\textit{Lee2019 SSVEP}} \\
Real EEG & \shortstack{$0.729$\\$\pm 0.000$} & \shortstack{$0.416$\\$\pm 0.007$} & \shortstack{$0.733$\\$\pm 0.005$} & \shortstack{$0.724$\\$\pm 0.019$} & 0.894 \\
PSD + cross-spectra & \shortstack{$0.714$\\$\pm 0.001$} & \shortstack{$0.425$\\$\pm 0.015$} & \shortstack{$0.723$\\$\pm 0.007$} & \shortstack{$0.721$\\$\pm 0.010$} & 0.882 \\
Covariance & \shortstack{$0.628$\\$\pm 0.008$} & \shortstack{$0.181$\\$\pm 0.020$} & \shortstack{$0.633$\\$\pm 0.005$} & \shortstack{$0.652$\\$\pm 0.002$} & 0.086 \\
PSD & \shortstack{$0.051$\\$\pm 0.001$} & \shortstack{$0.336$\\$\pm 0.024$} & \shortstack{$0.658$\\$\pm 0.005$} & \shortstack{$0.580$\\$\pm 0.016$} & 0.658 \\
Mean + variance & 0.000 & 0.111 & 0.624 & 0.386 & 0.000 \\
\addlinespace[2pt]
Initialization A & -0.343 & 0.337 & 0.651 & 0.454 & --- \\
Initialization B & -0.361 & 0.312 & 0.666 & 0.473 & --- \\
\addlinespace[2pt]
Real EEG, $5\times$ & 0.737 & 0.475 & 0.764 & 0.765 & --- \\
PSD + cross-spectra, $5\times$ & 0.719 & 0.487 & 0.765 & 0.747 & --- \\
\addlinespace[2pt]
\bottomrule
\end{tabular}
\end{table}

\begin{table}[!htbp]
\centering
\footnotesize
\setlength{\tabcolsep}{3pt}
\caption{Sleep and clinical EEG. Reconstruction is pooled $R_0^2$ on real EEG; the three frozen-readout columns report $\kappa$ on real EEG. For the four standard-budget conditions Real EEG, PSD + cross-spectra, Covariance, and PSD, these entries give mean $\pm$ sample SD across three REVE-small pretraining seeds. Attention first averages three classifier seeds within each pretraining run and corpus. Mean + variance, each individual initialization reference, and each $5\times$ condition use one encoder run and are reported as point estimates. EEGNetv4 entries remain means across three supervised training seeds on the row's signal version. Dashes denote inapplicable evaluations, not zero scores.}
\label{tab:s4_results_sleep_clinical}
\begin{tabular}{lrrrrr}
\toprule
Condition & $R_0^2$ & Mean pool & Electrodes kept & Attention & EEGNetv4 \\
\midrule
\multicolumn{6}{l}{\textit{ISRUC}} \\
Real EEG & \shortstack{$0.215$\\$\pm 0.014$} & \shortstack{$0.538$\\$\pm 0.003$} & \shortstack{$0.535$\\$\pm 0.005$} & \shortstack{$0.541$\\$\pm 0.005$} & 0.537 \\
PSD + cross-spectra & \shortstack{$0.207$\\$\pm 0.014$} & \shortstack{$0.544$\\$\pm 0.006$} & \shortstack{$0.546$\\$\pm 0.004$} & \shortstack{$0.541$\\$\pm 0.005$} & 0.554 \\
Covariance & \shortstack{$0.115$\\$\pm 0.017$} & \shortstack{$0.274$\\$\pm 0.005$} & \shortstack{$0.278$\\$\pm 0.005$} & \shortstack{$0.278$\\$\pm 0.003$} & 0.162 \\
PSD & \shortstack{$0.024$\\$\pm 0.003$} & \shortstack{$0.473$\\$\pm 0.008$} & \shortstack{$0.471$\\$\pm 0.006$} & \shortstack{$0.460$\\$\pm 0.002$} & 0.507 \\
Mean + variance & 0.000 & 0.100 & 0.090 & 0.167 & 0.003 \\
\addlinespace[2pt]
Initialization A & -0.319 & 0.321 & 0.275 & 0.090 & --- \\
Initialization B & -0.342 & 0.307 & 0.273 & 0.045 & --- \\
\addlinespace[2pt]
Real EEG, $5\times$ & 0.192 & 0.547 & 0.545 & 0.561 & --- \\
PSD + cross-spectra, $5\times$ & 0.177 & 0.579 & 0.576 & 0.567 & --- \\
\addlinespace[2pt]
Raw linear & --- & \multicolumn{3}{c}{0.026 (raw-window ridge)} & --- \\
\midrule
\multicolumn{6}{l}{\textit{TUAB}} \\
Real EEG & \shortstack{$0.664$\\$\pm 0.002$} & \shortstack{$0.597$\\$\pm 0.006$} & \shortstack{$0.610$\\$\pm 0.008$} & \shortstack{$0.593$\\$\pm 0.008$} & 0.532 \\
PSD + cross-spectra & \shortstack{$0.659$\\$\pm 0.002$} & \shortstack{$0.599$\\$\pm 0.004$} & \shortstack{$0.597$\\$\pm 0.007$} & \shortstack{$0.587$\\$\pm 0.007$} & 0.561 \\
Covariance & \shortstack{$0.565$\\$\pm 0.004$} & \shortstack{$0.392$\\$\pm 0.004$} & \shortstack{$0.420$\\$\pm 0.013$} & \shortstack{$0.396$\\$\pm 0.010$} & 0.297 \\
PSD & \shortstack{$0.062$\\$\pm 0.000$} & \shortstack{$0.477$\\$\pm 0.008$} & \shortstack{$0.489$\\$\pm 0.008$} & \shortstack{$0.469$\\$\pm 0.006$} & 0.460 \\
Mean + variance & 0.000 & 0.199 & 0.085 & 0.247 & -0.017 \\
\addlinespace[2pt]
Initialization A & -0.322 & 0.331 & 0.191 & 0.191 & --- \\
Initialization B & -0.344 & 0.344 & 0.226 & 0.235 & --- \\
\addlinespace[2pt]
Real EEG, $5\times$ & 0.675 & 0.640 & 0.626 & 0.632 & --- \\
PSD + cross-spectra, $5\times$ & 0.669 & 0.620 & 0.635 & 0.615 & --- \\
\addlinespace[2pt]
Raw linear & --- & \multicolumn{3}{c}{-0.008 (raw-window ridge)} & --- \\
\bottomrule
\end{tabular}
\end{table}

Changing the readout changes absolute performance, but the average comparison between real EEG and PSD + cross-spectra remains close. At the standard budget, their mean-pooling scores are 0.312 and 0.311; with electrodes kept, they are 0.415 and 0.406; with attention, they are 0.347 and 0.349. Each score averages three pretraining seeds and weights the six corpora equally. Electrode-wise pooling improves the mean score while also increasing the number of classifier parameters. Its gain cannot be assigned solely to either the retained structure or the larger classifier from this comparison alone.

The random-encoder reference over three encoder initializations reaches $0.146\pm 0.008$ in mean attention $\kappa$. Both real-EEG and PSD + cross-spectra pretraining remain well above this descriptive reference. The individual initialization references A and B are also reported by corpus; these reference comparisons do not assume identical starting weights across all signal conditions.

There are meaningful differences within this overall pattern. With electrodes kept, real EEG exceeds PSD + cross-spectra by 0.028 on PhysioNet MI, whereas the sign reverses under some other corpus--readout combinations. Mean-pooled real and PSD + cross-spectra features give near-zero scores on Lee2019 ERP; other readouts recover modest scores. These results limit what this probe demonstrates about ERP transfer. They do not establish that the embeddings contain no ERP information.

For supervised EEGNetv4, changing real EEG to PSD + cross-spectra reduces $\kappa$ from 0.440 to 0.168 on BCI IV-2a, from 0.395 to 0.075 on PhysioNet MI, and from 0.341 to 0.000 on Lee2019 ERP. The corresponding real-minus-synthetic subject-bootstrap intervals are $[0.229,0.316]$, $[0.271,0.372]$, and $[0.266,0.416]$. SSVEP changes only from 0.894 to 0.882; ISRUC and TUAB are slightly higher on the transformed signal. The supervised penalty is therefore concentrated in motor imagery and ERP under this protocol.

\subsection{The signal comparison across pretraining objectives}
\label{app:s4_objective_controls}

Table~\ref{tab:s4_controls_scores} reports the four-signal grid for REVE-small under MAE, JEPA, and JEPA + reconstruction, with three pretraining seeds per condition, together with the single-run REVE-base MAE size control. The objective comparison retains the same REVE-small encoder architecture, while the size control retains the MAE objective. Both use the data and evaluation procedure of the main experiment (Appendix~\ref{app:s4_objective_setup}). Table~\ref{tab:s4_controls_contrasts} isolates the effect of replacing real EEG with PSD + cross-spectra at each fixed encoder size, objective, and readout.

With attention pooling, real-EEG and PSD + cross-spectra pretraining give 0.347 and 0.349 under MAE, 0.354 and 0.349 under pure JEPA, and 0.352 and 0.349 under JEPA + reconstruction, averaging three pretraining seeds per condition. The real--synthetic proximity therefore persists when the target is a learned latent representation and no raw-sample reconstruction loss is used. It also persists when latent prediction and reconstruction are combined. Under REVE-base MAE, the corresponding scores are 0.368 and 0.364, extending the observation to a deeper encoder.

Across the three REVE-small objectives and all three readouts, the synthetic-minus-real differences in six-corpus mean $\kappa$ range from $-0.013$ to $+0.002$ after averaging pretraining seeds (computed before rounding). Individual run differences range from $-0.024$ to $+0.010$. The REVE-base size control also gives small mean differences. These results support a close average comparison across the tested objectives and readouts, rather than dependence on masked raw-signal reconstruction alone; they are not a formal equivalence test.

\begin{table}[!htbp]
\centering
\footnotesize
\setlength{\tabcolsep}{3pt}
\caption{Objective and encoder-size controls. All transfer scores use real EEG and the readouts of Appendix~\ref{app:s4_readouts}. Each run score averages six corpora equally; attention first averages three classifier seeds within each corpus and pretraining run. REVE-small entries give mean $\pm$ sample SD across three pretraining seeds. $R_0^2$ is the pooled reconstruction score within each corpus, averaged across corpora within each run. REVE-base entries are point estimates from one pretraining run. Dashes denote no evaluated trained reconstruction head. The REVE-small random reference summarizes three encoder initializations; the REVE-base random reference uses one.}
\label{tab:s4_controls_scores}
\begin{tabular}{llrrrr}
\toprule
Encoder / objective & Pretraining signal & $R_0^2$ & Mean pool & Electrodes kept & Attention \\
\midrule
REVE-small / MAE & Real EEG & \shortstack{$0.657$\\$\pm 0.003$} & \shortstack{$0.312$\\$\pm 0.005$} & \shortstack{$0.415$\\$\pm 0.002$} & \shortstack{$0.347$\\$\pm 0.005$} \\
 & PSD + cross-spectra & \shortstack{$0.648$\\$\pm 0.003$} & \shortstack{$0.311$\\$\pm 0.007$} & \shortstack{$0.406$\\$\pm 0.003$} & \shortstack{$0.349$\\$\pm 0.005$} \\
 & Covariance & \shortstack{$0.570$\\$\pm 0.008$} & \shortstack{$0.190$\\$\pm 0.005$} & \shortstack{$0.311$\\$\pm 0.008$} & \shortstack{$0.262$\\$\pm 0.006$} \\
 & PSD & \shortstack{$0.053$\\$\pm 0.001$} & \shortstack{$0.250$\\$\pm 0.002$} & \shortstack{$0.370$\\$\pm 0.008$} & \shortstack{$0.283$\\$\pm 0.006$} \\
\addlinespace[3pt]
REVE-small / JEPA & Real EEG & --- & \shortstack{$0.323$\\$\pm 0.002$} & \shortstack{$0.423$\\$\pm 0.010$} & \shortstack{$0.354$\\$\pm 0.002$} \\
 & PSD + cross-spectra & --- & \shortstack{$0.311$\\$\pm 0.007$} & \shortstack{$0.413$\\$\pm 0.007$} & \shortstack{$0.349$\\$\pm 0.003$} \\
 & Covariance & --- & \shortstack{$0.199$\\$\pm 0.003$} & \shortstack{$0.312$\\$\pm 0.006$} & \shortstack{$0.266$\\$\pm 0.004$} \\
 & PSD & --- & \shortstack{$0.270$\\$\pm 0.003$} & \shortstack{$0.375$\\$\pm 0.005$} & \shortstack{$0.307$\\$\pm 0.004$} \\
\addlinespace[3pt]
REVE-small / JEPA + recon. & Real EEG & \shortstack{$0.655$\\$\pm 0.001$} & \shortstack{$0.317$\\$\pm 0.010$} & \shortstack{$0.417$\\$\pm 0.001$} & \shortstack{$0.352$\\$\pm 0.004$} \\
 & PSD + cross-spectra & \shortstack{$0.647$\\$\pm 0.000$} & \shortstack{$0.319$\\$\pm 0.006$} & \shortstack{$0.412$\\$\pm 0.003$} & \shortstack{$0.349$\\$\pm 0.003$} \\
 & Covariance & \shortstack{$0.574$\\$\pm 0.006$} & \shortstack{$0.194$\\$\pm 0.007$} & \shortstack{$0.314$\\$\pm 0.008$} & \shortstack{$0.270$\\$\pm 0.005$} \\
 & PSD & \shortstack{$0.055$\\$\pm 0.001$} & \shortstack{$0.256$\\$\pm 0.005$} & \shortstack{$0.381$\\$\pm 0.011$} & \shortstack{$0.293$\\$\pm 0.003$} \\
\addlinespace[3pt]
REVE-base / MAE & Real EEG & 0.661 & 0.327 & 0.424 & 0.368 \\
 & PSD + cross-spectra & 0.653 & 0.330 & 0.420 & 0.364 \\
 & Covariance & 0.574 & 0.193 & 0.300 & 0.265 \\
 & PSD & 0.053 & 0.252 & 0.369 & 0.286 \\
\addlinespace[3pt]
\midrule
\multicolumn{6}{l}{\textit{Random encoder references}} \\
REVE-small & --- & --- & \shortstack{$0.203$\\$\pm 0.005$} & \shortstack{$0.266$\\$\pm 0.003$} & \shortstack{$0.146$\\$\pm 0.008$} \\
REVE-base & --- & --- & 0.200 & 0.297 & 0.181 \\
\bottomrule
\end{tabular}
\end{table}

\begin{table}[!htbp]
\centering
\footnotesize
\setlength{\tabcolsep}{3pt}
\caption{PSD + cross-spectra minus real-EEG pretraining at fixed encoder size, objective, and readout. Differences use unrounded scores and average the six corpora equally. REVE-small entries report mean $\pm$ sample SD of the three differences computed within corresponding pretraining-run indices. For the single-run REVE-base control, brackets retain 95\% intervals from 2,000 paired test-subject bootstrap resamples, with independent resampling across corpora; those intervals exclude pretraining-run uncertainty.}
\label{tab:s4_controls_contrasts}
\begin{tabular}{llrrr}
\toprule
Encoder & Objective & Mean pool & Electrodes kept & Attention \\
\midrule
REVE-small & MAE & \shortstack{$-0.001$\\$\pm 0.005$} & \shortstack{$-0.009$\\$\pm 0.005$} & \shortstack{$+0.002$\\$\pm 0.008$} \\
\addlinespace[3pt]
REVE-small & JEPA & \shortstack{$-0.013$\\$\pm 0.005$} & \shortstack{$-0.010$\\$\pm 0.017$} & \shortstack{$-0.005$\\$\pm 0.005$} \\
\addlinespace[3pt]
REVE-small & JEPA + recon. & \shortstack{$+0.001$\\$\pm 0.013$} & \shortstack{$-0.005$\\$\pm 0.002$} & \shortstack{$-0.003$\\$\pm 0.004$} \\
\addlinespace[3pt]
REVE-base & MAE & \shortstack{$+0.003$\\$[-0.008,+0.015]$} & \shortstack{$-0.004$\\$[-0.013,+0.005]$} & \shortstack{$-0.003$\\$[-0.012,+0.007]$} \\
\addlinespace[3pt]
\bottomrule
\end{tabular}
\end{table}

The reduced-statistic controls give a consistent separation across objectives. For REVE-small, preserving covariance alone or channel spectra alone yields lower six-corpus mean transfer than real EEG in all three pretraining runs under every objective and readout. With attention, mean deficits relative to real EEG are 0.082--0.088 for Covariance and 0.047--0.063 for PSD. REVE-base shows the same ordering. Thus, the experiment shows both the proximity of real EEG to PSD + cross-spectra and the lower average transfer obtained when preserving only one of these statistical components.

For configurations with a reconstruction head, real-EEG and PSD + cross-spectra pretraining also give close pooled reconstruction scores on real downstream EEG: 0.657 versus 0.648 for REVE-small MAE, 0.655 versus 0.647 for JEPA + reconstruction, and 0.661 versus 0.653 for REVE-base MAE. The REVE-small values average three pretraining seeds. These comparisons use the same 75\% mask and patch geometry. Pure JEPA is evaluated through transfer, without attributing a signal-reconstruction score to its latent predictor.

\paragraph{Results by corpus.}
Table~\ref{tab:s4_controls_corpora} gives the attention scores underlying the real--synthetic comparison. For all four encoder--objective configurations, the absolute difference in mean score is below 0.025 on every corpus; REVE-small means average three pretraining seeds. Thus, the six-corpus means do not conceal a large mean attention-pooling gap on one task. Corpus differences can be larger with other readouts: for example, under JEPA the PSD + cross-spectra condition is 0.029 below real EEG on PhysioNet MI when electrodes are kept. The results support a close average comparison, with some dependence on corpus and readout.

Attention scores on Lee2019 ERP remain close to zero under all four configurations. Similarity in that regime provides little evidence about preserving useful ERP transfer, and does not imply that the embeddings contain no ERP information. By contrast, JEPA achieves substantial transfer on SSVEP, ISRUC, and TUAB under both real and PSD + cross-spectra pretraining. The agreement between signal conditions is therefore not simply an agreement between uninformative encoders on every task.

\begin{table}[!htbp]
\centering
\scriptsize
\setlength{\tabcolsep}{3pt}
\caption{Attention-pooling transfer by corpus. Synthetic denotes PSD + cross-spectra pretraining; all readouts are fitted and tested on real EEG. REVE-small entries report mean $\pm$ sample SD across three pretraining seeds, after averaging three classifier seeds within each pretraining run and corpus. REVE-base entries average three classifier seeds for one pretraining run. BCI and PhysioNet denote the two motor-imagery corpora; ERP and SSVEP denote the Lee2019 tasks.}
\label{tab:s4_controls_corpora}
\begin{tabular}{llrrrrrr}
\toprule
Encoder / objective & Signal & BCI & PhysioNet & ERP & SSVEP & ISRUC & TUAB \\
\midrule
REVE-small / MAE & Real & \shortstack{$0.111$\\$\pm 0.012$} & \shortstack{$0.124$\\$\pm 0.002$} & \shortstack{$-0.013$\\$\pm 0.006$} & \shortstack{$0.724$\\$\pm 0.019$} & \shortstack{$0.541$\\$\pm 0.005$} & \shortstack{$0.593$\\$\pm 0.008$} \\
 & Synthetic & \shortstack{$0.134$\\$\pm 0.023$} & \shortstack{$0.115$\\$\pm 0.003$} & \shortstack{$-0.005$\\$\pm 0.006$} & \shortstack{$0.721$\\$\pm 0.010$} & \shortstack{$0.541$\\$\pm 0.005$} & \shortstack{$0.587$\\$\pm 0.007$} \\
\addlinespace[3pt]
REVE-small / JEPA & Real & \shortstack{$0.089$\\$\pm 0.008$} & \shortstack{$0.127$\\$\pm 0.010$} & \shortstack{$-0.001$\\$\pm 0.007$} & \shortstack{$0.725$\\$\pm 0.014$} & \shortstack{$0.582$\\$\pm 0.009$} & \shortstack{$0.604$\\$\pm 0.006$} \\
 & Synthetic & \shortstack{$0.099$\\$\pm 0.021$} & \shortstack{$0.127$\\$\pm 0.007$} & \shortstack{$-0.005$\\$\pm 0.011$} & \shortstack{$0.708$\\$\pm 0.024$} & \shortstack{$0.577$\\$\pm 0.004$} & \shortstack{$0.590$\\$\pm 0.008$} \\
\addlinespace[3pt]
REVE-small / JEPA + recon. & Real & \shortstack{$0.113$\\$\pm 0.027$} & \shortstack{$0.125$\\$\pm 0.004$} & \shortstack{$-0.003$\\$\pm 0.013$} & \shortstack{$0.713$\\$\pm 0.008$} & \shortstack{$0.557$\\$\pm 0.012$} & \shortstack{$0.607$\\$\pm 0.007$} \\
 & Synthetic & \shortstack{$0.117$\\$\pm 0.017$} & \shortstack{$0.122$\\$\pm 0.007$} & \shortstack{$0.000$\\$\pm 0.008$} & \shortstack{$0.725$\\$\pm 0.010$} & \shortstack{$0.549$\\$\pm 0.006$} & \shortstack{$0.584$\\$\pm 0.007$} \\
\addlinespace[3pt]
REVE-base / MAE & Real & 0.134 & 0.135 & 0.004 & 0.753 & 0.552 & 0.627 \\
 & Synthetic & 0.143 & 0.132 & -0.005 & 0.743 & 0.569 & 0.603 \\
\addlinespace[3pt]
\bottomrule
\end{tabular}
\end{table}

\paragraph{Published-checkpoint reference.}
\label{app:surrogate-official-references}
As an external reference point, we also evaluate the published REVE-base checkpoint under the same downstream protocol. It reaches a mean attention transfer of 0.347 (Table~\ref{tab:s4_official_references}), close to the 0.368 of our real-EEG MAE and the 0.364 of our PSD + cross-spectra MAE at the same encoder size. Our controlled runs therefore fall in the same range of average downstream performance as the public model. Note, however, that the published checkpoint was pretrained on different data with a different compute budget, so this comparison is a sanity check on absolute performance, not evidence of matched pretraining efficiency.

\begin{table}[!htbp]
\centering
\small
\setlength{\tabcolsep}{3pt}
\caption{Published REVE-base and the controlled REVE-base MAEs under the same frozen-readout protocol on real EEG. Values are mean $\kappa$ across six corpora; attention averages three classifier seeds. Pretraining data and compute are not matched to the published checkpoint.}
\label{tab:s4_official_references}
\begin{tabular}{lrrr}
\toprule
REVE-base pretraining & Mean pool & Electrodes kept & Attention \\
\midrule
Published checkpoint & 0.299 & 0.418 & 0.347 \\
Real EEG & 0.327 & 0.424 & 0.368 \\
PSD + cross-spectra & 0.330 & 0.420 & 0.364 \\
\bottomrule
\end{tabular}
\end{table}

\subsection{Five times the pretraining budget}
\label{app:s4_budget}

\begin{table}[!htbp]
\centering
\footnotesize
\setlength{\tabcolsep}{3pt}
\caption{Effect of five times the pretraining budget. Reconstruction uses pooled $R_0^2$ within each corpus; downstream values are equal-weight means across six corpora. Transfer columns report $\kappa$. Standard-budget entries report mean $\pm$ sample SD across three REVE-small pretraining seeds, after averaging classifier seeds where applicable. Each $5\times$ condition uses one pretraining run and is reported as a point estimate. The unequal numbers of pretraining runs make this a descriptive budget comparison.}
\label{tab:s4_budget}
\begin{tabular}{lrrrrr}
\toprule
Condition & \shortstack{$R_0^2$\\TUEG} & \shortstack{$R_0^2$\\Six corpora} & Mean pool & Electrodes kept & Attention \\
\midrule
Real EEG & \shortstack{$0.683$\\$\pm 0.001$} & \shortstack{$0.657$\\$\pm 0.003$} & \shortstack{$0.312$\\$\pm 0.005$} & \shortstack{$0.415$\\$\pm 0.002$} & \shortstack{$0.347$\\$\pm 0.005$} \\
Real EEG, $5\times$ & 0.695 & 0.657 & 0.335 & 0.443 & 0.371 \\
PSD + cross-spectra & \shortstack{$0.671$\\$\pm 0.000$} & \shortstack{$0.648$\\$\pm 0.003$} & \shortstack{$0.311$\\$\pm 0.007$} & \shortstack{$0.406$\\$\pm 0.003$} & \shortstack{$0.349$\\$\pm 0.005$} \\
PSD + cross-spectra, $5\times$ & 0.681 & 0.647 & 0.340 & 0.425 & 0.368 \\
\bottomrule
\end{tabular}
\end{table}

The extended-budget models achieve modestly higher mean transfer (Table~\ref{tab:s4_budget}). With attention pooling, real EEG reaches 0.371 at the larger budget, compared with the standard-budget three-seed mean of 0.347; PSD + cross-spectra reaches 0.368, compared with 0.349. Their scores remain close. These observations support persistence of the real-versus-synthetic comparison at the larger budget; they do not support the claim that further pretraining never helps. Each extended-budget condition has one pretraining run, so its pretraining-seed variability is not estimated.

\subsection{Paired differences and signal properties by task}
\label{app:paired_signal}
Table~\ref{tab:paired_configs} gives the paired difference between PSD + cross-spectra and original-EEG pretraining for every configuration, with equivalence tests. Table~\ref{tab:signal_summary} gives, for each signal version and task family, the share of the pretraining gain and of EEGNetv4's score that it keeps.

\begin{table}[!ht]
\centering
\footnotesize
\setlength{\tabcolsep}{3pt}
\caption{\textbf{Paired difference, PSD + cross-spectra minus original EEG.} Frozen transfer on real EEG with attention pooling (first five rows), and supervised EEGNetv4 trained and tested on each signal (last row). 95\% intervals resample test subjects within each corpus and pretraining seeds, paired across arms. $p_{\mathrm{TOST}}$: equivalence to zero within $\pm 0.02~\kappa$, 10\% of the mean gain from pretraining. Family columns give the point difference; $^{*}$ marks a 95\% interval excluding zero. REVE-base and the $5\times$ budget are single runs.}
\label{tab:paired_configs}
\resizebox{\linewidth}{!}{\begin{tabular}{lrrrrrrrr}
\toprule
Configuration & Six corpora & 95\% interval & $p_{\mathrm{TOST}}$ & MI & ERP & SSVEP & Sleep & Clinical \\
\midrule
REVE-small, MAE & $+0.002$ & $[-0.008, +0.012]$ & $<0.001$ & $+0.007$ & $+0.007$ & $-0.003$ & $+0.000$ & $-0.006$ \\
REVE-small, JEPA & $-0.005$ & $[-0.013, +0.004]$ & $<0.001$ & $+0.005$ & $-0.004$ & $-0.017$ & $-0.005$ & $-0.013$ \\
REVE-small, JEPA + recon. & $-0.003$ & $[-0.010, +0.005]$ & $<0.001$ & $+0.000$ & $+0.003$ & $+0.012$ & $-0.008$ & $-0.022^{*}$ \\
REVE-base, MAE & $-0.003$ & $[-0.012, +0.006]$ & $<0.001$ & $+0.003$ & $-0.009$ & $-0.009$ & $+0.018$ & $-0.024^{*}$ \\
REVE-small, MAE, $5\times$ budget & $-0.003$ & $[-0.013, +0.007]$ & $<0.001$ & $+0.012$ & $-0.014$ & $-0.018$ & $+0.006$ & $-0.017$ \\
\addlinespace[2pt]
EEGNetv4 (supervised) & $-0.150$ & $[-0.170, -0.128]$ & --- & $-0.296^{*}$ & $-0.341^{*}$ & $-0.012$ & $+0.017$ & $+0.029^{*}$ \\
\bottomrule
\end{tabular}}
\end{table}

\begin{table}[!ht]
\centering
\small
\setlength{\tabcolsep}{4pt}
\caption{\textbf{What each preserved signal property keeps, by task.} Top: share of the pretraining gain over the random encoder kept when pretraining on the transformed signal, $(\kappa_v-\kappa_0)/(\kappa_{\mathrm{orig}}-\kappa_0)$, frozen transfer on real EEG with attention pooling. Bottom: share of the EEGNetv4 score kept when training and testing on the transformed signal. Means over three seeds, except Mean + variance (one pretraining run). ERP has no pretraining gain to keep. The motor-imagery gain is small ($0.05~\kappa$), so its shares are imprecise (PSD + cross-spectra: 95\% interval 88--176\%).}
\label{tab:signal_summary}
\begin{tabular}{lrrrrrr}
\toprule
Signal version & MI & ERP & SSVEP & Sleep & Clinical & Six corpora \\
\midrule
\multicolumn{7}{l}{\textit{Share of the pretraining gain kept}} \\
PSD + cross-spectra & $115\%$ & --- & $99\%$ & $100\%$ & $98\%$ & $101\%$ \\
PSD & $57\%$ & --- & $46\%$ & $83\%$ & $68\%$ & $69\%$ \\
Covariance & $104\%$ & --- & $73\%$ & $44\%$ & $49\%$ & $58\%$ \\
Mean + variance & $-39\%$ & --- & $-28\%$ & $21\%$ & $10\%$ & $4\%$ \\
\addlinespace[2pt]
\multicolumn{7}{l}{\textit{Share of the EEGNetv4 score kept}} \\
PSD + cross-spectra & $29\%$ & $0\%$ & $99\%$ & $103\%$ & $106\%$ & $71\%$ \\
PSD & $9\%$ & $0\%$ & $74\%$ & $94\%$ & $86\%$ & $54\%$ \\
Covariance & $20\%$ & $1\%$ & $10\%$ & $30\%$ & $56\%$ & $23\%$ \\
Mean + variance & $1\%$ & $0\%$ & $0\%$ & $1\%$ & $-3\%$ & $0\%$ \\
\bottomrule
\end{tabular}
\end{table}

\subsection{Does scale change the task dependence?}
\label{app:scaling}

\begin{figure}[!ht]
\centering
\includegraphics[width=\linewidth]{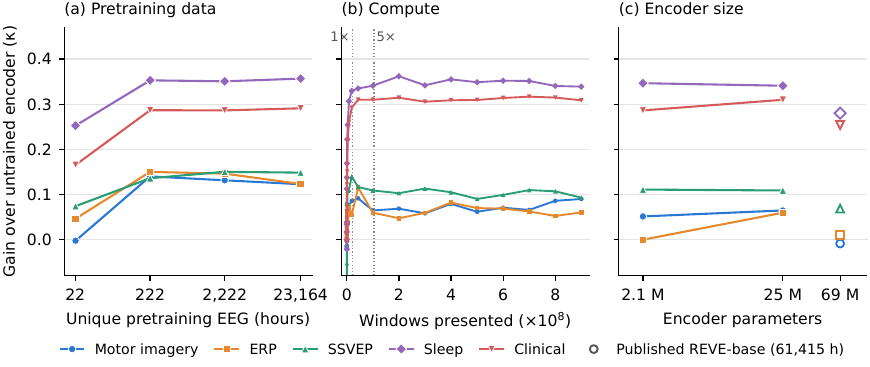}
\caption{\textbf{More data, compute or parameters does not close the gap between task families.} Gain in Cohen's $\kappa$ of the frozen encoder over the same architecture before pretraining, attention pooling, averaged over the corpora of each family (motor imagery 4, ERP 3, SSVEP 3, sleep 2, clinical 4; 21 corpora, subjects never seen). (a) REVE-small (12.7\,M; this ladder adds an auxiliary token-average loss to the recipe of Section~\ref{sec:surrogates}) on uniform subsets of TUEG from 22 to 23{,}164 hours, 42\,M window presentations each. (b) One REVE run (25.2\,M, 8 layers, 55\% masking) over 43 passes through TUEG; dotted lines mark the standard budget of Section~\ref{sec:surrogates} ($1\times$) and the extended one ($5\times$). (c) REVE with 2.1\,M and 25.2\,M encoder parameters, same recipe, 100\,M presentations; open markers: published REVE-base (69\,M, 61{,}415 hours, its own recipe), against its random copy. One pretraining run and one readout fit per point; neighbouring checkpoints in (b) differ by up to 0.06~$\kappa$ on ERPs.}
\label{fig:scaling_axes}
\end{figure}

\paragraph{Controlled scaling.} We vary pretraining data, compute and encoder size one at a time, using stored pretraining runs, and read every run on 21 corpora (subjects never seen) with the frozen attention readout of Section~\ref{sec:results}, against the same architecture before pretraining (Figure~\ref{fig:scaling_axes}). The gains on sleep and clinical EEG exceed those on motor imagery and ERPs by at least 0.12~$\kappa$ at every data and size point and at every checkpoint after the first pass. Beyond 222 hours of data, no family gains more than 0.012~$\kappa$; beyond the first pass, no family trends upwards (passes 5--14 and 34--43 differ by at most 0.017~$\kappa$). On the six corpora of Section~\ref{sec:surrogates}, where EEGNet gives an absolute reference (Table~\ref{tab:scaling_eegnet}), five times the budget or REVE-base adds about 0.02~$\kappa$ on motor imagery and ERPs: pretrained encoders then cover at most a fifth of the distance to EEGNet on motor imagery and almost none on ERPs, while they match or exceed EEGNet on sleep and clinical EEG. PSD + cross-spectra pretraining remains as good as original EEG at both larger scales (Table~\ref{tab:paired_configs}). Scales beyond 69 million parameters, 23{,}164 hours of distinct data and 43 passes remain untested.

\begin{table}[!ht]
\centering
\footnotesize
\setlength{\tabcolsep}{4pt}
\caption{\textbf{Share of the distance from a random encoder to EEGNet covered by pretraining, at three scales.} Six corpora and attention pooling of Section~\ref{sec:surrogates}, original-EEG MAE pretraining: $(\kappa_{\mathrm{pre}}-\kappa_{\mathrm{rand}})/(\kappa_{\mathrm{EEGNet}}-\kappa_{\mathrm{rand}})$ by family, each architecture against its own random encoder. REVE-small averages three pretraining seeds; the other columns are single runs. Values above 100\% mean that the frozen encoder beats EEGNet.}
\label{tab:scaling_eegnet}
\begin{tabular}{lrrrrr}
\toprule
 & EEGNet & REVE-small & REVE-small & REVE-base & REVE-base \\
 & $\kappa$ & $1\times$ budget & $5\times$ budget & $1\times$ budget & published \\
Family & & 12.7\,M & 12.7\,M & 69.2\,M & 61{,}415\,h \\
\midrule
Motor imagery & 0.418 & 14\% & 18\% & 19\% & 11\% \\
ERP & 0.341 & $-3$\% & 2\% & 2\% & $-5$\% \\
SSVEP & 0.894 & 61\% & 70\% & 67\% & 80\% \\
Sleep & 0.537 & 101\% & 105\% & 104\% & 89\% \\
Clinical & 0.532 & 119\% & 131\% & 140\% & 122\% \\
\bottomrule
\end{tabular}
\end{table}

\paragraph{Choice of pretraining data.} Table~\ref{tab:corpus_choice} keeps the 25.2\,M REVE encoder, recipe, seed and budget, and changes only the corpus: TUEG; Stieger2021, a single motor-imagery corpus (615 hours); or a mixture of 23 public corpora (769 hours) that contains the unlabeled recordings of six read-out corpora, test subjects included. The pattern across families is the same for the three. Motor-imagery data alone gives the lowest motor-imagery gain, and on the six corpora whose recordings the mixture contains, it changes the gain relative to TUEG by $+0.010$~$\kappa$ on motor imagery and $-0.003$ on ERPs. Similar data helps where the objective already helps: on TUAB and TUEV, from the same hospital archive as TUEG (TUAB recordings excluded), TUEG gives 0.09~$\kappa$ more than the mixture.

\begin{table}[!ht]
\centering
\footnotesize
\setlength{\tabcolsep}{4pt}
\caption{\textbf{The pretraining corpus barely changes the gain by family.} Gain in $\kappa$ over the same untrained encoder, 21-corpus read-out of Figure~\ref{fig:scaling_axes}, REVE (25.2\,M) pretrained for $10^8$ and $2\times10^8$ window presentations on each corpus. One pretraining run and one readout fit per cell.}
\label{tab:corpus_choice}
\begin{tabular}{llrrrrr}
\toprule
Presentations & Pretraining corpus & MI & ERP & SSVEP & Sleep & Clinical \\
\midrule
$10^8$ & TUEG (23{,}164\,h) & 0.065 & 0.060 & 0.109 & 0.341 & 0.310 \\
 & Stieger2021, MI only (615\,h) & 0.028 & 0.007 & 0.116 & 0.330 & 0.237 \\
 & 23 public corpora (769\,h) & 0.058 & 0.027 & 0.125 & 0.359 & 0.265 \\
\midrule
$2\times10^8$ & TUEG & 0.068 & 0.047 & 0.103 & 0.362 & 0.314 \\
 & Stieger2021, MI only & $-0.016$ & 0.052 & 0.135 & 0.317 & 0.258 \\
 & 23 public corpora & 0.080 & 0.069 & 0.132 & 0.365 & 0.275 \\
\bottomrule
\end{tabular}
\end{table}

\begin{figure}[!ht]
\centering
\includegraphics[width=0.8\linewidth]{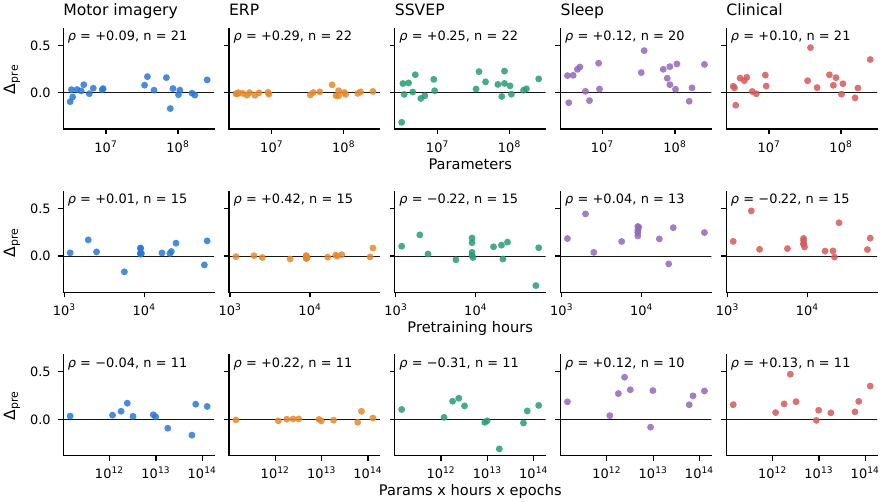}
\caption{\textbf{Across published encoders, the pretraining gain does not follow size, data or compute.} Each point is one encoder, averaged over the datasets of a family (attention pooling). Spearman $\rho$ with $n$ encoders; none of the 15 associations survives Holm correction. Pretraining hours are those stated in each model's paper (15 encoders state them, from 1{,}186 hours for LEAD to 61{,}415 for REVE), and compute is the proxy parameters $\times$ hours $\times$ epochs, available for fewer encoders. These encoders differ in architecture, objective and data at once, and with $n=10$--$22$ only a strong monotonic trend could be detected. Controlled comparisons, where only one factor changes, are in Figure~\ref{fig:scaling_axes}.}
\label{fig:gain_vs_scale}
\end{figure}

\paragraph{Published encoders.} Across the published encoders of Section~\ref{sec:results}, the pretraining gain shows no association with parameters, pretraining hours or compute (Figure~\ref{fig:gain_vs_scale}).

\section{Experimental setup for Section~\ref{sec:retention}}
\label{app:section5}
\label{app:retention}
\suppressfloats[t]

\paragraph{The question: once a useful representation has been learned, what keeps it useful?}
A model can reconstruct an EEG signal accurately without making the target task easy to predict from its features. The experiments in Section~\ref{sec:retention} therefore ask a practical question: \emph{if supervision first teaches an encoder a useful representation, can reconstruction alone preserve that benefit?} This matters because acquisition and retention are different requirements. A representation must first become useful, and it must remain useful as training continues.

The experiment follows this question directly. We first train with reconstruction and task labels together. We then make two identical copies of the resulting training state: one \textbf{keeps supervision}, while the other \textbf{removes supervision} and continues with reconstruction alone. A third model, trained with \textbf{reconstruction only from the beginning}, provides a reference at the same training budget. The two continuation branches see the same subjects and windows, in the same order, with the same reconstruction masks. New classifiers fitted to each frozen encoder on separate subjects measure what its representation can support at each checkpoint.

This design makes the comparison informative: it tests the retention of an already acquired predictive advantage, checks whether continued supervision helps preserve it, and measures whether either outcome comes with a substantial change in reconstruction loss. Each corpus is a separate experiment with a randomly initialized encoder.

\subsection{Data and subject separation}
\label{app:s5_data}

We use Stieger 2021 for four-class motor imagery, Lee2019 ERP for binary event-related potential decoding, ISRUC for five-class sleep staging, and TUAB for normal versus abnormal EEG classification. These provide two event-related tasks and two state-related tasks. Signals and labels from different corpora are never mixed.

We use the benchmark preprocessing: 4-s windows at 200\,Hz, standardized within each window and channel. The montage contains 19 electrodes for Stieger, ERP, and TUAB, and six for ISRUC. The fixed evaluation panel contains at most 6,000 windows per corpus (5,994 for ERP). A separate encoder-training stream is drawn from the corpus's labeled pool, capped at 60,000 windows before restricting it to encoder-training subjects.

Subjects are partitioned once, with seed 20260908, into three disjoint groups:
\begin{itemize}
 \item \textbf{Encoder training (A, approximately 40\%):} all data used to update the encoder, whether or not labels are used.
 \item \textbf{Readout development (B, 30\%):} data used to fit and select the classifiers that evaluate frozen representations. These subjects are further divided into fitting and validation groups in an approximately 2:1 ratio.
 \item \textbf{Testing (C, 30\%):} subjects held out for evaluation, never used to fit or select a model.
\end{itemize}
All sessions of a subject remain in the same group. TUAB splits are stratified by each subject's majority label in the evaluation panel. Table~\ref{tab:s5_split} gives the realized counts. No window from B or C enters encoder training, even without its label. Consequently, the readouts test whether the learned representation is useful for new subjects. The task classes are shared across groups; transfer to new tasks or classes is not tested.

\begin{table}[htbp]
\centering
\footnotesize
\setlength{\tabcolsep}{3pt}
\renewcommand{\arraystretch}{1.15}
\caption{Subject groups and evaluation windows. Encoder-training subjects (A), readout-development subjects (B), and test subjects (C) are disjoint. B is divided into fitting and validation subjects. Window counts describe the fixed evaluation panel, not the encoder-training stream.}
\label{tab:s5_split}
\begin{tabular*}{\linewidth}{@{\extracolsep{\fill}}lrrrrrr@{}}
\toprule
Corpus & Classes & \shortstack{Training\\subjects (A)} & \shortstack{Readout subjects (B)\\fit / val.} & \shortstack{Test\\subjects (C)} & \shortstack{Readout windows\\fit / val.} & \shortstack{Test\\windows} \\
\midrule
Stieger 2021 & 4 & 25 & 13 / 6 & 18 & 1160 / 703 & 1801 \\
Lee2019 ERP & 2 & 22 & 11 / 5 & 16 & 1221 / 555 & 1776 \\
ISRUC & 5 & 39 & 19 / 10 & 30 & 1166 / 631 & 1818 \\
TUAB & 2 & 930 & 465 / 233 & 698 & 1200 / 571 & 1803 \\
\bottomrule
\end{tabular*}
\end{table}

The finite A stream contains roughly 24,000 windows per corpus, which are repeatedly shuffled and presented during training. An implementation ``epoch'' contains 4,096,000 presentations and therefore includes many passes through this finite set. Presentation counts measure exposure, not distinct examples. Incomplete batches at the end of each shuffle are dropped, and a new permutation is drawn.

\subsection{Training: learn with supervision, then keep or remove it}
\label{app:s5_training}

\paragraph{Architecture and initialization.}
We use the small REVE-style masked autoencoder: embedding width 512, four encoder transformer layers, eight attention heads, and a two-layer decoder. Patches contain 200 samples with 20-sample overlap, giving four temporal patches per electrode. Random masking removes 75\% of electrode--patch tokens. Each independent encoder training run begins from random weights. Within each corpus and encoder seed, the reconstruction-only model and the model trained with supervision share exactly the same initialization. Encoder seeds 0, 1, and 2 correspond to random seeds.

\paragraph{What the model is trained to do.}
The reconstruction objective asks the model to recover masked signal patches. The supervised objective asks its features to predict the task label. Let $f_\theta$ and $g_\phi$ denote the encoder and reconstruction decoder, and let $m$ denote the reconstruction mask. The reconstruction loss $L_{\mathrm{recon}}$ is the mean absolute error on masked patches, using the window-standardized signal as the target, without additional target standardization within each patch. A linear classifier $q_\psi$ predicts the label from the mean of the encoder tokens of the complete window:
\begin{align}
 L_{\mathrm{label}}(x,y)
 &= \operatorname{CE}\!\left(q_\psi\!\left(\frac{1}{L}\sum_{\ell=1}^{L}f_\theta(x)_\ell\right),y\right), \\
 L_{\mathrm{joint}}(x,y,m)
 &= L_{\mathrm{recon}}(x,m)+\lambda L_{\mathrm{label}}(x,y).
\end{align}
Both terms use the same batch of A windows: a masked view for reconstruction and a complete view for classification. The main setting is $\lambda=0.02$; a control uses stronger supervision, $\lambda=0.1$, on encoder seed 0. The encoder and decoder are optimized together. The training-time classifier is never used to score frozen representations.

\paragraph{Step 1: establish what supervision adds.}
We train two models from the shared random initialization: one with reconstruction only, and one with reconstruction plus task supervision. At the end of this first phase, denoted $t_{\mathrm{split}}$, their frozen representations are compared on held-out subjects. This establishes whether supervision has produced a predictive advantage worth tracking. We call the supervised model at this point the \textbf{supervised start}: it is the shared starting checkpoint for the next step.

\paragraph{Step 2: test whether reconstruction preserves that advantage.}
We copy the complete supervised start into two continuation branches:
\begin{itemize}
 \item \textbf{Supervision kept:} continue training with reconstruction and task supervision together.
 \item \textbf{Supervision removed:} set the supervision weight to zero and continue with reconstruction alone.
\end{itemize}
Both branches inherit the same encoder, decoder, classifier, optimizer moments and step counters, training-loop state, and random-number state. The unused classifier remains present when supervision is removed, but contributes no supervised gradient. The \textbf{reconstruction-only reference} continues uninterrupted to the same final exposure, denoted $t_{\mathrm{end}}$.

The two reconstruction-only continuations have different histories: the reference has never used labels, whereas the supervision-removed branch starts from a representation that has already learned from labels. This distinction is central to the retention question.

\paragraph{Optimizer controls: repeat the comparison with a fresh optimizer.}
To assess the role of optimizer history, we repeat both continuations from the same supervised start, clearing optimizer moments and step counters in both. We also continue the reconstruction-only reference from its first-phase checkpoint with a fresh optimizer. These conditions keep the same descriptive names; the optimizer setting is listed separately in Table~\ref{tab:s5_branches}.

The main reconstruction-only, supervised-learning, supervision-kept, and supervision-removed conditions are run on all three encoder seeds for every corpus. The three optimizer-reset controls use encoder seed 0 and $\lambda=0.02$. The stronger-supervision control repeats supervised learning and both continuations on seed 0, using the same reconstruction-only reference.

\begin{table}[htbp]
\centering
\footnotesize
\setlength{\tabcolsep}{3pt}
\renewcommand{\arraystretch}{1.15}
\caption{Training conditions in plain language. ``Supervised start'' is the shared checkpoint at the end of the first phase. Reconstruction continues in every condition. The main retention comparison is \emph{supervision kept} versus \emph{supervision removed}, with the complete optimizer state copied into both branches. Reset controls change the optimizer setting in both branches.}
\label{tab:s5_branches}
\begin{tabular*}{\linewidth}{@{\extracolsep{\fill}}llll@{}}
\toprule
Condition & Starting point & Training objective & Optimizer \\
\midrule
\shortstack[l]{Reconstruction only\\(reference)} & Random initialization & Reconstruction & Continuous \\
\shortstack[l]{Supervised learning\\(first phase)} & Random initialization & \shortstack[l]{Reconstruction\\and labels} & Continuous \\
Supervision kept & Supervised start & \shortstack[l]{Reconstruction\\and labels} & Copied \\
Supervision removed & Supervised start & Reconstruction & Copied \\
\midrule
\multicolumn{4}{l}{\textit{Controls with a fresh optimizer}} \\
Supervision kept & Supervised start & \shortstack[l]{Reconstruction\\and labels} & Reset \\
Supervision removed & Supervised start & Reconstruction & Reset \\
Reconstruction only & \shortstack[l]{Reference at end\\of first phase} & Reconstruction & Reset \\
\bottomrule
\end{tabular*}
\end{table}

\paragraph{Optimization and training budget.}
Training uses AdamW with constant learning rate $10^{-3}$, weight decay 0.05, $(\beta_1,\beta_2)=(0.9,0.95)$, $\epsilon=10^{-8}$, batch size 2,048, and bfloat16 mixed precision. The first phase lasts 2,000 updates, or $t_{\mathrm{split}}=4,096,000$ window presentations. Continuation lasts 4,000 additional updates, or 8,192,000 presentations, giving a final cumulative budget of $t_{\mathrm{end}}=12,288,000$. These budgets are fixed independently of test performance.

Checkpoints are saved at first-phase steps 16, 64, 250, 500, 1,000, and 2,000, and at 16, 64, 250, 500, 1,000, 2,000, 3,000, and 4,000 updates after the split. The initial state and continuation starting states are also saved. In the tables, \emph{end of first phase} and \emph{end of training} refer to $t_{\mathrm{split}}$ and $t_{\mathrm{end}}$, respectively.

\paragraph{What is matched between the continuations.}
A private generator determines data order from encoder seed and stream epoch. Mask and position-noise generators are re-seeded before each update using encoder seed and cumulative update number. The split occurs at a stream-epoch boundary, and resumed branches start at the corresponding next epoch. Thus, the additional supervised forward pass cannot alter later reconstruction masks or data order. The implementation checks mask identity and batch fingerprints. Saved readout scores and reconstruction losses at the start of each continuation equal those of the parent checkpoint.

The branches are matched in data exposure and update count. Their computation differs: training with labels adds a forward and backward pass through the encoder on complete windows. A token-layer count estimates approximately 2.3 times the work per update compared with reconstruction alone; this is not a measured FLOP count. No control matching that additional computation through extra reconstruction views was run.

\subsection{Evaluation: can a new classifier use the frozen representation?}
\label{app:s5_readouts}

At every checkpoint, we freeze the encoder and fit a new classifier, or \emph{readout}, on group B. This classifier is independent of the one used during encoder training. It tests whether the current representation remains useful even when a new classifier is allowed to adapt to it. Consequently, a decline cannot be explained solely by the original classification head becoming outdated.

Encoder features, denoted $Z$, are extracted from complete, unmasked windows. All feature standardization, dimensionality reduction, and classifier selection use only the designated B fitting and validation data; C is used only for evaluation. We evaluate both the encoder features alone and their combination with classical EEG descriptors, denoted $H$. Giving the same descriptors to both competing readouts tests whether the encoder comparison still matters when these established features are available.

\paragraph{Primary attention readout.}
Token features are standardized per embedding dimension over fitting windows and token positions. One learned query $q$ pools the standardized tokens $z_\ell$:
\begin{equation}
 \alpha_\ell = \frac{\exp(q^\top z_\ell/\sqrt{512})}
 {\sum_j\exp(q^\top z_j/\sqrt{512})},\qquad
 z_{\mathrm{pool}}=\sum_\ell\alpha_\ell z_\ell.
\end{equation}
The query is initialized with independent $\mathcal{N}(0,1/512)$ components. A linear classifier receives either $z_{\mathrm{pool}}$ alone or its concatenation with the classical descriptor block. Query and classifier are trained by cross-entropy using Adam, learning rate $5\times10^{-4}$, weight decay $10^{-3}$, batches of 128, and at most 80 epochs. Validation pooled $\kappa$ selects the checkpoint, with patience 10 and an improvement threshold of $10^{-4}$. The best fitting-subject model is retained without refitting on validation subjects. Three readout initializations use seeds 20260908, 20260909, and 20260910.

\paragraph{Classical descriptors: a compact block and the full block.}
$H$ concatenates multi-resolution log-power spectra, band-specific log-covariance descriptors, and coherency descriptors from the benchmark. Their respective widths are 5,966, 7,980, and 2,052 on the 19-electrode corpora (15,998 total), and 1,884, 882, and 180 on ISRUC (2,946 total). Each group is standardized on fitting windows; near-constant columns are set to zero. Dividing each group by the square root of three times its number of active columns gives equal total variance to the three groups. The concatenation is scaled by $\sqrt{d_Z}$, where $d_Z$ is the readout's encoder-feature width.

The primary analysis uses $H_{512}$, the leading 512 principal directions of this group-balanced block, fitted on the readout-fitting windows. Principal-component scores retain their relative variance and are not whitened or independently re-standardized. We also evaluate full-width $H$: an advantage measured with a compact descriptor block need not persist when all descriptors are supplied. The descriptors alone are evaluated under the same classifier recipe, using an all-zero placeholder token in place of encoder features.

\paragraph{Alternative ways to pool encoder features.}
We evaluate two fixed alternatives to attention: global mean pooling ($d_Z=512$), and temporal mean pooling followed by concatenation of electrodes ($d_Z=512C_{\mathrm{electrodes}}$). The latter retains a separate feature block for each electrode.

Their linear classifiers minimize cross-entropy plus an explicit squared-weight penalty, using full-batch Adam with learning rate $10^{-2}$ for at most 300 steps. Penalties in $\{10^{-3},10^{-2},10^{-1}\}$ and stopping steps are selected by validation $\kappa$, evaluated every five updates. Ties between penalty values favor the larger penalty. Each selected classifier is refitted on all of B for the selected number of steps, including refitting feature transformations on B. Three classifier initializations use seeds 0, 1, and 2. Global mean pooling is evaluated with both full $H$ and $H_{512}$; electrode-preserving pooling uses full $H$.

\subsection{What each comparison tells us}
\label{app:s5_statistics}

\paragraph{Overall performance and paired subject comparisons.}
We report two complementary summaries. The \textbf{pooled score}, $\kappa_{\mathrm{pool}}$, is Cohen's $\kappa$ over all test windows in C. It is computed separately for each of the three fitted readout heads and then averaged; predictions are not ensembled. A \textbf{paired subject comparison} first computes the score difference within each eligible subject and then averages these differences, giving each eligible subject equal weight.

For the latter, let $s_c^b(t;V)$ be the mean readout-head $\kappa$ within test subject $c$, for condition $b$, checkpoint $t$, and readout input $V$. The eligible set $\mathcal{C}_*$ contains subjects with at least two test windows and at least two observed target classes. It includes all 18 Stieger, 16 ERP, and 30 ISRUC test subjects, but only 18 of the 698 TUAB test subjects. A paired subject comparison therefore need not equal the difference of two pooled scores. TUAB's paired estimates concern only this small subset with both labels and cannot provide uncertainty intervals for its full test population.

In the equations below, the superscripts $\mathrm{recon}$, $\mathrm{start}$, $\mathrm{kept}$, and $\mathrm{removed}$ refer to the reconstruction-only reference, the supervised start, and the two continuation conditions. The supervised start is always evaluated at $t_{\mathrm{split}}$; its time argument is omitted.

\paragraph{1. Initial gain: did supervision make the representation more useful?}
At the end of the first phase, we compare the supervised start with the reconstruction-only reference using encoder features alone:
\begin{equation}
 \Delta_Z = \frac{1}{|\mathcal{C}_*|}\sum_{c\in\mathcal{C}_*}
 \left[s_c^{\mathrm{start}}(Z)-s_c^{\mathrm{recon}}(t_{\mathrm{split}};Z)\right].
\end{equation}
A positive value establishes an acquired predictive advantage under the tested readout. This is the benefit whose subsequent retention we examine.

\paragraph{2. Gain with descriptors: does the encoder advantage remain when classical features are available?}
At any checkpoint, we give both readouts the same descriptor block and compare condition $b$ with the reconstruction-only reference at the same training budget:
\begin{equation}
 G_H^b(t)=\frac{1}{|\mathcal{C}_*|}\sum_{c\in\mathcal{C}_*}
 \left[s_c^b(t;[H_{512},Z])-s_c^{\mathrm{recon}}(t;[H_{512},Z])\right].
\end{equation}
Full-descriptor controls replace $H_{512}$ by $H$. This \emph{gain with descriptors} compares two encoders while holding the supplied descriptors fixed. Improvement over the descriptors alone is a separate quantity, defined using pooled scores:
\begin{equation}
 A_{H,\mathrm{pool}}^b(t)
 =\kappa_{\mathrm{pool}}([H,Z_b(t)])-\kappa_{\mathrm{pool}}(H).
\end{equation}
The same definition applies to the compact block by replacing $H$ with $H_{512}$.

\paragraph{3. Change after removal: does performance actually fall?}
We compare the supervision-removed branch with its own starting representation:
\begin{equation}
 \mathrm{Change}_Z(t)=\frac{1}{|\mathcal{C}_*|}\sum_{c\in\mathcal{C}_*}
 \left[s_c^{\mathrm{removed}}(t;Z)-s_c^{\mathrm{start}}(Z)\right].
\end{equation}
A negative value means that predictive performance has declined since the supervised start.

\paragraph{4. Benefit of keeping supervision: how far apart do the continuations become?}
We compare the two branches at the same checkpoint:
\begin{equation}
 \mathrm{Gap}_Z(t)=\frac{1}{|\mathcal{C}_*|}\sum_{c\in\mathcal{C}_*}
 \left[s_c^{\mathrm{kept}}(t;Z)-s_c^{\mathrm{removed}}(t;Z)\right].
\end{equation}
A positive gap favors keeping supervision. The change and the gap answer different questions: a gap could arise because one branch improves while the other stays stable. Reporting the change from the shared starting point is therefore necessary to establish an actual decline. The optimizer-reset comparison uses the same gap definition, with both branches reset.

\paragraph{How much of the initial advantage remains?}
When the initial gain is substantial, we also report the fraction of the initial pooled advantage that remains above the reconstruction-only reference:
\begin{equation}
 F_b(t)=\frac{\kappa_{\mathrm{pool}}(Z_b(t))-\kappa_{\mathrm{pool}}(Z_{\mathrm{recon}}(t))}
 {\kappa_{\mathrm{pool}}(Z_{\mathrm{start}})-\kappa_{\mathrm{pool}}(Z_{\mathrm{recon}}(t_{\mathrm{split}}))}.
\end{equation}
Both numerator and denominator use pooled scores. This is a fraction of the measured predictive advantage, not a fraction of all task information in the encoder. It can exceed 100\% if the advantage over the reference grows. We interpret it only for Stieger and ERP, where the denominator is clearly positive.

\paragraph{Uncertainty.}
Within an encoder seed, 95\% intervals use 2,000 paired bootstrap resamples of the eligible test subjects, averaging the per-subject differences. These intervals condition on the trained encoder and fitted readouts. Encoder-training variability is reported separately as individual seed results or mean $\pm$ sample standard deviation across the three encoder seeds (divisor $3-1$). The three classifier initializations are not treated as additional encoder-training replications. Intervals are pointwise and do not include variability from alternative subject partitions.

\paragraph{Reconstruction quality: is the predictive benefit compatible with the reconstruction objective?}
We also evaluate reconstruction loss. This tests whether representations with different predictive performance can nevertheless reconstruct comparably well. The model's own masked-patch L1 loss is measured in evaluation mode on fixed banks drawn from A, B, and C. A shares subject identities with encoder training; B and C contain subjects unseen during encoder training. Bank sizes are given in Table~\ref{tab:s5_banks}.

Each bank is split into batches of at most 32 windows. Masks are fixed by batch index and seed across all conditions and checkpoints. The stored bank loss is the unweighted mean of these batch losses, including a final smaller batch when present. Each batch contributes one loss value, called a \emph{loss cell} below.

\begin{table}[htbp]
\centering
\small
\setlength{\tabcolsep}{3pt}
\renewcommand{\arraystretch}{1.15}
\caption{Reconstruction-bank sizes and numbers of evaluation batches (loss cells), in encoder-training / readout-development / test order (A / B / C). Each batch contains 32 windows except a possible final smaller batch.}
\label{tab:s5_banks}
\begin{tabular*}{\linewidth}{@{\extracolsep{\fill}}lrr@{}}
\toprule
Corpus & Windows (A / B / C) & Batches (A / B / C) \\
\midrule
Stieger 2021 & 2048 / 1863 / 1801 & 64 / 59 / 57 \\
Lee2019 ERP & 2048 / 1776 / 1776 & 64 / 56 / 56 \\
ISRUC & 2048 / 1797 / 1818 & 64 / 57 / 57 \\
TUAB & 2048 / 1771 / 1803 & 64 / 56 / 57 \\
\bottomrule
\end{tabular*}
\end{table}

Let $\overline{L}_B^{\,b}(t)$ be the stored reconstruction loss on bank B for condition $b$. Its relative excess over the reconstruction-only reference at the same budget is
\begin{equation}
 E_B^b(t)=\frac{\overline{L}_B^{\,b}(t)}{\overline{L}_B^{\,\mathrm{recon}}(t)}-1.
\end{equation}
Negative values mean better reconstruction than the reference. The 95\% interval uses 2,000 paired bootstrap resamples of loss cells, recomputing the ratio in each resample. The pre-specified compatibility criterion requires the upper interval bound to be at most $+1\%$ on bank B. These intervals describe variability over the fixed evaluation batches; they are not subject-bootstrap intervals and do not establish population-wide equivalence. We apply the criterion to the supervised start and to both continuation branches at the end of training.

\subsection{Main results: what is learned, and what remains?}
\label{app:s5_main_results}

\paragraph{Supervision produces a substantial advantage on the two event-related tasks.}
Tables~\ref{tab:s5_scores_z} and~\ref{tab:s5_scores_hz} report the absolute attention-readout scores for every encoder seed. At the end of the first phase, mean pooled $\kappa$ increases from 0.074 to 0.367 on Stieger and from 0.005 to 0.277 on ERP when supervision is added. The paired subject-level initial gains are $+0.267\pm0.008$ and $+0.259\pm0.015$. With $H_{512}$ supplied to both classifiers, the gains remain $+0.203\pm0.004$ and $+0.258\pm0.018$ (Table~\ref{tab:s5_conditional}). These gains show that the architecture can learn features that improve prediction on new subjects, including when the supplied descriptor block is already available. There is therefore a clear acquired benefit whose retention can be tested.

The advantage is smaller on ISRUC: pooled $\kappa$ increases from 0.402 to 0.461, and the subject-level gain with descriptors is $+0.021\pm0.015$. TUAB shows little pooled improvement, from 0.497 to 0.507; its pooled score with $H_{512}$ slightly decreases, from 0.533 to 0.526. Its paired estimates concern a small eligible subset and differ from these pooled comparisons (Table~\ref{tab:s5_contrasts}). The strongest evidence for a substantial acquired advantage is thus concentrated in Stieger and ERP.

\begin{table}[htbp]
\centering
\footnotesize
\setlength{\tabcolsep}{3pt}
\renewcommand{\arraystretch}{1.15}
\caption{Performance using encoder features alone ($Z$), with the attention readout and $\lambda=0.02$. Each entry is pooled $\kappa$ over all test windows, averaged over three independently fitted readout heads for one encoder seed. The first phase ends after 4.096 million presentations; training ends after 12.288 million. Both continuation branches begin from the supervised start. The reconstruction-only reference never uses labels.}
\label{tab:s5_scores_z}
\begin{tabular*}{\linewidth}{@{\extracolsep{\fill}}lrrrrrr@{}}
\toprule
Corpus & Seed & \multicolumn{2}{c}{End of first phase} & \multicolumn{3}{c}{End of training} \\
\cmidrule(lr){3-4}\cmidrule(lr){5-7}
 & & \shortstack{Reconstruction\\only} & \shortstack{Supervised\\start} & \shortstack{Reconstruction\\only} & \shortstack{Supervision\\kept} & \shortstack{Supervision\\removed} \\
\midrule
Stieger 2021 & 0 & $0.073$ & $0.367$ & $0.065$ & $0.273$ & $0.151$ \\
 & 1 & $0.073$ & $0.375$ & $0.065$ & $0.312$ & $0.161$ \\
 & 2 & $0.076$ & $0.360$ & $0.064$ & $0.290$ & $0.125$ \\
\addlinespace
Lee2019 ERP & 0 & $0.002$ & $0.274$ & $-0.006$ & $0.263$ & $0.003$ \\
 & 1 & $0.012$ & $0.303$ & $0.015$ & $0.245$ & $0.021$ \\
 & 2 & $0.001$ & $0.253$ & $-0.019$ & $0.271$ & $0.016$ \\
\addlinespace
ISRUC & 0 & $0.417$ & $0.456$ & $0.427$ & $0.463$ & $0.427$ \\
 & 1 & $0.393$ & $0.455$ & $0.413$ & $0.443$ & $0.432$ \\
 & 2 & $0.396$ & $0.472$ & $0.409$ & $0.437$ & $0.448$ \\
\addlinespace
TUAB & 0 & $0.492$ & $0.509$ & $0.501$ & $0.516$ & $0.483$ \\
 & 1 & $0.508$ & $0.511$ & $0.525$ & $0.524$ & $0.498$ \\
 & 2 & $0.491$ & $0.499$ & $0.520$ & $0.495$ & $0.504$ \\
\bottomrule
\end{tabular*}
\end{table}

\begin{table}[htbp]
\centering
\footnotesize
\setlength{\tabcolsep}{3pt}
\renewcommand{\arraystretch}{1.15}
\caption{Performance using classical descriptors and encoder features together ($[H_{512},Z]$), with the attention readout and $\lambda=0.02$. Each entry is pooled $\kappa$ over all test windows, averaged over three independently fitted readout heads for one encoder seed. The first phase ends after 4.096 million presentations; training ends after 12.288 million. The corresponding descriptor-only scores are 0.130 (Stieger), $-0.031$ (ERP), 0.544 (ISRUC), and 0.532 (TUAB).}
\label{tab:s5_scores_hz}
\begin{tabular*}{\linewidth}{@{\extracolsep{\fill}}lrrrrrr@{}}
\toprule
Corpus & Seed & \multicolumn{2}{c}{End of first phase} & \multicolumn{3}{c}{End of training} \\
\cmidrule(lr){3-4}\cmidrule(lr){5-7}
 & & \shortstack{Reconstruction\\only} & \shortstack{Supervised\\start} & \shortstack{Reconstruction\\only} & \shortstack{Supervision\\kept} & \shortstack{Supervision\\removed} \\
\midrule
Stieger 2021 & 0 & $0.140$ & $0.363$ & $0.133$ & $0.281$ & $0.173$ \\
 & 1 & $0.153$ & $0.370$ & $0.135$ & $0.315$ & $0.159$ \\
 & 2 & $0.142$ & $0.356$ & $0.134$ & $0.292$ & $0.171$ \\
\addlinespace
Lee2019 ERP & 0 & $0.012$ & $0.275$ & $0.013$ & $0.202$ & $0.015$ \\
 & 1 & $0.001$ & $0.295$ & $0.017$ & $0.211$ & $0.007$ \\
 & 2 & $-0.018$ & $0.239$ & $-0.007$ & $0.260$ & $0.005$ \\
\addlinespace
ISRUC & 0 & $0.545$ & $0.547$ & $0.549$ & $0.553$ & $0.546$ \\
 & 1 & $0.527$ & $0.552$ & $0.534$ & $0.552$ & $0.553$ \\
 & 2 & $0.532$ & $0.565$ & $0.536$ & $0.550$ & $0.562$ \\
\addlinespace
TUAB & 0 & $0.530$ & $0.524$ & $0.538$ & $0.523$ & $0.527$ \\
 & 1 & $0.536$ & $0.541$ & $0.548$ & $0.534$ & $0.529$ \\
 & 2 & $0.533$ & $0.512$ & $0.536$ & $0.543$ & $0.543$ \\
\bottomrule
\end{tabular*}
\end{table}

\paragraph{Removing supervision erodes much of that advantage.}
By the end of training, Stieger's pooled scores are $0.291\pm0.020$ with supervision kept, $0.146\pm0.019$ with supervision removed, and $0.065\pm0.001$ for the reconstruction-only reference. ERP reaches $0.260\pm0.013$, $0.013\pm0.009$, and $-0.003\pm0.017$, respectively.

Two comparisons establish what happened. First, after supervision is removed, the subject-level change from the supervised start is $-0.206\pm0.014$ on Stieger and $-0.256\pm0.015$ on ERP: performance actually falls. Second, keeping supervision produces a final advantage over removing it of $+0.138\pm0.022$ and $+0.240\pm0.016$. For every encoder seed on both tasks, the change interval lies below zero and the gap interval lies above zero (Table~\ref{tab:s5_contrasts}).

The same pattern appears when classical descriptors are available. The final gain over the reconstruction-only reference is $+0.028\pm0.010$ on Stieger and $+0.002\pm0.007$ on ERP after supervision is removed, compared with $+0.150\pm0.015$ and $+0.205\pm0.041$ when it is kept. Across seeds, removing supervision retains 22--32\% of the initial pooled advantage on Stieger and 2--14\% on ERP. Keeping it retains 70--82\% and 79--115\%, respectively.

Keeping supervision therefore helps preserve the learned advantage, but does not guarantee constant performance: on Stieger, this branch also declines from its starting score. Table~\ref{tab:s5_trajectory} reports every scheduled continuation checkpoint for both tasks.

ISRUC shows a smaller subject-level decline after removal ($-0.026\pm0.003$), while the gap between keeping and removing supervision varies across seeds ($+0.013\pm0.022$), with little difference remaining when descriptors are supplied. TUAB also lacks a consistent subject-level retention gap. Neither corpus provides an initial advantage with descriptors comparable to Stieger or ERP.

\begin{table}[htbp]
\centering
\footnotesize
\setlength{\tabcolsep}{3pt}
\renewcommand{\arraystretch}{1.15}
\caption{What supervision adds and what happens after its removal: paired subject-level comparisons with 95\% bootstrap intervals, attention readout, and $\lambda=0.02$. Initial gains compare the supervised start with the reconstruction-only reference. ``Change after removal'' compares the final supervision-removed score with its starting score; negative values indicate decline. ``Kept $-$ removed'' compares the final continuation scores. All columns use $Z$ alone except ``with descriptors'', which uses $[H_{512},Z]$. These are averages of within-subject differences, not differences of pooled scores. TUAB intervals concern only 18 eligible subjects.}
\label{tab:s5_contrasts}
\begin{tabular*}{\linewidth}{@{\extracolsep{\fill}}lrrrrr@{}}
\toprule
Corpus & Seed & \multicolumn{2}{c}{End of first phase} & \multicolumn{2}{c}{End of training} \\
\cmidrule(lr){3-4}\cmidrule(lr){5-6}
 & & \shortstack{Initial gain\\($Z$ alone)} & \shortstack{Initial gain\\with descriptors} & \shortstack{Change after\\removal} & \shortstack{Kept $-$\\removed} \\
\midrule
Stieger 2021 & 0 & \shortstack{$+0.264$\\$[+0.193,+0.335]$} & \shortstack{$+0.204$\\$[+0.150,+0.261]$} & \shortstack{$-0.194$\\$[-0.246,-0.139]$} & \shortstack{$+0.114$\\$[+0.086,+0.143]$} \\
 & 1 & \shortstack{$+0.276$\\$[+0.221,+0.333]$} & \shortstack{$+0.205$\\$[+0.156,+0.258]$} & \shortstack{$-0.203$\\$[-0.239,-0.167]$} & \shortstack{$+0.144$\\$[+0.107,+0.183]$} \\
 & 2 & \shortstack{$+0.260$\\$[+0.190,+0.335]$} & \shortstack{$+0.199$\\$[+0.141,+0.257]$} & \shortstack{$-0.221$\\$[-0.277,-0.169]$} & \shortstack{$+0.155$\\$[+0.109,+0.203]$} \\
\addlinespace
Lee2019 ERP & 0 & \shortstack{$+0.262$\\$[+0.196,+0.330]$} & \shortstack{$+0.249$\\$[+0.186,+0.311]$} & \shortstack{$-0.260$\\$[-0.328,-0.193]$} & \shortstack{$+0.246$\\$[+0.190,+0.300]$} \\
 & 1 & \shortstack{$+0.272$\\$[+0.195,+0.346]$} & \shortstack{$+0.279$\\$[+0.185,+0.371]$} & \shortstack{$-0.269$\\$[-0.337,-0.193]$} & \shortstack{$+0.222$\\$[+0.167,+0.278]$} \\
 & 2 & \shortstack{$+0.243$\\$[+0.191,+0.294]$} & \shortstack{$+0.246$\\$[+0.184,+0.309]$} & \shortstack{$-0.240$\\$[-0.284,-0.191]$} & \shortstack{$+0.253$\\$[+0.182,+0.323]$} \\
\addlinespace
ISRUC & 0 & \shortstack{$+0.041$\\$[+0.024,+0.057]$} & \shortstack{$+0.005$\\$[-0.015,+0.026]$} & \shortstack{$-0.029$\\$[-0.043,-0.015]$} & \shortstack{$+0.036$\\$[+0.019,+0.053]$} \\
 & 1 & \shortstack{$+0.067$\\$[+0.046,+0.088]$} & \shortstack{$+0.024$\\$[+0.005,+0.044]$} & \shortstack{$-0.025$\\$[-0.046,-0.003]$} & \shortstack{$+0.012$\\$[-0.006,+0.028]$} \\
 & 2 & \shortstack{$+0.076$\\$[+0.051,+0.102]$} & \shortstack{$+0.034$\\$[+0.016,+0.052]$} & \shortstack{$-0.024$\\$[-0.041,-0.008]$} & \shortstack{$-0.008$\\$[-0.025,+0.012]$} \\
\addlinespace
TUAB & 0 & \shortstack{$+0.020$\\$[-0.072,+0.114]$} & \shortstack{$+0.008$\\$[-0.084,+0.083]$} & \shortstack{$+0.044$\\$[-0.065,+0.176]$} & \shortstack{$-0.061$\\$[-0.191,+0.025]$} \\
 & 1 & \shortstack{$-0.041$\\$[-0.127,+0.036]$} & \shortstack{$-0.047$\\$[-0.138,+0.028]$} & \shortstack{$+0.019$\\$[-0.034,+0.071]$} & \shortstack{$+0.025$\\$[-0.029,+0.085]$} \\
 & 2 & \shortstack{$-0.078$\\$[-0.147,-0.006]$} & \shortstack{$-0.015$\\$[-0.075,+0.040]$} & \shortstack{$+0.097$\\$[+0.030,+0.170]$} & \shortstack{$-0.033$\\$[-0.103,+0.028]$} \\
\bottomrule
\end{tabular*}
\end{table}

\begin{table}[htbp]
\centering
\small
\setlength{\tabcolsep}{3pt}
\renewcommand{\arraystretch}{1.15}
\caption{Gain over the reconstruction-only reference when the same classical descriptors ($H_{512}$) are available to both readouts. The reference is evaluated at the same training budget as the condition shown. Values are paired subject-level gains, mean $\pm$ sample SD across three encoder seeds; $\lambda=0.02$. This SD measures variability across encoder seeds and is separate from subject-bootstrap uncertainty. TUAB estimates concern only its 18 eligible subjects.}
\label{tab:s5_conditional}
\begin{tabular*}{\linewidth}{@{\extracolsep{\fill}}lrrr@{}}
\toprule
Corpus & \shortstack{Supervised start\\(end of first phase)} & \shortstack{Supervision kept\\(end of training)} & \shortstack{Supervision removed\\(end of training)} \\
\midrule
Stieger 2021 & $+0.203\,\pm\,0.004$ & $+0.150\,\pm\,0.015$ & $+0.028\,\pm\,0.010$ \\
Lee2019 ERP & $+0.258\,\pm\,0.018$ & $+0.205\,\pm\,0.041$ & $+0.002\,\pm\,0.007$ \\
ISRUC & $+0.021\,\pm\,0.015$ & $+0.012\,\pm\,0.007$ & $+0.014\,\pm\,0.015$ \\
TUAB & $-0.018\,\pm\,0.028$ & $-0.005\,\pm\,0.018$ & $-0.004\,\pm\,0.006$ \\
\bottomrule
\end{tabular*}
\end{table}

\begin{table}[htbp]
\centering
\footnotesize
\setlength{\tabcolsep}{3pt}
\renewcommand{\arraystretch}{1.15}
\caption{How predictive performance evolves during continuation on the two event-related tasks. Values are pooled $\kappa$ on encoder features alone, using the attention readout and $\lambda=0.02$, reported as mean $\pm$ sample SD across three encoder seeds. At zero added updates, both continuation branches are exactly the supervised start. Added presentations count repeated window exposures after the first phase, not distinct windows or cumulative training exposure.}
\label{tab:s5_trajectory}
\begin{tabular*}{\linewidth}{@{\extracolsep{\fill}}lrrrrr@{}}
\toprule
Corpus & \shortstack{Added\\updates} & \shortstack{Added presentations\\(millions)} & \shortstack{Reconstruction\\only} & \shortstack{Supervision\\kept} & \shortstack{Supervision\\removed} \\
\midrule
Stieger 2021 & 0 & 0.000 & $0.074\,\pm\,0.002$ & $0.367\,\pm\,0.007$ & $0.367\,\pm\,0.007$ \\
 & 16 & 0.033 & $0.069\,\pm\,0.013$ & $0.370\,\pm\,0.007$ & $0.349\,\pm\,0.005$ \\
 & 64 & 0.131 & $0.081\,\pm\,0.008$ & $0.363\,\pm\,0.004$ & $0.330\,\pm\,0.004$ \\
 & 250 & 0.512 & $0.074\,\pm\,0.006$ & $0.364\,\pm\,0.006$ & $0.290\,\pm\,0.004$ \\
 & 500 & 1.024 & $0.069\,\pm\,0.007$ & $0.346\,\pm\,0.017$ & $0.253\,\pm\,0.023$ \\
 & 1000 & 2.048 & $0.076\,\pm\,0.002$ & $0.339\,\pm\,0.018$ & $0.214\,\pm\,0.014$ \\
 & 2000 & 4.096 & $0.074\,\pm\,0.009$ & $0.319\,\pm\,0.005$ & $0.166\,\pm\,0.016$ \\
 & 3000 & 6.144 & $0.069\,\pm\,0.002$ & $0.293\,\pm\,0.026$ & $0.151\,\pm\,0.022$ \\
 & 4000 & 8.192 & $0.065\,\pm\,0.001$ & $0.291\,\pm\,0.020$ & $0.146\,\pm\,0.019$ \\
\addlinespace
Lee2019 ERP & 0 & 0.000 & $0.005\,\pm\,0.006$ & $0.277\,\pm\,0.025$ & $0.277\,\pm\,0.025$ \\
 & 16 & 0.033 & $0.000\,\pm\,0.007$ & $0.290\,\pm\,0.001$ & $0.267\,\pm\,0.029$ \\
 & 64 & 0.131 & $0.004\,\pm\,0.003$ & $0.274\,\pm\,0.032$ & $0.240\,\pm\,0.030$ \\
 & 250 & 0.512 & $0.003\,\pm\,0.008$ & $0.301\,\pm\,0.021$ & $0.094\,\pm\,0.030$ \\
 & 500 & 1.024 & $-0.003\,\pm\,0.006$ & $0.294\,\pm\,0.027$ & $0.054\,\pm\,0.017$ \\
 & 1000 & 2.048 & $-0.002\,\pm\,0.002$ & $0.278\,\pm\,0.036$ & $0.019\,\pm\,0.010$ \\
 & 2000 & 4.096 & $0.009\,\pm\,0.006$ & $0.265\,\pm\,0.019$ & $0.004\,\pm\,0.004$ \\
 & 3000 & 6.144 & $-0.004\,\pm\,0.011$ & $0.240\,\pm\,0.031$ & $0.009\,\pm\,0.006$ \\
 & 4000 & 8.192 & $-0.003\,\pm\,0.017$ & $0.260\,\pm\,0.013$ & $0.013\,\pm\,0.009$ \\
\bottomrule
\end{tabular*}
\end{table}

\paragraph{The useful representation can coexist with comparable reconstruction quality.}
All 12 comparisons of the supervised start with the reconstruction-only reference satisfy the bank-B compatibility criterion at $\lambda=0.02$ (Table~\ref{tab:s5_recon}). At the end of training, 23 of the 24 continuation comparisons satisfy it. The exception is TUAB with supervision removed, seed 1: the excess is 0.84\%, with an upper interval bound of 1.06\%, narrowly above the threshold.

Every Stieger and ERP comparison satisfies the criterion. On these two tasks, both acquiring a large predictive advantage and retaining much of it are compatible with reconstruction loss within the chosen tolerance. This gives the retention result its practical relevance: losing the acquired advantage is not necessary to meet that reconstruction criterion. Comparable reconstruction loss can accompany markedly different predictive performance. Table~\ref{tab:s5_other_banks} gives the corresponding excesses on A and C.

\begin{table}[htbp]
\centering
\footnotesize
\setlength{\tabcolsep}{3pt}
\renewcommand{\arraystretch}{1.15}
\caption{Relative reconstruction excess on held-out bank B, in percent, with paired 95\% batch-bootstrap intervals. Each condition is compared with the reconstruction-only reference at the same training budget. Negative values mean lower L1 loss. At $\lambda=0.02$, the upper bound is below the pre-specified $+1\%$ margin in all 12 first-phase comparisons and 23 of 24 continuation comparisons. The exception is TUAB with supervision removed, seed 1. Resampling units are fixed evaluation batches, not independent subjects.}
\label{tab:s5_recon}
\begin{tabular*}{\linewidth}{@{\extracolsep{\fill}}lrrrr@{}}
\toprule
Corpus & Seed & \shortstack{Supervised start\\(end of first phase)} & \shortstack{Supervision kept\\(end of training)} & \shortstack{Supervision removed\\(end of training)} \\
\midrule
Stieger 2021 & 0 & \shortstack{$+0.34$\\$[+0.20,+0.48]$} & \shortstack{$+0.45$\\$[+0.23,+0.68]$} & \shortstack{$+0.23$\\$[+0.03,+0.44]$} \\
 & 1 & \shortstack{$+0.74$\\$[+0.61,+0.89]$} & \shortstack{$-0.01$\\$[-0.18,+0.18]$} & \shortstack{$+0.23$\\$[+0.04,+0.44]$} \\
 & 2 & \shortstack{$+0.60$\\$[+0.46,+0.74]$} & \shortstack{$-0.02$\\$[-0.20,+0.17]$} & \shortstack{$+0.12$\\$[-0.07,+0.30]$} \\
\addlinespace
Lee2019 ERP & 0 & \shortstack{$+0.00$\\$[-0.10,+0.11]$} & \shortstack{$+0.29$\\$[+0.14,+0.43]$} & \shortstack{$+0.48$\\$[+0.33,+0.63]$} \\
 & 1 & \shortstack{$+0.20$\\$[+0.11,+0.30]$} & \shortstack{$-0.15$\\$[-0.33,+0.03]$} & \shortstack{$+0.08$\\$[-0.10,+0.28]$} \\
 & 2 & \shortstack{$+0.11$\\$[+0.02,+0.20]$} & \shortstack{$-0.09$\\$[-0.23,+0.04]$} & \shortstack{$+0.29$\\$[+0.15,+0.42]$} \\
\addlinespace
ISRUC & 0 & \shortstack{$-0.55$\\$[-0.84,-0.24]$} & \shortstack{$-0.94$\\$[-1.42,-0.53]$} & \shortstack{$-0.63$\\$[-1.06,-0.25]$} \\
 & 1 & \shortstack{$-0.65$\\$[-0.93,-0.38]$} & \shortstack{$-0.78$\\$[-1.12,-0.43]$} & \shortstack{$-0.96$\\$[-1.29,-0.62]$} \\
 & 2 & \shortstack{$-0.66$\\$[-0.94,-0.37]$} & \shortstack{$-0.45$\\$[-0.77,-0.13]$} & \shortstack{$-0.19$\\$[-0.55,+0.15]$} \\
\addlinespace
TUAB & 0 & \shortstack{$+0.39$\\$[+0.25,+0.52]$} & \shortstack{$+0.21$\\$[+0.01,+0.40]$} & \shortstack{$+0.26$\\$[+0.07,+0.45]$} \\
 & 1 & \shortstack{$+0.83$\\$[+0.69,+0.96]$} & \shortstack{$+0.69$\\$[+0.44,+0.93]$} & \shortstack{$+0.84$\\$[+0.60,+1.06]$} \\
 & 2 & \shortstack{$+0.46$\\$[+0.30,+0.60]$} & \shortstack{$+0.48$\\$[+0.25,+0.71]$} & \shortstack{$+0.57$\\$[+0.32,+0.82]$} \\
\bottomrule
\end{tabular*}
\end{table}

\begin{table}[htbp]
\centering
\scriptsize
\setlength{\tabcolsep}{1.5pt}
\renewcommand{\arraystretch}{1.15}
\caption{Relative reconstruction excess on banks A and C, in percent, compared with the reconstruction-only reference at the same training budget. Values are mean $\pm$ sample SD across three encoder seeds, with $\lambda=0.02$. The supervised start is evaluated at the end of the first phase; the two continuations are evaluated at the end of training. Bank A shares encoder-training subject identities; bank C contains test subjects.}
\label{tab:s5_other_banks}
\begin{tabular*}{\linewidth}{@{\extracolsep{\fill}}lrrrrrr@{}}
\toprule
Corpus & \multicolumn{2}{c}{Supervised start} & \multicolumn{2}{c}{Supervision kept} & \multicolumn{2}{c}{Supervision removed} \\
\cmidrule(lr){2-3}\cmidrule(lr){4-5}\cmidrule(lr){6-7}
 & Training (A) & Test (C) & Training (A) & Test (C) & Training (A) & Test (C) \\
\midrule
Stieger 2021 & $+0.921\,\pm\,0.242$ & $+0.490\,\pm\,0.071$ & $+0.379\,\pm\,0.250$ & $-0.028\,\pm\,0.447$ & $+0.212\,\pm\,0.069$ & $+0.165\,\pm\,0.306$ \\
Lee2019 ERP & $+0.314\,\pm\,0.055$ & $+0.100\,\pm\,0.121$ & $+0.361\,\pm\,0.148$ & $+0.016\,\pm\,0.211$ & $+0.639\,\pm\,0.157$ & $+0.354\,\pm\,0.314$ \\
ISRUC & $+0.708\,\pm\,0.235$ & $-0.552\,\pm\,0.220$ & $+0.498\,\pm\,0.371$ & $-0.589\,\pm\,0.428$ & $+0.458\,\pm\,0.568$ & $-0.454\,\pm\,0.264$ \\
TUAB & $+0.633\,\pm\,0.155$ & $+0.480\,\pm\,0.211$ & $+0.529\,\pm\,0.205$ & $+0.372\,\pm\,0.384$ & $+0.589\,\pm\,0.237$ & $+0.487\,\pm\,0.410$ \\
\bottomrule
\end{tabular*}
\end{table}

\subsection{Controls: optimizer history, supervision strength, and readout choice}
\label{app:s5_controls}

\paragraph{Does the result depend on inherited optimizer history?}
Repeating both continuations with a fresh optimizer preserves the direction of the event-task result (Table~\ref{tab:s5_reset}). The subject-level gap favoring supervision kept over supervision removed is $+0.129$ on Stieger and $+0.201$ on ERP, compared with $+0.114$ and $+0.246$ for the corresponding seed-0 pair with inherited optimizer state. The magnitudes differ, particularly for ERP, but both reset gaps have positive 95\% intervals.

When supervision is removed, the endpoint pooled score is similar with a reset and an inherited optimizer: 0.149 versus 0.151 on Stieger, and 0.024 versus 0.003 on ERP. Resetting the reconstruction-only reference changes its endpoint pooled score by less than 0.017 in absolute value on every corpus. Thus, the decline occurs both when optimizer history is preserved and when it is cleared in both branches. The result is not specific to either optimizer setting.

\begin{table}[htbp]
\centering
\footnotesize
\setlength{\tabcolsep}{3pt}
\renewcommand{\arraystretch}{1.15}
\caption{Optimizer-reset controls at the end of training, encoder seed 0 and $\lambda=0.02$. All three conditions shown use a reset optimizer. The first three numeric columns give pooled attention-readout $\kappa$ on $Z$. ``Kept $-$ removed'' is the paired subject-level gap with its 95\% interval. The final column gives the change in the reconstruction-only reference\textquotesingle s pooled score caused by resetting its optimizer. TUAB\textquotesingle s paired gap uses only 18 subjects.}
\label{tab:s5_reset}
\begin{tabular*}{\linewidth}{@{\extracolsep{\fill}}lrrrrr@{}}
\toprule
Corpus & \shortstack{Supervision\\kept} & \shortstack{Supervision\\removed} & \shortstack{Reconstruction\\only} & \shortstack{Kept $-$ removed\\95\% CI} & \shortstack{Reference:\\reset $-$ inherited} \\
\midrule
Stieger 2021 & $0.288$ & $0.149$ & $0.082$ & \shortstack{$+0.129$\\$[+0.098,+0.159]$} & $+0.016$ \\
Lee2019 ERP & $0.248$ & $0.024$ & $-0.006$ & \shortstack{$+0.201$\\$[+0.129,+0.273]$} & $+0.000$ \\
ISRUC & $0.438$ & $0.436$ & $0.421$ & \shortstack{$+0.003$\\$[-0.015,+0.022]$} & $-0.006$ \\
TUAB & $0.514$ & $0.492$ & $0.487$ & \shortstack{$-0.046$\\$[-0.101,+0.005]$} & $-0.014$ \\
\bottomrule
\end{tabular*}
\end{table}

\begin{figure}[htbp]
\centering
\includegraphics[width=\linewidth]
{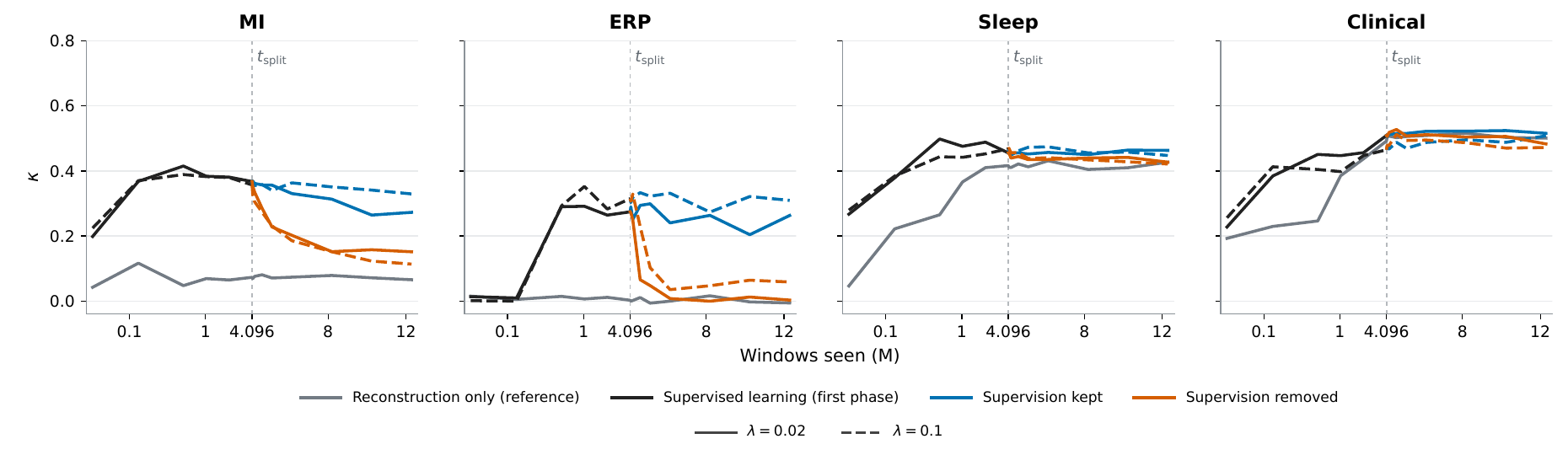}
\caption{
Effect of supervision strength on acquisition and retention. Panels show motor imagery (Stieger 2021), ERP (Lee2019 ERP), sleep staging (ISRUC), and clinical classification (TUAB). Curves report pooled test $\kappa$ from the attention readout on encoder features $Z$, averaged over three readout initializations at each checkpoint.
Solid and dashed curves use $\lambda=0.02$ and $\lambda=0.1$, respectively, both on encoder seed 0. The reconstruction-only reference is shared and shown once.
For each supervision weight, the kept and removed branches inherit the complete training state at $t_{\mathrm{split}}=4.096\times10^6$ window presentations; optimizer-reset controls are not included. The horizontal axis is logarithmic before $t_{\mathrm{split}}$ and linear afterwards, giving equal width to both phases. The checkpoint at zero is omitted. This single-encoder-seed comparison does not estimate variability across encoder initializations.}

\label{fig:s5_lambda_trajectories}
\end{figure}

\paragraph{What changes with stronger supervision?}
Figure~\ref{fig:s5_lambda_trajectories} compares the full training trajectories for $\lambda=0.02$ and $\lambda=0.1$ on the same encoder seed, showing how supervision strength affects both the supervised start and the subsequent continuations.

With $\lambda=0.1$, keeping supervision on Stieger gives a final pooled $\kappa=0.329$, compared with 0.113 after removal, retaining 93\% versus 17\% of the initial advantage. ERP ends at 0.309 versus 0.058, retaining 101\% versus 21\%. The gains with descriptors also remain larger when supervision is kept (Table~\ref{tab:s5_lambda}).

The reconstruction criterion is not satisfied throughout this control. At the supervised start, Stieger exceeds the reconstruction-only reference's bank-B loss by 2.07\%, with a stored 95\% interval of $[1.82,2.31]\%$. ERP, ISRUC, and TUAB remain within the $1\%$ tolerance. By the end of training, the supervision-kept branch has a lower bank-B reconstruction loss than the reference by 1.93\%, 0.99\%, 2.88\%, and 2.16\%, respectively for Stieger, ERP, ISRUC, and TUAB. Stronger supervision can therefore improve retention, although acquisition is not uniformly within the chosen reconstruction tolerance. This control uses only one encoder seed.

\begin{table}[htbp]
\centering
\footnotesize
\setlength{\tabcolsep}{3pt}
\renewcommand{\arraystretch}{1.15}
\caption{Stronger supervision ($\lambda=0.1$), encoder seed 0. Pooled scores use the attention readout on $Z$. Gains with descriptors are paired subject-level comparisons with the reconstruction-only reference at the same training budget, using $[H_{512},Z]$ in both readouts. ``Start'' is the supervised start at the end of the first phase; ``kept'' and ``removed'' are evaluated at the end of training. TUAB gains concern only its 18 eligible subjects.}
\label{tab:s5_lambda}
\begin{tabular*}{\linewidth}{@{\extracolsep{\fill}}lrrrrrr@{}}
\toprule
Corpus & \multicolumn{3}{c}{Pooled score on encoder features} & \multicolumn{3}{c}{Gain with descriptors} \\
\cmidrule(lr){2-4}\cmidrule(lr){5-7}
 & \shortstack{Supervised\\start} & \shortstack{Supervision\\kept} & \shortstack{Supervision\\removed} & \shortstack{Supervised\\start} & \shortstack{Supervision\\kept} & \shortstack{Supervision\\removed} \\
\midrule
Stieger 2021 & $0.357$ & $0.329$ & $0.113$ & $+0.211$ & $+0.185$ & $+0.028$ \\
Lee2019 ERP & $0.315$ & $0.309$ & $0.058$ & $+0.285$ & $+0.273$ & $+0.027$ \\
ISRUC & $0.468$ & $0.448$ & $0.421$ & $+0.012$ & $-0.001$ & $-0.006$ \\
TUAB & $0.465$ & $0.508$ & $0.472$ & $+0.027$ & $+0.026$ & $+0.010$ \\
\bottomrule
\end{tabular*}
\end{table}

\paragraph{Does the pattern persist with other descriptors and readouts?}
Table~\ref{tab:s5_readouts} reports the full-descriptor and fixed-pooling controls, using the same subjects, checkpoint states, and training budgets. With attention and full $H$, the initial gains with descriptors remain $+0.159\pm0.010$ on Stieger and $+0.223\pm0.017$ on ERP. After supervision is removed, the residual gains fall to $+0.031\pm0.011$ and $-0.003\pm0.005$.

Global mean pooling also shows an acquired advantage followed by a decline after supervision removal. The electrode-preserving readout gives a weaker Stieger gap between keeping and removing supervision: $+0.027\pm0.030$, compared with $+0.158\pm0.039$ on ERP. Thus, the pattern extends beyond attention and the compact descriptor block, while its magnitude and consistency depend on the readout. All these classifiers are trained with cross-entropy. Their scores measure predictive performance accessible to the specified readouts; they do not measure all task information in the representation.

\begin{table}[htbp]
\centering
\footnotesize
\setlength{\tabcolsep}{3pt}
\renewcommand{\arraystretch}{1.15}
\caption{Readout and descriptor controls, $\lambda=0.02$. All entries are paired subject-level comparisons, mean $\pm$ sample SD across three encoder seeds. Gains with descriptors compare the indicated encoder with the reconstruction-only reference, supplying the listed descriptors to both readouts. ``Change after removal'' and ``kept $-$ removed'' use $Z$ alone. Initial gains are measured at the end of the first phase; all other columns use the end of training. Attention with full $H$ shares the primary analysis\textquotesingle s $Z$ readout. The electrode-preserving control uses full $H$ only. TUAB estimates concern only 18 eligible subjects.}
\label{tab:s5_readouts}
\begin{tabular*}{\linewidth}{@{\extracolsep{\fill}}lrrrr@{}}
\toprule
Readout / descriptors & \shortstack{Initial gain\\with descriptors} & \shortstack{Change after\\removal} & \shortstack{Kept $-$\\removed} & \shortstack{Final gain with\\descriptors\\(removed)} \\
\midrule
\multicolumn{5}{l}{\textit{Stieger 2021}} \\
Attention, full $H$ & $+0.159\,\pm\,0.010$ & $-0.206\,\pm\,0.014$ & $+0.138\,\pm\,0.022$ & $+0.031\,\pm\,0.011$ \\
Global mean, $H_{512}$ & $+0.184\,\pm\,0.014$ & $-0.171\,\pm\,0.007$ & $+0.110\,\pm\,0.019$ & $+0.055\,\pm\,0.017$ \\
Global mean, full $H$ & $+0.159\,\pm\,0.028$ & $-0.171\,\pm\,0.007$ & $+0.110\,\pm\,0.019$ & $+0.055\,\pm\,0.016$ \\
Electrodes kept, full $H$ & $+0.055\,\pm\,0.023$ & $-0.053\,\pm\,0.017$ & $+0.027\,\pm\,0.030$ & $+0.009\,\pm\,0.053$ \\
\addlinespace
\multicolumn{5}{l}{\textit{Lee2019 ERP}} \\
Attention, full $H$ & $+0.223\,\pm\,0.017$ & $-0.256\,\pm\,0.015$ & $+0.240\,\pm\,0.016$ & $-0.003\,\pm\,0.005$ \\
Global mean, $H_{512}$ & $+0.278\,\pm\,0.029$ & $-0.269\,\pm\,0.030$ & $+0.254\,\pm\,0.017$ & $+0.014\,\pm\,0.028$ \\
Global mean, full $H$ & $+0.265\,\pm\,0.012$ & $-0.269\,\pm\,0.030$ & $+0.254\,\pm\,0.017$ & $-0.005\,\pm\,0.029$ \\
Electrodes kept, full $H$ & $+0.168\,\pm\,0.008$ & $-0.180\,\pm\,0.005$ & $+0.158\,\pm\,0.039$ & $+0.017\,\pm\,0.038$ \\
\addlinespace
\multicolumn{5}{l}{\textit{ISRUC}} \\
Attention, full $H$ & $+0.010\,\pm\,0.004$ & $-0.026\,\pm\,0.003$ & $+0.013\,\pm\,0.022$ & $+0.002\,\pm\,0.004$ \\
Global mean, $H_{512}$ & $+0.005\,\pm\,0.006$ & $-0.019\,\pm\,0.025$ & $+0.000\,\pm\,0.017$ & $+0.013\,\pm\,0.011$ \\
Global mean, full $H$ & $+0.017\,\pm\,0.018$ & $-0.019\,\pm\,0.025$ & $+0.000\,\pm\,0.017$ & $+0.011\,\pm\,0.004$ \\
Electrodes kept, full $H$ & $+0.001\,\pm\,0.011$ & $-0.025\,\pm\,0.005$ & $+0.001\,\pm\,0.018$ & $+0.013\,\pm\,0.020$ \\
\addlinespace
\multicolumn{5}{l}{\textit{TUAB}} \\
Attention, full $H$ & $+0.004\,\pm\,0.024$ & $+0.053\,\pm\,0.040$ & $-0.023\,\pm\,0.044$ & $-0.003\,\pm\,0.002$ \\
Global mean, $H_{512}$ & $-0.005\,\pm\,0.039$ & $+0.010\,\pm\,0.048$ & $-0.022\,\pm\,0.033$ & $-0.011\,\pm\,0.018$ \\
Global mean, full $H$ & $-0.025\,\pm\,0.014$ & $+0.010\,\pm\,0.048$ & $-0.022\,\pm\,0.033$ & $-0.018\,\pm\,0.023$ \\
Electrodes kept, full $H$ & $+0.013\,\pm\,0.005$ & $+0.039\,\pm\,0.019$ & $-0.012\,\pm\,0.061$ & $+0.017\,\pm\,0.045$ \\
\bottomrule
\end{tabular*}
\end{table}

\subsection{Cue-aligned, generic and more numerous unlabeled epochs}
\label{app:s5_task_epochs}

The encoder-training stream of this section is made of the benchmark's labeled windows: one 4-s window per trial, at a fixed position relative to the event ($-0.5$ to $3.5$~s around the motor-imagery cue on Stieger2021, $-1.5$ to $2.5$~s around each stimulus on Lee2019 ERP). The reconstruction-only reference is therefore already self-supervised learning on unlabeled, cue-aligned task epochs. Two arms change only the stream (same subjects, initialization, recipe, masks, budget of 12.288 million presentations and readout). \emph{Generic windows}: as many windows per subject, cut from the continuous recordings of the same subjects at a fixed stride regardless of events, rest periods included. \emph{All aligned epochs}: every cue-aligned epoch of the encoder-training subjects instead of the capped draw, $2.1\times$ more on Stieger2021 and $7.5\times$ more on Lee2019 ERP.

Generic and cue-aligned windows yield the same decoding (Table~\ref{tab:windows_control}): generic minus cue-aligned is $+0.005$ $[-0.006, 0.018]$ on Stieger2021 and $+0.004$ $[-0.018, 0.025]$ on Lee2019 ERP (three seeds per arm, paired by seed, 95\% subject bootstrap). All cue-aligned epochs improve held-out reconstruction (loss 0.334 against 0.351--0.354 on Lee2019 ERP, 0.367 against 0.384--0.385 on Stieger2021), yet leave ERP decoding at chance and raise motor-imagery decoding only from $0.065$ to $0.088$ (one seed). In this setting, exposure to many unlabeled, event-locked task epochs is not what the objective lacks.

\begin{table}[!ht]
\centering
\footnotesize
\setlength{\tabcolsep}{4pt}
\caption{\textbf{Frozen decoding after reconstruction on differently built streams}, by pooling, after 12.288 million presentations. Mean over pretraining seeds (number in parentheses), each averaged over three readout seeds; encoder features alone, test subjects. Labels: supervised start, read at the split (4.1 million presentations).}
\label{tab:windows_control}
\begin{tabular}{llrrr}
\toprule
Corpus & Encoder & Attention & Mean pool & Electrodes kept \\
\midrule
Stieger2021 (MI) & Untrained (3) & 0.014 & 0.098 & 0.142 \\
 & Reconstruction, generic windows (3) & 0.073 & 0.089 & 0.231 \\
 & Reconstruction, cue-aligned epochs (3) & 0.065 & 0.106 & 0.237 \\
 & Reconstruction, all cue-aligned epochs (1) & 0.088 & 0.123 & 0.264 \\
 & Cue-aligned epochs with labels (3) & 0.367 & 0.366 & 0.330 \\
\midrule
Lee2019 ERP & Untrained (3) & 0.012 & 0.012 & 0.112 \\
 & Reconstruction, generic windows (3) & $-0.002$ & 0.014 & 0.066 \\
 & Reconstruction, cue-aligned epochs (3) & $-0.004$ & 0.011 & 0.087 \\
 & Reconstruction, all cue-aligned epochs (2) & $-0.002$ & $-0.004$ & 0.106 \\
 & Cue-aligned epochs with labels (3) & 0.277 & 0.277 & 0.281 \\
\bottomrule
\end{tabular}
\end{table}

\subsection{What the experiment establishes, and how to reproduce it}
\label{app:s5_scope}

\paragraph{Why the design is informative.}
The principal comparison starts from exactly the same learned model and optimizer state, then keeps or removes the supervised term while preserving the reconstruction data, their order, their masks, and the update budget. It therefore measures the effect of continuing that supervision under this training procedure without changing subject exposure or optimizer initialization. Fitting a new readout at every checkpoint also prevents an outdated training-time classifier from being the sole explanation for a decline.

On the two event-related tasks, the results establish a concrete separation between learning a useful representation and retaining its usefulness. Supervision produces a substantial advantage on new subjects; reconstruction-only continuation erodes much of that advantage; and keeping supervision preserves more of it while remaining within the chosen reconstruction tolerance. This is why evaluating retention adds value to evaluating reconstruction quality: reconstruction loss alone does not reveal whether an acquired predictive capability remains accessible to the tested readouts.

\paragraph{Scope of the conclusion.}
The inference is limited to this architecture, training recipe, four corpora, and fixed subject partitions. The A pools are small relative to the presentation budget and undergo extensive reuse. Training with supervision uses additional computation. The optimizer-reset and stronger-supervision controls each use only one encoder seed. TUAB's subject-level inference is particularly restricted because most subjects lack both labels in the small evaluation sample.

Neither the reconstruction tolerance nor the available readouts establishes that all task information is preserved or lost. The experiment does not test transfer to unseen tasks or classes, and it does not establish that reconstruction is the unique limitation of other EEG foundation models.

\paragraph{Reproducibility.}
The implementation separates training, frozen evaluation, and analysis. Training saves the random initialization and the complete supervised start at $t_{\mathrm{split}}$. Evaluation records per-head pooled scores, eligible per-subject scores, selected readout settings, and per-cell reconstruction losses at every checkpoint. The analysis reconstructs the initial-gain, continuation, optimizer-reset, and reconstruction comparisons from those records.

Reported endpoint comparisons use the fixed $t_{\mathrm{split}}$ and $t_{\mathrm{end}}$ budgets. Intermediate checkpoints describe the trajectory and are not used to choose a stopping point that maximizes test performance. All numerical tables in this appendix are based exclusively on the single-corpus experiments described above.

\section{Is the encoder undertrained?}
\label{app:undertrained}

Masked reconstruction might learn the dominant low frequencies first and mu/beta structure only later. We read the stored checkpoints of the Section~\ref{sec:surrogates} REVE-small encoder pretrained on original EEG (seed 0), from initialization to 9.5 times the standard budget of 20.8 million window presentations, with the paper's masks, splits and head; no model was retrained. Reconstruction by band is $R^2_{\text{band}} = 1 - E_{\text{band}}(x-\hat{x}) / E_{\text{band}}(x)$, with $E_{\text{band}}$ the Hann-tapered spectral energy of all hidden one-second patches in the band.

\begin{figure}[!ht]
\centering
\includegraphics[width=0.94\linewidth]{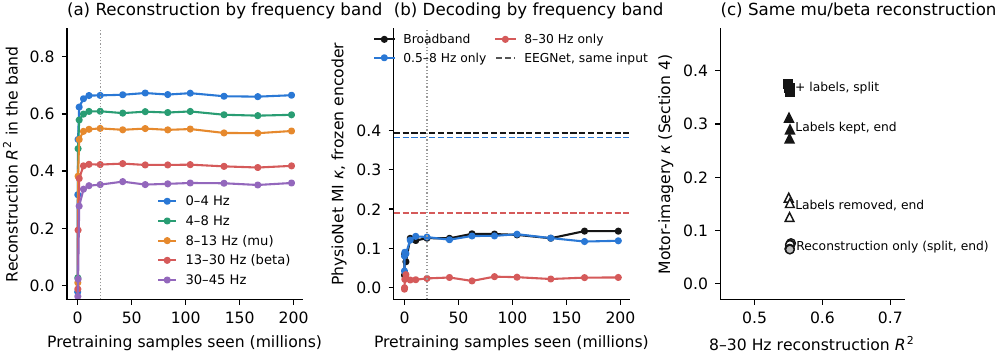}
\caption{\textbf{Reconstruction plateaus early in every band, and reconstruction quality does not decide whether motor imagery is decodable.} (a) Reconstruction $R^2$ by band along one pretraining run (five of the six Section~\ref{sec:surrogates} corpora). (b) PhysioNet MI, frozen head on broadband or band-limited input; dashed: EEGNet on the same input. (c) Section~\ref{sec:retention} encoders, Stieger2021 held-out subjects: 8--30~Hz reconstruction against motor-imagery $\kappa$. Vertical line: standard budget.}
\label{fig:undertrained}
\end{figure}

\paragraph{Every band plateaus early.}
After 90 thousand presentations, only 0--4~Hz is reconstructed ($R^2 = 0.32$), but by 1.4 million, 7\% of the standard budget, every band, mu and beta included, has completed 90\% of its rise (Figure~\ref{fig:undertrained}a). From the standard budget to 9.5 times it, 8--30~Hz reconstruction does not improve ($R^2$ 0.485 to 0.478), and frozen motor-imagery decoding with the paper's head barely moves ($0.126$ to $0.144$ on PhysioNet MI, $+0.018$ $[-0.000, 0.034]$; Figure~\ref{fig:undertrained}b).

\paragraph{The same mu/beta reconstruction supports very different decoding.}
At the split on Stieger2021, the Section~\ref{sec:retention} encoders trained with and without labels decode motor imagery at $\kappa = 0.37$ and $0.07$ on held-out subjects, yet reconstruct 8--30~Hz equally well ($R^2$ $0.551$ and $0.553$; Figure~\ref{fig:undertrained}c): better mu/beta reconstruction is not what the label loss adds.

\paragraph{Near the reconstruction limit, the label is worth little to the objective.}
A perfect reconstruction would contain the task information, but the objective rewards it only by the error the label could remove. On 29 corpora, giving the label (as class-mean waveforms or class spectra and cross-spectra) to the optimal linear predictor of a hidden half-window lowers its error as much as a planted signature of $0.15\sigma$ on sleep and $0.11\sigma$ on clinical EEG, but only $0.03\sigma$ on motor imagery and $0.015\sigma$ on ERPs, below the smallest planted one ($0.05\sigma$). An optimal linear reconstructor thus gains almost nothing from encoding these labels; a non-linear or subject-specific one could gain more.

\clearpage
\section{Do the conclusions persist with few downstream labels?}
\label{app:low_label}

The main experiments use the full annotation budget of each retained benchmark
subset. Because label efficiency is a central motivation for foundation models,
we repeat the three analyses with $K\in\{32,128\}$ labeled windows per class
(validation labels included). The qualitative picture is unchanged: pretraining
gains stay concentrated in sleep and clinical tasks and largely overlap with
classical descriptors; pretraining on signals that preserve spectra and
cross-spectra closely matches real-EEG pretraining; and reconstruction does not
keep most of the extra motor-imagery and ERP information acquired under
supervision. At $K=32$, a small residual pretraining gain remains when both
readouts receive classical features, but the five-family mean gain of adding
pretrained features to classical descriptors alone is small at both budgets,
with intervals that include zero.

\subsection{Label budgets and evaluation protocol}
\label{app:low_label_protocol}

\paragraph{What is reduced.}
We reuse the encoder checkpoints, EEG windows, subject partitions, and test
windows of the corresponding main experiment. Only the labels available to the
downstream classifier are reduced. In the retention experiment, the labels used
to train the encoder on subject group A are unchanged; the budget applies to
the fresh readout fitted on group B and evaluated on group C. These experiments
therefore measure the label efficiency of frozen representations, not
low-label encoder training or fine-tuning.

\paragraph{Sampling and validation.}
For each class, we draw without replacement $3K/4$ windows from the training
subjects and $K/4$ from the validation subjects
(Table~\ref{tab:low_label_budgets}). We use five label draws, nested so that
the $K=32$ subset is contained in the $K=128$ subset. Draws and test windows
are shared across the compared conditions. Model selection uses only the
budgeted validation labels; the selected classifier is then refitted on all
budgeted labels. The budget counts labeled windows, not annotated subjects or
the screening cost of building class-stratified subsets.

\begin{table}[!ht]
\centering
\small
\caption{Nominal downstream label budgets, per class. For $C$ classes, the total
budget is $CK$ windows. Validation is included, rather than supplied for free.}
\label{tab:low_label_budgets}
\begin{tabular}{lrr}
\toprule
 & $K=32$ & $K=128$ \\
\midrule
Training labels per class & 24 & 96 \\
Validation labels per class & 8 & 32 \\
Total labels, binary task & 64 & 256 \\
Total labels, five-class task & 160 & 640 \\
\bottomrule
\end{tabular}
\end{table}

\paragraph{Available labels and unlabeled recordings.}
All corpora reach the nominal $K=32$ budget. At $K=128$, classes with too few
windows contribute all available windows of the relevant partition. In the
published-encoder panel, actual per-class totals are 79--100 for BETA, 120 for
Kalunga2016 and Nakanishi2015, 123--128 for Sleep-EDF, and 47--128 for TUEV;
all other corpora reach 128. Token standardization, descriptor
standardization, and PCA use all downstream training and validation windows
without labels (all of B for retention), identically at every budget; test
windows never enter these fits. This is therefore a label-scarce setting with
access to extra unlabeled recordings, not a setting with only $CK$ recordings.

\paragraph{Readout recipe.}
The readout uses a single learned attention query, initialized at zero, followed by a linear classifier. Full-batch Adam minimizes cross-entropy with learning rate $10^{-2}$, at most 300 updates, and an L2 penalty chosen from $\{10^{-3},10^{-2},10^{-1}\}$. Validation $\kappa$ is computed every five updates to select the penalty and stopping step; ties go to stronger regularization and earlier stopping. Test scores are averaged over head seeds 0, 1, and 2, then over label draws. For the surrogate and retention experiments, we also re-evaluate the full budget with this recipe, so that the low-label curves have a within-recipe reference. These full-budget scores need not match the main-text scores, which used the original readout recipes.

\clearpage
\subsection{Published encoders: the pretraining gain still overlaps with classical descriptors}
\label{app:low_label_panel}
The panel covers 23 encoder configurations on 34 datasets and tasks. The
attention comparison uses 515 model--corpus pairs across 27 corpora in the
five main families, out of 519 in the full-budget panel; the full-budget
reference is restricted to the same pairs. Scores average the available pairs
within each family, then weight the five families equally.

\begin{figure}[!ht]
\centering
\includegraphics[width=\linewidth]{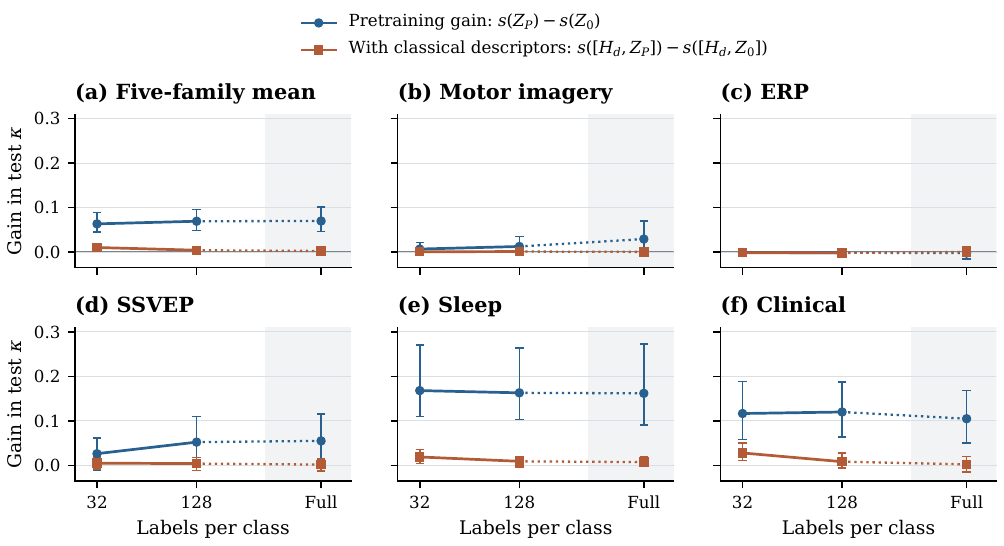}
\caption{\textbf{Pretraining gains under restricted label budgets.}
Attention readouts on the same 515 pairs, averaged over five label draws and
three head seeds. Error bars are 95\% intervals from 2,000 bootstrap resamples
of models and corpora independently within each family; the overall mean
weights families equally. They quantify matrix variation, not subject or draw
uncertainty. Shaded ``Full'' points use the paper's matched pairs with
corpus-dependent annotation counts; horizontal positions are categorical.}
\label{fig:low_label_gains}
\end{figure}

\begin{table}[!ht]
\centering
\small
\setlength{\tabcolsep}{3pt}
\caption{Five-family means with 95\% matrix-bootstrap intervals. $A_H(K)$ is
the gain from adding pretrained features to $H_d$ alone, defined below.
The full-budget $A_H$ is omitted because the paper's standalone $H_d$ score
used a different classifier recipe.}
\label{tab:low_label_overall}
\begin{tabular*}{\linewidth}{@{\extracolsep{\fill}}lccc@{}}
\toprule
Labels / class & $\Delta_{\mathrm{pre}}$ & $\Delta_H$ & $A_H$ \\
\midrule
32 & $+0.063\;[+0.045, +0.089]$ & $+0.010\;[+0.005, +0.017]$ & $+0.004\;[-0.001, +0.013]$ \\
128 & $+0.069\;[+0.049, +0.096]$ & $+0.004\;[-0.001, +0.009]$ & $+0.001\;[-0.003, +0.006]$ \\
Full (paper) & $+0.070\;[+0.047, +0.101]$ & $+0.002\;[-0.003, +0.008]$ & -- \\
\bottomrule
\end{tabular*}
\end{table}

The average pretraining gain is $+0.063$ at 32 labels per class and $+0.069$
at 128, versus $+0.070$ at the full budget: it does not grow as labels become
scarce. Classical descriptors remove about 84\% and 94\% of these gains,
respectively. Motor imagery and ERP remain weakly served; sleep and clinical
tasks account for most of the benefit (Figure~\ref{fig:low_label_gains}).

\clearpage
\subsection{Residual gains and comparison with standalone classical descriptors}
\label{app:low_label_residual}

A positive $\Delta_H$ does not by itself show that pretrained features improve
on $H_d$ alone, because appending random features can hurt the baseline. We
therefore also measure
\begin{equation}
A_H(K)=s_K([H_d,Z_P])-s_K(H_d),
\end{equation}
with the same descriptor block, labels, classifier family, and selection
recipe in both arms. Across the five families, $\Delta_H$ is $+0.010$ at
$K=32$ and $+0.004$ at $K=128$, while $A_H$ is $+0.004$
$[-0.001,+0.013]$ and $+0.001$ $[-0.003,+0.006]$. The positive residual
contrast at $K=32$ therefore does not establish a positive mean improvement
over standalone classical descriptors. At $K=128$, $A_H$ is only $+0.003$ in
sleep and $+0.001$ in clinical tasks.

\begin{figure}[!ht]
\centering
\includegraphics[width=\linewidth]{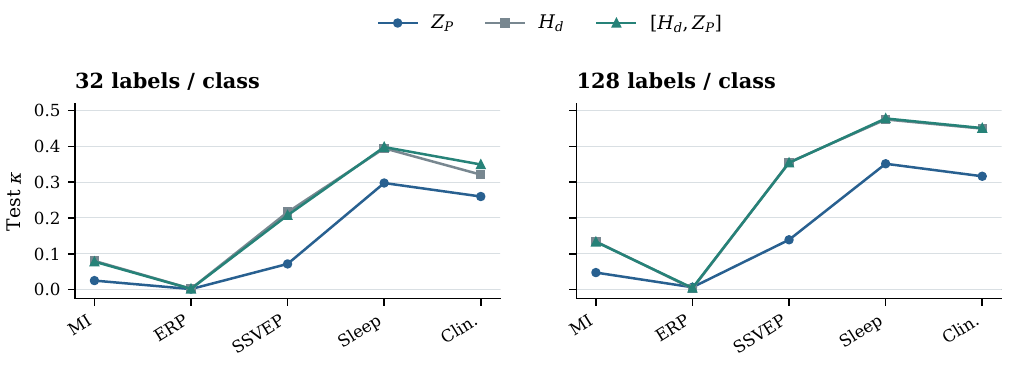}
\caption{\textbf{Classical descriptors remain a strong low-label reference.}
Absolute attention-readout scores, averaged over the available model--corpus
pairs within each family. $H_d$ alone uses cross-entropy fitting here and its
width matches the descriptor block paired with each model. Each point averages
five label draws and three head initializations. Lines connect task families
only to aid visual comparison; they do not represent a continuous variable.}
\label{fig:low_label_scores}
\end{figure}

\paragraph{Absolute scores.}
In motor imagery, SSVEP, and sleep, $H_d$ alone outperforms $Z_P$ at both
restricted budgets. ERP stays near chance. In short, classical descriptors
account for most of the pretraining advantage at both budgets; the small
residual contrast at $K=32$ does not translate into a clear gain over
descriptors alone.

\clearpage
\subsection{Surrogate pretraining transfers with few downstream labels}
\label{app:low_label_surrogates}

We re-evaluate the real-EEG and surrogate pretraining checkpoints on the same
six real-EEG datasets: BNCI2014-001, PhysionetMI, Lee2019 ERP, Lee2019 SSVEP,
ISRUC, and TUAB. For each encoder seed, we average over head initializations
and label draws, then weight the six corpora equally. Mean and standard
deviation are across the three pretraining seeds, except for the
mean-and-variance surrogate and the REVE-base controls, which have one run per
condition.

\begin{figure}[!ht]
\centering
\includegraphics[width=\linewidth]{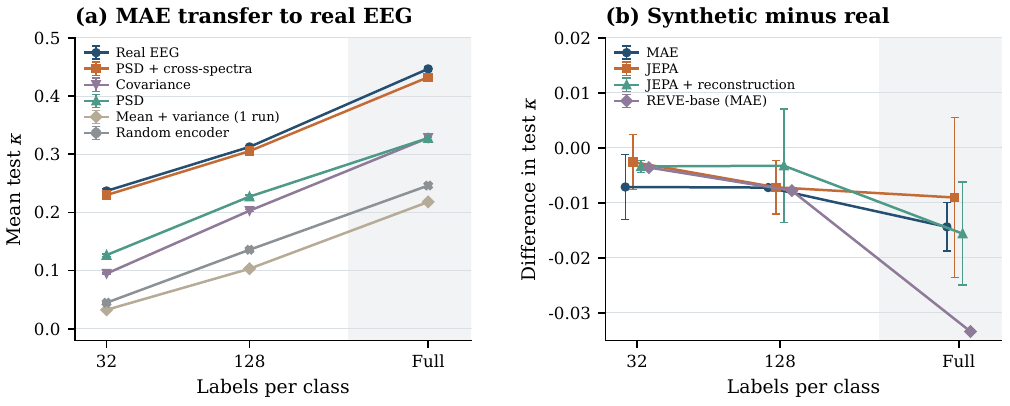}
\caption{\textbf{The real-versus-synthetic transfer comparison persists at low
label budgets.} (a) Six-corpus mean transfer of frozen REVE-small encoders
pretrained by masked reconstruction (MAE). (b) Paired difference between
pretraining with the PSD-plus-cross-spectra surrogate and with real EEG, under
three objectives and the larger REVE-base architecture. Error bars show one
standard deviation across three encoder seeds; the single-run controls have
no uncertainty bar. All downstream inputs are real EEG. The full budget is
re-evaluated with the same classifier recipe as the restricted budgets.}
\label{fig:low_label_surrogates}
\end{figure}

\begin{table}[!ht]
\centering
\small
\setlength{\tabcolsep}{3pt}
\caption{Six-corpus mean transfer $\kappa$ of REVE-small MAE encoders, with
standard deviation across encoder seeds. The random encoder has three
initializations; the mean-and-variance surrogate has one pretraining run.
``Full'' uses the new readout recipe throughout.}
\label{tab:low_label_surrogates}
\begin{tabular*}{\linewidth}{@{\extracolsep{\fill}}lccc@{}}
\toprule
Pretraining signal & 32 / class & 128 / class & Full \\
\midrule
Real EEG & $0.237\,\pm\,0.005$ & $0.313\,\pm\,0.001$ & $0.447\,\pm\,0.004$ \\
PSD + cross-spectra & $0.230\,\pm\,0.002$ & $0.305\,\pm\,0.001$ & $0.432\,\pm\,0.002$ \\
Covariance & $0.095\,\pm\,0.001$ & $0.203\,\pm\,0.002$ & $0.328\,\pm\,0.007$ \\
PSD & $0.127\,\pm\,0.002$ & $0.227\,\pm\,0.003$ & $0.328\,\pm\,0.002$ \\
Mean + variance (1 run) & $0.032$ & $0.103$ & $0.218$ \\
Random encoder & $0.045\,\pm\,0.001$ & $0.136\,\pm\,0.001$ & $0.246\,\pm\,0.005$ \\
\bottomrule
\end{tabular*}
\end{table}

At 32 labels per class, real-EEG pretraining reaches $0.237\,\kappa$ and the
PSD-plus-cross-spectra surrogate $0.230$, against $0.045$ for random weights.
At 128 labels per class, the scores are $0.313$, $0.305$, and $0.136$. The
paired synthetic-minus-real difference, computed before rounding, is $-0.007$
at both restricted budgets and $-0.014$ at the full budget under this readout.
Covariance-only and PSD-only pretraining fall well below real EEG. The same
close comparison holds under JEPA, JEPA plus reconstruction, and REVE-base
(Figure~\ref{fig:low_label_surrogates}b). These are small observed
differences, not a formal equivalence test, but they support the main-text
interpretation: most of the low-label transfer measured here can be learned
from spectra and cross-spectra.

\clearpage
\subsection{Reconstruction still loses the extra information acquired under supervision}
\label{app:low_label_retention}

\begin{figure}[!ht]
\centering
\includegraphics[width=\linewidth]{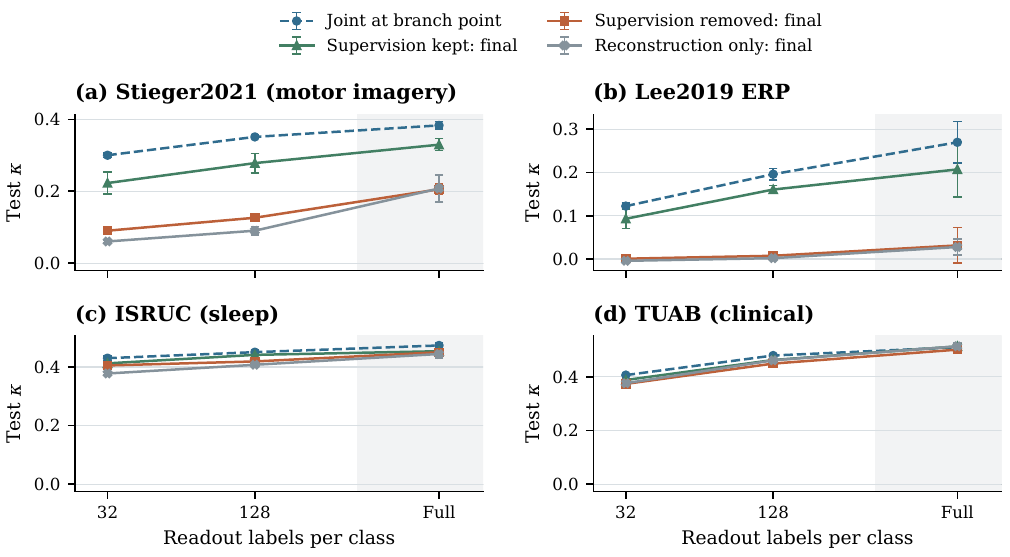}
\caption{\textbf{Restricting readout labels does not remove the retention gap.}
Attention-readout scores at the branch point and final exposure. Encoder
training is unchanged; only fresh readout labels vary. Error bars show one
standard deviation over three encoder seeds. Full-budget points use the same
readout recipe as low-budget points, rather than the original retention readout.}
\label{fig:low_label_retention}
\end{figure}

We reuse the single-corpus bifurcation checkpoints ($\lambda=0.02$, three
encoder seeds). R is trained with reconstruction only. J adds task supervision
until 4.096 million window presentations; from that point, JC keeps
supervision and JR removes it, both running to 12.288 million presentations.
Each checkpoint receives a fresh budgeted readout on B.
Figure~\ref{fig:low_label_retention} reports pooled test $\kappa$, averaged
over five label draws and three readout seeds per encoder.

On Stieger2021 at $K=32$, branch-point scores are $0.059$ for R and $0.300$
for J; at final exposure, R, JC, and JR score $0.060$, $0.223$, and $0.090$.
On Lee2019 ERP, branch-point scores are $0.000$ and $0.122$, and final scores
are $-0.004$, $0.093$, and $0.001$. Most of the supervised advantage is thus
lost once supervision is removed. When both readouts also receive $H_{512}$,
the final JR-over-R contrast stays small at $K=32$ and $K=128$, whereas
keeping supervision preserves a larger contrast
(Table~\ref{tab:low_label_conditional_retention}).

\begin{table}[!ht]
\centering
\small
\setlength{\tabcolsep}{3pt}
\caption{The same comparison when both attention readouts receive $H_{512}$.
Entries are differences of pooled test $\kappa$, mean $\pm$ standard deviation
across three encoder seeds. $P_1$ is the branch point and $T$ is final exposure.
They are not the subject-averaged contrasts reported elsewhere in the paper.}
\label{tab:low_label_conditional_retention}
\begin{tabular*}{\linewidth}{@{\extracolsep{\fill}}llccc@{}}
\toprule
Corpus & $K$ & $J(P_1)-R(P_1)$ & $JC(T)-R(T)$ & $JR(T)-R(T)$ \\
\midrule
Stieger2021 & 32 & $+0.158\,\pm\,0.021$ & $+0.071\,\pm\,0.022$ & $+0.004\,\pm\,0.002$ \\
Stieger2021 & 128 & $+0.180\,\pm\,0.015$ & $+0.110\,\pm\,0.024$ & $+0.021\,\pm\,0.003$ \\
Lee2019 ERP & 32 & $+0.096\,\pm\,0.008$ & $+0.066\,\pm\,0.015$ & $+0.003\,\pm\,0.004$ \\
Lee2019 ERP & 128 & $+0.125\,\pm\,0.006$ & $+0.098\,\pm\,0.019$ & $-0.003\,\pm\,0.003$ \\
\bottomrule
\end{tabular*}
\end{table}

\clearpage
\subsection{Supervised decoding controls and scope of the conclusion}
\label{app:low_label_scope}

\begin{figure}[!ht]
\centering
\includegraphics[width=\linewidth]{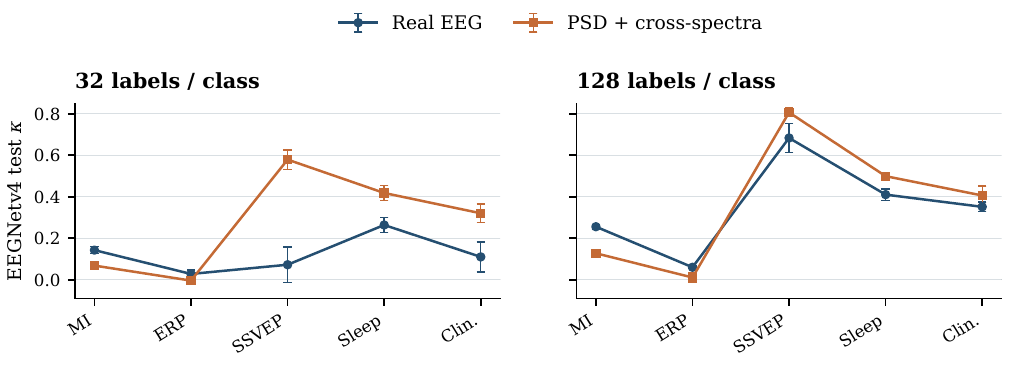}
\caption{\textbf{The supervised surrogate control is not unchanged at low
budgets.} EEGNetv4 trained from scratch for 1,000 updates on all available
budgeted labels, with training and test inputs both real or both synthetic.
Motor imagery averages BNCI2014-001 and PhysionetMI; each other family is
represented by one corpus. Error bars are standard deviations over five draws,
which jointly vary the label subset and network initialization. A fresh
surrogate is sampled for each training batch; validation/test surrogates use
fixed random seeds. These scores are distinct from frozen-encoder transfer.}
\label{fig:low_label_eegnet}
\end{figure}

\paragraph{The supervised surrogate control depends on the budget.}
Frozen transfer should not be confused with training a supervised decoder
directly on each version of the downstream signal. We repeat the EEGNetv4
control on the six corpora, training and testing either on real EEG or on its
PSD-plus-cross-spectra surrogate. At $K=32$, only eight validation windows per
class exist, so validation-based stopping can select very short schedules. We
therefore also run a fixed schedule of 1,000 updates on all budgeted labels,
without validation-based selection (Figure~\ref{fig:low_label_eegnet}).

The surrogate lowers motor-imagery decoding, but the mean degradation seen at
the full budget does not hold for every low-label task. At 32 labels per class
under the fixed schedule, synthetic versus real inputs give roughly $0.58$
versus $0.07$ for SSVEP, $0.42$ versus $0.26$ for ISRUC, and $0.32$ versus
$0.11$ for TUAB. Fresh phase randomization at each batch also acts as a
training-time augmentation, which may help these data-limited decoders, so
these results cannot be read purely as an information-removal effect. ERP is
near chance in both conditions, so this control says nothing about ERP
information at these budgets.

\paragraph{What the low-label results establish.}
Reducing downstream labels does not make the evaluated encoders uniformly
useful across task families. Pretraining still helps mainly where classical
spectral and spatial structure is useful; synthetic pretraining that preserves
this structure still transfers about as well as real-EEG pretraining; and
reconstruction still loses most of the extra motor-imagery and ERP advantage
installed by supervision. A small residual pretraining contrast remains at
$K=32$ when descriptors are given to both readouts, but the mean gain over
standalone descriptors is small at both restricted budgets. The conclusion is
about the structure and task dependence of the benefit; it does not claim that
pretraining never helps with few labels.

\paragraph{Limits.}
The budgets restrict labeled windows, not annotated subjects, and assume
access to extra non-test recordings for label-free preprocessing. All encoders
stay frozen during downstream fitting. We do not extrapolate to fine-tuning,
subject-budget scaling, or an annotation process that must first discover
class membership. Full-budget values are corpus-dependent and are not a common
number of labels. Matrix-bootstrap intervals, encoder-seed standard
deviations, and label-draw standard deviations measure different sources of
variation and are labeled separately throughout.
\section{Where pretraining barely helps, much remains to be learned}
\label{sec:results_headroom}

\begin{table}[t]
\centering
\small
\setlength{\tabcolsep}{4pt}
\caption{\textbf{Where pretraining helps least, supervised decoding goes far beyond the descriptors.} Test Cohen's $\kappa$ on the six datasets of Section~\ref{sec:surrogates} ($\kappa = 0$ at chance). The four frozen-encoder columns average the encoder configurations evaluated on each dataset (16 for BCI IV-2a, 21 for the others), with attention pooling. EEGNetv4 is trained from scratch on the same windows and subject splits; its score averages three training seeds.}
\label{tab:headroom}
\begin{tabular}{llccccc}
\toprule
 & & \multicolumn{4}{c}{Frozen encoders (mean over models)} & Supervised \\
\cmidrule(lr){3-6}\cmidrule(lr){7-7}
Family & Dataset & $Z_0$ & $Z_P$ & $[H_d, Z_0]$ & $[H_d, Z_P]$ & EEGNetv4 \\
\midrule
Motor imagery & BCI IV-2a     & 0.056 & 0.095 & 0.187 & 0.193 & 0.440 \\
              & PhysioNet MI  & 0.077 & 0.081 & 0.190 & 0.186 & 0.395 \\
ERP           & Lee2019 ERP   & 0.006 & 0.005 & 0.004 & 0.001 & 0.341 \\
SSVEP         & Lee2019 SSVEP & 0.271 & 0.377 & 0.654 & 0.653 & 0.894 \\
Sleep         & ISRUC         & 0.298 & 0.432 & 0.591 & 0.595 & 0.537 \\
Clinical      & TUAB          & 0.298 & 0.437 & 0.568 & 0.578 & 0.532 \\
\bottomrule
\end{tabular}
\end{table}

A residual gain near zero has two possible meanings. The descriptors may already capture what the task requires, leaving nothing for pretraining to add; or the task may depend on information that neither the descriptors nor the pretrained features provide. Telling them apart requires a reference for how well each task can be decoded from the same windows. We therefore train a supervised network, EEGNetv4 \citep{eegnet}, from scratch on the same windows and splits of the six datasets we use again in Section~\ref{sec:surrogates} (Table~\ref{tab:headroom}).

On sleep staging (ISRUC) and abnormality detection (TUAB), $[H_d, Z_0]$ already scores at least as high as EEGNetv4, leaving little headroom relative to this supervised reference. On motor imagery and ERP, EEGNetv4 scores substantially higher than $[H_d, Z_0]$, while pretrained features add little on average beyond the descriptors. On Lee2019 ERP, descriptors and pretrained features remain near chance, while EEGNetv4 reaches $\kappa = 0.34$. SSVEP also shows a substantial gap that pretrained features do not close. The frozen readout can nevertheless detect substantial gains when task-relevant features are learned with supervision: in Section~\ref{sec:retention}, with descriptors supplied to both heads, the paired comparison detects gains of $0.20$~$\kappa$ on motor imagery and $0.26$ on ERP for an encoder trained with task labels.

Under these readouts, the supervised EEGNet reference reveals substantial remaining headroom on MI and ERP, while pretrained features add little on average beyond the classical descriptors.

\section{Replication of section~\ref{sec:surrogates} results with LaBraM and its native pretraining objective}
\label{app:s4_labram_short}

\paragraph{Why this experiment.}
Section~\ref{sec:surrogates} keeps the REVE-small encoder fixed and
varies only the pretraining signal and objective. This isolates the
effect of the signal, but it leaves open whether the finding depends on
REVE. Here we test the same question on a second architecture,
LaBraM-base~\citep{labram}, trained with its own discrete prediction
objective.

\paragraph{Why REVE is the main backbone.}
\label{app:why_reve}
Pretraining EEG foundation models from scratch is expensive, so we could not run every architecture under every objective, signal, and seed. We chose REVE for the main comparisons because it is the strongest model in our five-family benchmark and a leading model in NeuralBench
and EEG-FM-Bench~\citep{neuralbench,eegfmbench}. With the REVE-small
encoder fixed, we varied the objective across masked reconstruction
(MAE), masked latent prediction (JEPA), and their combination
(inspired by EEGPT~\citep{eegpt}), which covers the main
reconstruction and latent-prediction families.

The main grid is 3~objectives $\times$ 4~signal conditions $\times$
3~seeds, i.e.\ 36 pretraining runs. We added four REVE-base MAE runs
(one per signal condition; 69.2M vs.\ 12.7M parameters) and one
REVE-small MAE run on the Mean + variance control, for 41 runs at the
standard budget. Separate runs test a fivefold larger budget for real
EEG and PSD + cross-spectra. Repeating all of this for a second
architecture was out of budget, so we instead ran the targeted
replication below.

\paragraph{Protocol.}
LaBraM predicts discrete codes from a learned VQ-NSP tokenizer. For
each signal condition we train both the tokenizer and the encoder from
scratch, with a separate tokenizer per condition. Both stages use the
same TUEG data as the REVE experiments and about 20.8 million window
presentations. We then freeze the encoder and evaluate transfer on the
same six real-EEG dataset--task combinations with the same
attention-pooling readout. Table~\ref{tab:s4_labram_short} lists the
number of completed runs per condition.

\begin{table}[!htbp]
\centering
\small
\caption{Frozen transfer with LaBraM to real EEG.
Each run's score averages three readout initializations and weights
the six dataset--task combinations equally. Values are mean
$\pm$ sample SD across $n$ pretraining runs, or $n$ encoder
initializations for the random baseline. No SD is reported for
the single-run Mean + variance condition. The Covariance results
include one run with tokenizer collapse.}
\label{tab:s4_labram_short}
\begin{tabular}{lcc}
\toprule
\textbf{Pretraining signal} & $n$ & \textbf{Transfer $\kappa$} \\
\midrule
Random initialization & 3 & $0.144 \pm 0.006$ \\
Real EEG & 3 & $0.234 \pm 0.008$ \\
PSD + cross-spectra & 3 & $0.249 \pm 0.014$ \\
PSD & 2 & $0.202 \pm 0.017$ \\
Covariance & 2 & $0.181 \pm 0.078$ \\
Mean + variance & 1 & $0.118$ \\
\bottomrule
\end{tabular}
\end{table}

\paragraph{Results.}
PSD + cross-spectra reaches a mean transfer score of $0.249$~$\kappa$,
against $0.234$ for real EEG and $0.144$ for random initialization.
The paired synthetic-minus-real difference is
$+0.015 \pm 0.019$~$\kappa$ (mean $\pm$ sample SD over three seeds).
PSD and Covariance score lower, and Mean + variance does not beat
random initialization. The Covariance estimate is noisy and includes
one run with tokenizer collapse.

The pattern matches Section~\ref{sec:surrogates}: with a second
architecture and a discrete prediction objective, pretraining on
signals that preserve whole-window spectra and cross-spectra matches
the average transfer of real EEG. As before, this is an aggregate
result; ERP performance stays near chance in both conditions.

\section{Beyond REVE: published encoders, other architectures and other objectives}
\label{app:beyond_reve}

\begin{figure}[!ht]
\centering
\includegraphics[width=0.85\linewidth]{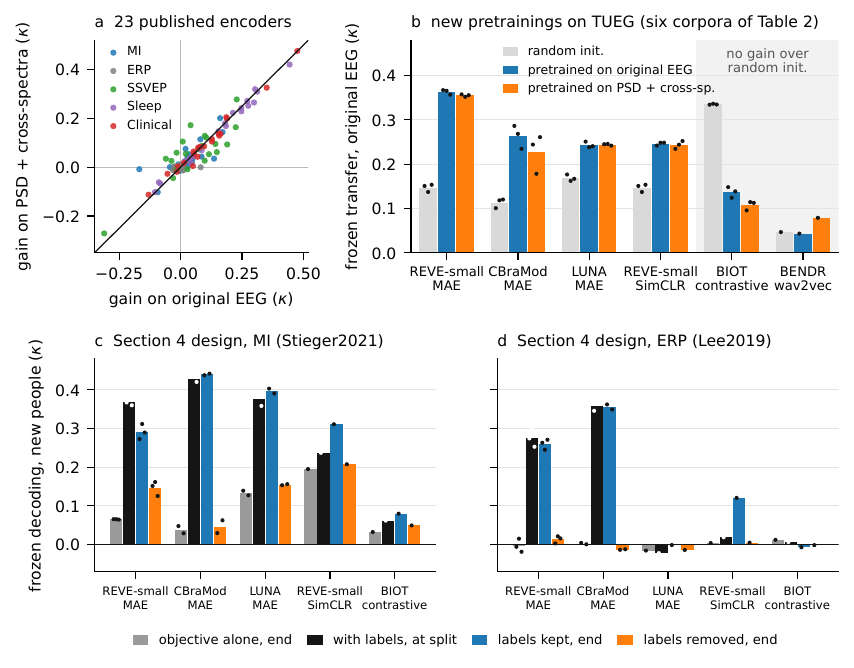}
\caption{\textbf{The mechanism holds beyond REVE.} \textbf{(a)} The 23 published encoders of Figure~\ref{fig:benchmark_gains}, frozen: gain over the random copy when the 27 five-family corpora are read on original EEG ($x$) or on their PSD + cross-spectra surrogate ($y$), with an attention head trained, selected and tested on that signal. One point per encoder and family. \textbf{(b)} New pretrainings on TUEG of other architectures and objectives, on original EEG or on PSD + cross-spectra, read frozen on the six original-EEG corpora of Table~\ref{tab:synthetic_objective_controls}. Dots: pretraining seeds. \textbf{(c, d)} The design of Section~\ref{sec:retention} with other encoders, $\lambda = 0.02$: frozen attention readout on test subjects for the objective alone (end of training), the encoder trained with labels (at the split), and its two continuations (end). Dots: encoder seeds.}
\label{fig:beyond_reve}
\end{figure}

Sections~\ref{sec:surrogates} and~\ref{sec:retention} use REVE. We test their conclusions on other encoders in three ways (Figure~\ref{fig:beyond_reve}), with the paper's readouts and pretraining budget (20.8 million window presentations).

\textbf{Published encoders read on surrogates.} We read every published encoder of Figure~\ref{fig:benchmark_gains} and its random copy on PSD + cross-spectra surrogates of the 27 downstream corpora (Appendix~\ref{app:s4_joint}), with the Section~\ref{sec:results} attention head trained, selected and tested on the surrogate. If the gain from pretraining relied on structure beyond spectra and cross-spectra, it would shrink there. It does not (Figure~\ref{fig:beyond_reve}a): surrogate minus original gain lies between $-0.005$ and $+0.007$~$\kappa$ in every family, with 95\% intervals that include zero, and the two gains correlate at $r = 0.95$ over the 107 encoder--family pairs. The test can fail: on six corpora, PSD-only surrogates, which drop the cross-spectra (Appendix~\ref{app:s4_indep}), lower the gain by $0.036$~$\kappa$ [$0.007$, $0.065$] on clinical EEG and by $0.083$ [$0.047$, $0.122$] on SSVEP (21 encoders).

\textbf{New pretrainings.} We pretrained, on original EEG and on PSD + cross-spectra, CBraMod~\citep{cbramod} (criss-cross attention over per-electrode patches) and LUNA~\citep{luna} (electrodes compressed into learned queries) with masked reconstruction; REVE-small with SimCLR~\citep{simclr}, an instance-contrastive objective on two augmented views of a window (augmentations of SeqCLR,~\citealp{seqclr}); and BIOT~\citep{biot} and BENDR~\citep{bendr} with their native contrastive objectives. REVE-small with masked reconstruction was re-run in the same pipeline as their anchor ($0.363$ and $0.355$~$\kappa$ on original EEG and PSD + cross-spectra; $0.347$ and $0.349$ in Table~\ref{tab:synthetic_objective_controls}). Wherever pretraining helps, PSD + cross-spectra gives the same transfer as original EEG (Table~\ref{tab:beyond_reve_pretraining}): within $0.008$~$\kappa$ on the six corpora for REVE-small, LUNA and SimCLR, against gains of $0.07$ to $0.22$ over random initialization; on all 27 corpora, original EEG is slightly ahead by $0.007$ to $0.015$~$\kappa$, as for the paper's REVE-small runs ($0.009$). CBraMod's six-corpus difference comes from Lee2019 SSVEP, where its attention head is unstable across seeds on both signals; without that corpus it is $-0.002 \pm 0.005$, and a readout that keeps the electrodes gives both signals $\kappa = 0.79$ there. BIOT and BENDR gain nothing from pretraining, like their published checkpoints, so they leave no gain to explain.

\begin{table}[!ht]
\centering
\scriptsize
\setlength{\tabcolsep}{1.6pt}
\caption{\textbf{Original EEG against PSD + cross-spectra pretraining, other architectures and objectives.} Six corpora: frozen attention readout of Table~\ref{tab:synthetic_objective_controls}, mean $\pm$ SD over pretraining seeds; $\Delta$ is surrogate minus original within seed. 27 corpora: Section~\ref{sec:results} attention head, mean over pretraining seeds (one for LUNA, BIOT and BENDR), 95\% interval resampling corpora within families. All runs are read at 20.8 million presentations. $^{\dagger}$BIOT reads 16 bipolar derivations and skips the corpora that cannot build them (4 of 6, 20 of 27).}
\label{tab:beyond_reve_pretraining}
\begin{tabular}{llcccccc}
\toprule
\textbf{Encoder} & \textbf{Objective} & \textbf{Seeds} & \textbf{Random} & \textbf{Original} & \textbf{PSD + cross-sp.} & $\boldsymbol{\Delta}$, six corpora & $\boldsymbol{\Delta}$, 27 corpora \\
\midrule
REVE-small & MAE & 3 & $0.147$ & $0.363 \pm 0.006$ & $0.355 \pm 0.003$ & $-0.008 \pm 0.006$ & $-0.012$ [$-0.020$, $-0.004$] \\
CBraMod & MAE & 3 & $0.113$ & $0.263 \pm 0.026$ & $0.228 \pm 0.044$ & $-0.035 \pm 0.059$ & $-0.012$ [$-0.019$, $-0.005$] \\
LUNA & MAE & 3 & $0.169$ & $0.243 \pm 0.007$ & $0.244 \pm 0.002$ & $+0.001 \pm 0.007$ & $-0.015$ [$-0.034$, $+0.007$] \\
REVE-small & SimCLR & 3 & $0.147$ & $0.246 \pm 0.004$ & $0.244 \pm 0.009$ & $-0.003 \pm 0.006$ & $-0.007$ [$-0.017$, $+0.003$] \\
BIOT$^{\dagger}$ & native, contrastive & 3 & $0.335$ & $0.137 \pm 0.012$ & $0.107 \pm 0.010$ & $-0.029 \pm 0.022$ & $-0.042$ [$-0.062$, $-0.024$] \\
BENDR & native, wav2vec-style & 1 & $0.047$ & $0.043$ & $0.079$ & $+0.036$ & $+0.023$ [$+0.006$, $+0.039$] \\
\bottomrule
\end{tabular}

\end{table}

\textbf{Acquisition and retention.} We repeated the design of Section~\ref{sec:retention} on Stieger2021 and Lee2019 ERP with the same windows, subjects, $\lambda$, split and budget (Figure~\ref{fig:beyond_reve}c, d). Labels raise frozen motor-imagery decoding from $0.02$ to $0.43$ with CBraMod and from $0.14$ to $0.38$ with LUNA, and CBraMod's ERP decoding from $0.02$ to $0.36$. Reconstruction alone then removes almost all of the gain (at most 14\% retained), while continued supervision keeps it. With the contrastive objectives, the label term barely changes decoding at the split, so retention cannot be measured; nor for LUNA on ERP, whose decoding of new people stays at chance.

\textbf{Other pretexts on the target corpus.} Table~\ref{tab:target_pretexts} trains two more pretexts in the setting of Section~\ref{sec:retention} (same people, cue-aligned windows, budget, initial weights and readouts). SimCLR decodes motor imagery better than reconstruction ($0.22$ against $0.07$~$\kappa$) and stays at chance on ERP; trained instead on a fresh PSD + cross-spectra surrogate of every window and read on original windows, it nearly matches ($0.19$; paired difference over training $-0.02$ $[-0.05, -0.00]$), so what it learns about motor imagery is carried mostly by spectra and cross-spectra. An objective that second-order statistics cannot solve, telling each window from a surrogate that keeps its spectra, cross-spectra and amplitude values (one random phase per frequency shared by all channels, then rank matching), is learned and transfers to new people, yet decodes motor imagery no better than reconstruction, and ERP stays at chance.

\begin{table}[!ht]
\centering
\scriptsize
\setlength{\tabcolsep}{4pt}
\caption{\textbf{Pretexts trained on the target corpus} (Section~\ref{sec:retention} setting). Frozen decoding of new people, $\kappa$, at the split (4.1 million windows), every encoder read on original windows; attention, global-mean and electrodes-kept poolings on motor imagery (Stieger2021), attention on ERP (Lee2019). Mean over encoder seeds (SimCLR on ERP: one seed).}
\label{tab:target_pretexts}
\begin{tabular}{lccccc}
\toprule
 & & \multicolumn{3}{c}{Motor imagery} & ERP \\
\cmidrule(lr){3-5}\cmidrule(lr){6-6}
Pretext & Seeds & Attention & Global mean & Electrodes kept & Attention \\
\midrule
Untrained & 3 & $0.01$ & $0.10$ & $0.14$ & $0.01$ \\
Reconstruction & 3 & $0.07$ & $0.12$ & $0.26$ & $0.01$ \\
SimCLR, original EEG & 3 & $0.22$ & $0.27$ & $0.23$ & $0.00$ \\
SimCLR, PSD + cross-spectra & 3 & $0.19$ & $0.25$ & $0.23$ & -- \\
Real vs.\ surrogate & 1 & $0.07$ & $0.16$ & $0.22$ & $0.00$ \\
Reconstruction + labels ($\lambda = 0.02$) & 3 & $0.37$ & $0.37$ & $0.33$ & $0.28$ \\
\bottomrule
\end{tabular}
\end{table}

\section{Full fine-tuning evaluation}
\label{app:full_finetuning}

\subsection{Why we mainly focus on frozen encoders}
\label{app:why_frozen}
Our main focus in this paper is studying the predictive value accessible in an encoder before supervised adaptation. Our objective is to characterize what pretraining
encodes in the weights rather than to maximize downstream performance.
Keeping the encoder fixed and training only the readout makes this possible:
the representation under evaluation is the one produced by pretraining, so
the predictive signal we measure reflects what that representation makes
accessible to a small readout. This choice lets us compare pretrained and
randomly initialized representations, assess their predictive overlap with
classical EEG descriptors, and track the effects of pretraining and
continuation without allowing target labels to reshape the encoder. The
conclusions concern the representations and readouts evaluated under this
protocol.

Full fine-tuning evaluates the utility of a pretrained initialization when
target labels also update the encoder. Because supervised training modifies
the encoder, the resulting representation mixes what pretraining produced
with what the target labels induced, and predictive signal can no longer be
attributed to pretraining alone; this is why the interpretation and
diagnosis in the main paper rely on frozen encoders. Fine-tuning instead
addresses a different question: comparing pretrained and random
initializations under the same adaptation protocol measures a benefit of
pretraining as an initialization in this setting, including its interaction
with supervised learning. We examine this question for REVE-base on five
task families, with two label budgets.

\paragraph{Setup.}
We compare the published REVE-base checkpoint with a randomly
initialized copy on PhysioNet MI \cite{physionet_eegmmidb}, ERP~\cite{lee2019openbmi}, SSVEP, ISRUC,
and TUAB, using the paper's windows, subject-disjoint splits, and
attention-pooling head, without classical descriptors. We evaluate
direct end-to-end training (FT) and head training with a frozen encoder
followed by end-to-end training (LP-FT), at the full label budget and
at 32 labeled windows per class. Results weight the five corpora equally.
Within each adaptation protocol, the pretrained and random
initializations receive the same optimization and hyperparameter search
budget; encoder learning rates and checkpoints are selected separately
on validation performance.

We use AdamW with a head learning rate of $10^{-3}$ and weight decay
of $10^{-4}$. Encoder learning rates are
$\{3\!\times\!10^{-5},10^{-4},3\!\times\!10^{-4},10^{-3}\}$ at the
full budget and $\{10^{-5},3\!\times\!10^{-5},10^{-4}\}$ at 32 labels
per class. Full-budget learning-rate selection averages validation
scores over three training seeds. The smaller budget uses five label
draws, each with 24 training and 8 validation windows per class, and
one training seed per draw. Fine-tuning does not refit on validation
data, whereas the frozen references retain the main paper's
train--validation refit. Comparisons between frozen and adapted scores
should therefore be interpreted with this protocol difference in mind.

\begin{figure}[t]
    \centering
    \includegraphics[width=\linewidth]{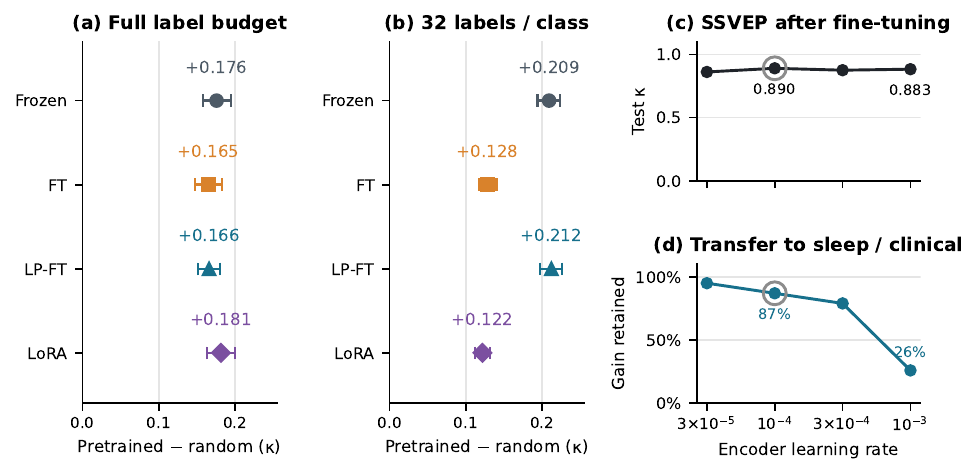}
    \caption{\textbf{Pretraining as a frozen representation and as an
    initialization for adaptation.}
    \textbf{(a--b)} Mean pretrained--random gap across five equally
    weighted corpora, with attention pooling and without classical
    descriptors. Error bars are 95\% paired test-subject bootstrap
    intervals, conditional on the fitted runs. Frozen heads use three
    seeds; FT, LP-FT and LoRA use three training seeds at the full budget
    and one seed for each of five label draws at 32 labels per class.
    \textbf{(c--d)} SSVEP fine-tuning from pretrained weights, with one
    training seed and four encoder learning rates; the ring marks the
    rate selected on SSVEP validation performance. Retained gain
    averages $(s_{\mathrm{after}}-s_0)/(s_P-s_0)$ over ISRUC and TUAB,
    using fresh frozen readouts. Here, $s_P$ and $s_0$ are the
    pretrained and random frozen references for each task, and
    $s_{\mathrm{after}}$ is the score after SSVEP fine-tuning.}
    \label{fig:full_finetuning}
\end{figure}

\paragraph{Absolute performance and the pretraining advantage.}
At the full label budget, FT improves mean absolute performance for both
initializations relative to their frozen references
($+0.051$~$\kappa$ for pretrained weights and $+0.061$ for random
weights). The mean pretrained--random gap is $+0.176$~$\kappa$ with
frozen encoders, $+0.165$ after FT, and $+0.166$ after LP-FT
(Figure~\ref{fig:full_finetuning}a). At 32 labels per class, the
corresponding gaps are $+0.209$, $+0.128$, and $+0.212$ (panel~b).
Thus, under the main attention-pooling readout, adaptation can improve
downstream scores without a correspondingly larger average advantage
from pretrained initialization.

This aggregate pattern does not establish equivalence between the
protocols: task-specific gains vary, and with global mean pooling the
full-budget gap increases from $+0.122$~$\kappa$ frozen to $+0.200$
after FT. The effect of adaptation therefore depends on the readout
and training procedure.

\paragraph{Parameter-efficient adaptation.}
We also adapt both initializations with LoRA~\citep{hu2022lora}: the encoder weights stay frozen, and rank-8 adapters ($\alpha = 16$) on the four linear layers of each of the 22 transformer blocks are trained together with a fresh attention head (1.4\,M trained encoder parameters, 2.1\% of the encoder). Everything else follows the FT protocol above, including seeds, label draws, early stopping and the selection of the learning rate on validation, with an adapter learning-rate grid that is the same for both initializations and extended upward whenever either selected its top value. At the full budget, the mean pretrained--random gap with LoRA is $+0.181$~$\kappa$ $[0.163, 0.199]$, close to the frozen ($+0.176$), FT ($+0.165$) and LP-FT ($+0.166$) gaps (Figure~\ref{fig:full_finetuning}a and Table~\ref{tab:lora}). The LoRA-adapted pretrained encoder remains below EEGNetv4 on motor imagery ($0.384$ against $0.395$) and ERPs ($0.284$ against $0.341$). With 32 labels per class, LoRA loses about 40\% of the advantage, as FT does ($+0.12$ and $+0.13$~$\kappa$, against $+0.21$ frozen and with LP-FT), mostly on SSVEPs ($+0.07$, against $+0.38$ with LP-FT; Figure~\ref{fig:full_finetuning}b). With few labels, a fresh head trained jointly with the encoder thus erodes the advantage whether all weights or 2\% of them are adapted, whereas training the head first preserves it.

\begin{table}[!ht]
\centering
\footnotesize
\setlength{\tabcolsep}{4pt}
\caption{\textbf{Pretraining advantage under four adaptation regimes.} Left: pretrained minus random $\kappa$ per task family, attention pooling, no descriptors (95\% paired test-subject bootstrap intervals for the means in Figure~\ref{fig:full_finetuning}a--b). Right: REVE-base after LoRA and after FT, and EEGNetv4 trained from scratch.}
\label{tab:lora}
\begin{tabular}{llccccccc}
\toprule
& & \multicolumn{4}{c}{Pretrained $-$ random ($\kappa$)} & \multicolumn{3}{c}{REVE-base ($\kappa$)} \\
\cmidrule(lr){3-6}\cmidrule(lr){7-9}
Budget & Family & Frozen & LP-FT & FT & LoRA & LoRA & FT & EEGNetv4 \\
\midrule
Full & Motor imagery & $+0.053$ & $+0.118$ & $+0.120$ & $+0.110$ & $0.384$ & $0.379$ & $0.395$ \\
 & ERP & $+0.079$ & $+0.101$ & $+0.100$ & $+0.215$ & $0.284$ & $0.209$ & $0.341$ \\
 & SSVEP & $+0.232$ & $+0.156$ & $+0.174$ & $+0.167$ & $0.886$ & $0.896$ & $0.894$ \\
 & Sleep & $+0.231$ & $+0.200$ & $+0.166$ & $+0.163$ & $0.556$ & $0.552$ & $0.537$ \\
 & Clinical & $+0.284$ & $+0.254$ & $+0.264$ & $+0.253$ & $0.590$ & $0.586$ & $0.532$ \\
 & Mean & $+0.176$ & $+0.166$ & $+0.165$ & $+0.181$ & $0.540$ & $0.524$ & $0.540$ \\
\midrule
32 / class & Motor imagery & $-0.008$ & $+0.012$ & $-0.013$ & $-0.009$ & $0.032$ & $0.019$ & $0.147$ \\
 & ERP & $-0.002$ & $+0.005$ & $+0.005$ & $-0.007$ & $0.006$ & $0.007$ & $0.028$ \\
 & SSVEP & $+0.336$ & $+0.379$ & $+0.077$ & $+0.067$ & $0.145$ & $0.133$ & $0.072$ \\
 & Sleep & $+0.325$ & $+0.279$ & $+0.273$ & $+0.261$ & $0.399$ & $0.389$ & $0.264$ \\
 & Clinical & $+0.395$ & $+0.384$ & $+0.300$ & $+0.296$ & $0.325$ & $0.345$ & $0.110$ \\
 & Mean & $+0.209$ & $+0.212$ & $+0.128$ & $+0.122$ & $0.181$ & $0.178$ & $0.124$ \\
\bottomrule
\end{tabular}

\end{table}

\paragraph{Target-task performance and transfer after adaptation.}
In a separate diagnostic, we fine-tune the pretrained encoder on SSVEP,
freeze it again, and fit fresh readouts on ISRUC and TUAB. Across four
encoder learning rates, SSVEP scores range from $0.861$ to
$0.890$~$\kappa$, while transfer to the other two tasks varies more
substantially (Figure~\ref{fig:full_finetuning}c--d). At learning rates
$10^{-4}$ and $10^{-3}$, SSVEP scores are $0.890$ and $0.883$, but the
mean fraction of the original pretraining gain retained on ISRUC and
TUAB is $87\%$ and $26\%$, respectively. The validation-selected rate
is $10^{-4}$. This single-seed diagnostic illustrates that similar
target-task scores can coexist with different transfer performance
after adaptation; it does not establish that fine-tuning inevitably
degrades transfer.

\paragraph{Convergence and cost.}
The same runs record validation $\kappa$ after every period of optimizer steps (about one epoch at the full budget). Pretrained weights speed up fine-tuning where the frozen readout already shows their benefit: on SSVEP, sleep and clinical EEG, the fine-tuned REVE-base reaches 90\% of its best validation score within 1 to 2.3 epochs, against 5 to 9 for the random initialization. On motor imagery, both reach most of their final level within three epochs. On ERPs, the pretrained model improves over 10 to 20 epochs, while the random initialization stays near $0.1$~$\kappa$. An epoch costs the same for both initializations, and a full-budget run takes 83~s against 107~s on one A40 GPU, because early stopping ends the pretrained runs sooner. The fine-tuned pretrained model remains below EEGNetv4 trained from scratch on motor imagery ($0.379$ against $0.395$) and ERPs ($0.209$ against $0.341$). At 32 labels per class, its advantage over the random initialization on these two families is at most $0.013$~$\kappa$ in absolute value under the frozen, FT and LP-FT protocols.

\paragraph{Scope.}
These experiments cover one published checkpoint, one random encoder
initialization, and five datasets under the tested adaptation budgets.
The full campaign, including pooling controls and head-only runs,
required approximately 62 GPU-hours. They provide a complementary
assessment of pretraining as an initialization, while the frozen
analyses characterize its accessible predictive value before
task-specific encoder adaptation. The two evaluations thus address
complementary aspects of the contribution of pretraining.

\end{document}